\pdfoutput=1

\documentclass[10pt,twocolumn,letterpaper]{article}

\usepackage{cvpr}              

\definecolor{cvprblue}{rgb}{0.21,0.49,0.74}
\usepackage[pagebackref,breaklinks,colorlinks,allcolors=cvprblue]{hyperref}
\usepackage{hyperref}
\usepackage{url}
\usepackage{listings}
\usepackage{multirow}
\usepackage{graphicx}
\usepackage{caption}
\usepackage{subcaption}
\usepackage{makecell}
\usepackage{longtable}
\usepackage{array}     
\usepackage{graphicx}  
\usepackage{lineno}

\usepackage{wrapfig} 
\usepackage{amssymb}
\usepackage{pifont}

\RequirePackage[breakable,skins]{tcolorbox}
\RequirePackage{xcolor}

\usepackage{array}
\usepackage{multirow}
\usepackage{amssymb}
\usepackage{caption}
\usepackage{subcaption}
\usepackage{booktabs}
\usepackage{pifont}
\usepackage{colortbl}
\usepackage{newtxtext}
\usepackage{twemojis}
\newlength\savewidth\newcommand\shline{\noalign{\global\savewidth\arrayrulewidth
  \global\arrayrulewidth 1pt}\hline\noalign{\global\arrayrulewidth\savewidth}}
\newcommand{\tablestyle}[2]{\setlength{\tabcolsep}{#1}\renewcommand{\arraystretch}{#2}\centering\footnotesize}

\newcommand{\dalle}{DALL$\cdot$E3 \xspace}

\newcommand{\ourapproach}{\textsc{BizGen}\xspace}
\newcommand{\ourbenchmark}{\textsc{BizEval}\xspace}
\newcommand{\ourinfo}{\textsc{Infographics-650K}\xspace}
\newcommand{\ourslide}{\textsc{Slides-500K}\xspace}

\definecolor{colorcommentbg_layerprompt}{HTML}{ABDCC1}
\definecolor{colorcommentframe_layerprompt}{HTML}{2E7D6F}
\newenvironment{layerprompt}[1][]{
\begin{tcolorbox}[adjusted title={Layer-wise Quality Assurance Prompt for GPT-4o}, fonttitle={\bfseries\footnotesize}, fontupper=\scriptsize, colback={colorcommentbg_layerprompt!30}, colframe={colorcommentframe_layerprompt!80},coltitle={white},#1]
}{\end{tcolorbox}}

\definecolor{colorcommentbg_mainlayerprompt}{HTML}{CCCCFF}
\definecolor{colorcommentframe_mainlayerprompt}{HTML}{76608A}
\newenvironment{mainlayerprompt}[1][]{
\begin{tcolorbox}[adjusted title={Select dominant object layer Prompt for GPT-4o}, fonttitle={\bfseries\footnotesize}, fontupper=\scriptsize, colback={colorcommentbg_mainlayerprompt!30}, colframe={colorcommentframe_mainlayerprompt!80},coltitle={white},#1]
}{\end{tcolorbox}}

\definecolor{colorcommentbg_infoqualityprompt}{HTML}{A0522D}
\definecolor{colorcommentframe_infoqualityprompt}{HTML}{6D1F00}
\newenvironment{infoqualityprompt}[1][]{
\begin{tcolorbox}[adjusted title={Infographics Global Visual Quality Assurance Prompt for GPT-4o}, fonttitle={\bfseries\footnotesize}, fontupper=\scriptsize, colback={colorcommentbg_infoqualityprompt!30}, colframe={colorcommentframe_infoqualityprompt!80},coltitle={white},#1]
}{\end{tcolorbox}}

\definecolor{colorcommentbg_slidequalityprompt}{HTML}{87CEEB}
\definecolor{colorcommentframe_slidequalityprompt}{HTML}{00BFFF}
\newenvironment{slidequalityprompt}[1][]{
\begin{tcolorbox}[adjusted title={Slides Global Visual Quality Assurance Prompt for GPT-4o}, fonttitle={\bfseries\footnotesize}, fontupper=\scriptsize, colback={colorcommentbg_slidequalityprompt!30}, colframe={colorcommentframe_slidequalityprompt!80},coltitle={white},#1]
}{\end{tcolorbox}}
\newcommand\codeurl[1]{{{\color{blue}{\url{#1}}}}}

\def\confName{CVPR}
\def\confYear{2025}

\title{\Large \textsc{BizGen}: Advancing Article-level Visual Text Rendering for\\ Infographics Generation}

\author{\normalsize  Yuyang Peng$^{*}$\; Shishi Xiao$^{*}$ \; Keming Wu$^{*}$ \; Qisheng Liao$^{*}$ \; Bohan Chen$^{*}$ \; Kevin Lin \; Danqing Huang \; Ji Li \; Yuhui Yuan\\
  \vspace{1em}
 \normalsize Tsinghua University,\; Brown University,\; University of Liverpool,\; Microsoft Research Asia,\; Microsoft
 \\[-1em]
\small \codeurl{https://bizgen-msra.github.io}}

\begin{document}

\twocolumn[{%
\renewcommand\twocolumn[1][]{#1}%
\maketitle

\begin{center}
\begin{minipage}[t]{1\linewidth}
\centering
\vspace{-9mm}
\captionsetup{type=figure} 
\begin{subfigure}[b]{0.19\textwidth}
\includegraphics[width=\textwidth]{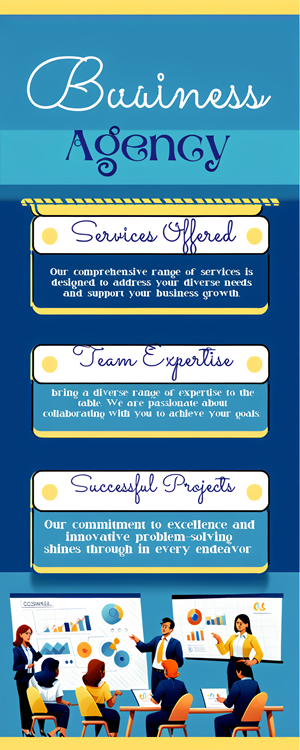}
\caption{\scriptsize{$\underline{386}$ characters / OCR: $\underline{93\%}$}}
\end{subfigure}
\hfill
\begin{subfigure}[b]{0.19\textwidth}
\includegraphics[width=\textwidth]{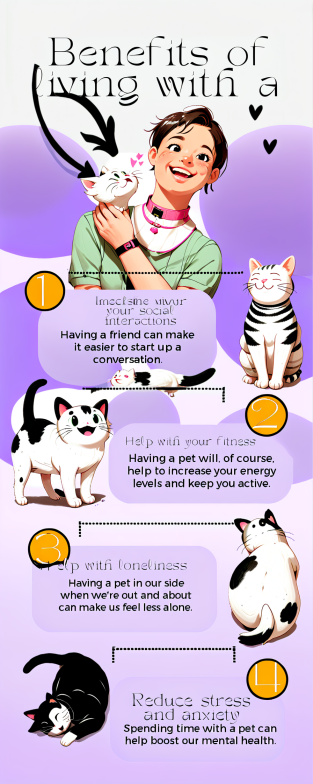}
\caption{\scriptsize{$\underline{426}$ characters / OCR: $\underline{96\%}$}}
\end{subfigure}
\hfill
\begin{subfigure}[b]{0.19\textwidth}
{\includegraphics[width=\textwidth]{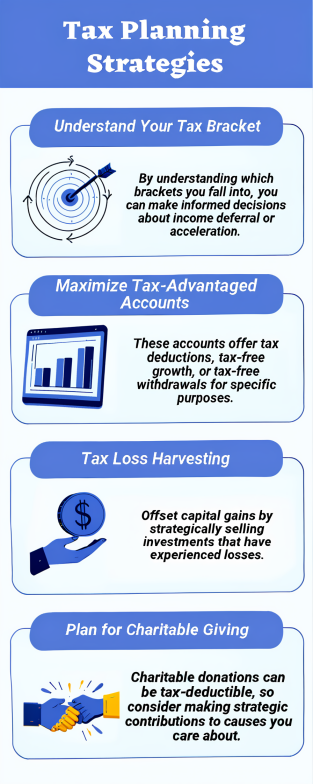}}
\caption{\scriptsize{$\underline{545}$ characters / OCR: $\underline{99\%}$}}
\end{subfigure}
\hfill
\begin{subfigure}[b]{0.19\textwidth}
{\includegraphics[width=\textwidth]{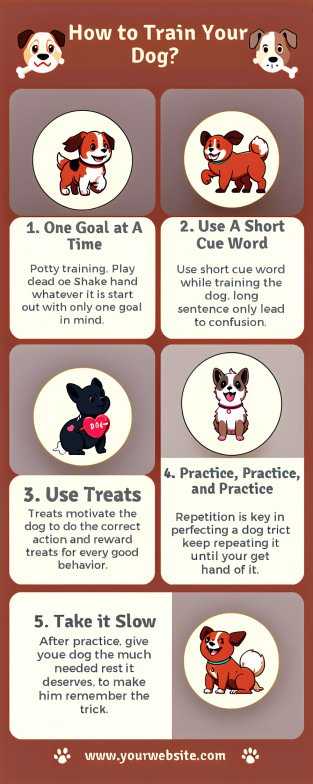}}
\caption{\scriptsize{$\underline{594}$ characters / OCR: $\underline{99\%}$}}
\end{subfigure}
\hfill
\begin{subfigure}[b]{0.19\textwidth}
{\includegraphics[width=\textwidth]{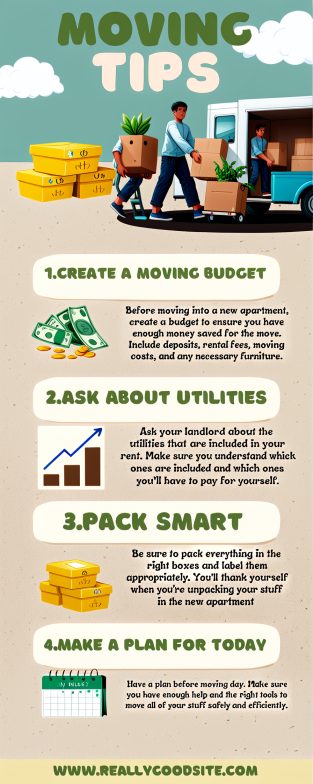}}
\caption{\scriptsize{$\underline{737}$ characters / OCR: $\underline{97\%}$}}
\end{subfigure}
\vspace{-2mm}
\caption{\footnotesize{\textbf{Accurate article-level visual text rendering results} in infographics generated with \ourapproach, ranging from $386$ to $737$ characters.}}
\label{fig:teaser}
\end{minipage}
\end{center}
}]
\renewcommand{\thefootnote}{\relax} 
\footnotetext{$^{*}$ Research intern at Microsoft. Corresponding author: \\
\twemoji{e-mail} \texttt{researcher.yuanyuhui@gmail.com}}

\begin{abstract}
Recently, state-of-the-art text-to-image generation models, such as FLUX and Ideogram 2.0, have made significant progress in sentence-level visual text rendering. In this paper, we focus on the more challenging scenarios of article-level visual text rendering and address a novel task of generating high-quality business content, including infographics and slides, based on user provided article-level descriptive prompts and ultra-dense layouts.  The fundamental challenges are twofold: significantly longer context lengths and the scarcity of high-quality business content data.

 In contrast to most previous works that focus on a limited number of sub-regions and sentence-level prompts, ensuring precise adherence to ultra-dense layouts with tens or even hundreds of sub-regions in business content is far more challenging. We make two key technical contributions: (i) the construction of scalable, high-quality business content dataset, i.e., \textsc{Infographics-650K}, equipped with ultra-dense layouts and prompts by implementing a layer-wise retrieval-augmented infographic generation scheme; and (ii) a layout-guided cross attention scheme, which injects tens of region-wise prompts into a set of cropped region latent space according to the ultra-dense layouts, and refine each sub-regions flexibly during inference using a layout conditional CFG. We demonstrate the  strong results of our system compared to previous SOTA systems such as FLUX and SD3 on our \ourbenchmark prompt set. Additionally, we conduct thorough ablation experiments to verify the effectiveness of each component. We hope our constructed \ourinfo and \ourbenchmark can encourage the broader community to advance the progress of business content generation.

\end{abstract}
    
\section{Introduction}
\label{sec:intro}

\fboxsep=0.1mm
\fboxrule=1pt

\begin{figure}[!t]
\begin{minipage}[!t]{1\linewidth}
\begin{subfigure}[b]{0.19\textwidth}
\centering
\includegraphics[width=1\textwidth]{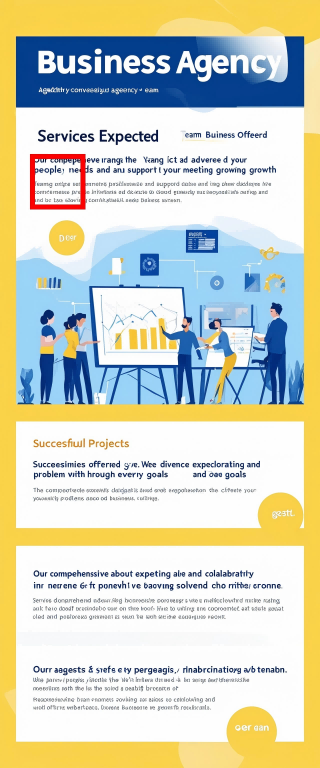}
\fcolorbox{red}{red}{\includegraphics[width=0.95\textwidth]{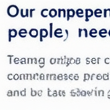}}
\vspace{-3mm}
\caption{\scriptsize{$\underline{386}$ characters / OCR: $\underline{11\%}$}}
\end{subfigure}
\hfill
\begin{subfigure}[b]{0.19\textwidth}
\centering
\includegraphics[width=1\textwidth]{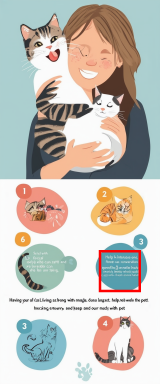}
\fcolorbox{red}{red}{\includegraphics[width=0.95\textwidth]{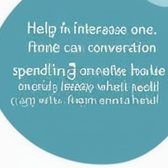}}
\vspace{-3mm}
\caption{\scriptsize{$\underline{426}$ characters / OCR: $\underline{30\%}$}}
\end{subfigure}
\hfill
\begin{subfigure}[b]{0.19\textwidth}
\centering
\includegraphics[width=1\textwidth]{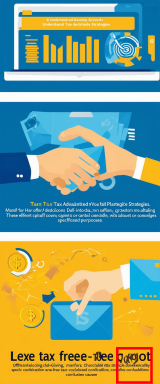}
\fcolorbox{red}{red}{\includegraphics[width=0.95\textwidth]{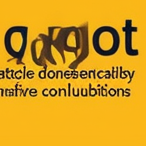}}
\vspace{-3mm}
\caption{\scriptsize{$\underline{545}$ characters / OCR: $\underline{22\%}$}}
\end{subfigure}
\hfill
\begin{subfigure}[b]{0.19\textwidth}
\centering
\includegraphics[width=1\textwidth]{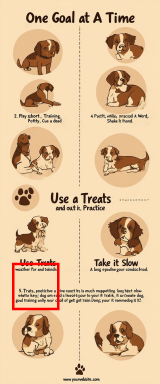}
\fcolorbox{red}{red}{\includegraphics[width=0.95\textwidth]{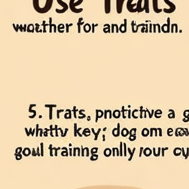}}
\vspace{-3mm}
\caption{\scriptsize{$\underline{594}$ characters / OCR: $\underline{55\%}$}}
\end{subfigure}
\hfill
\begin{subfigure}[b]{0.19\textwidth}
\centering
\includegraphics[width=1\textwidth]{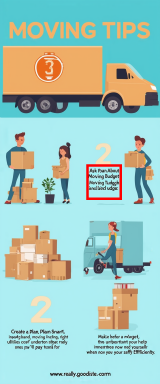}
\fcolorbox{red}{red}{\includegraphics[width=0.95\textwidth]{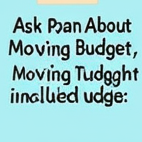}}
\vspace{-3mm}
\caption{\scriptsize{$\underline{737}$ characters / OCR: $\underline{46\%}$}}
\end{subfigure}
\vspace{-2mm}
\end{minipage}
\caption{ \footnotesize{
\textbf{Infographics generation results based on SD3-Large}. While these results appear appealing, the spelling accuracy of the article-level visual text is inadequate, as shown in the zoomed-in marked rectangle regions in the second row.
}}
\label{fig:intro_compare_infographics}
\vspace{-2mm}
\end{figure}

Recently, graphic design generation~\cite{jia2023cole,inoue2024opencole,kikuchi2024multimodal,cheng2024graphic,lin2023designbench,chen2023textdiffuser2,yang2023idea2img,ma2023glyphdraw,yang2023glyphcontrol,ji2023improving,chendiffute,zhao2023udifftext} has attracted increasing interest, as modern text-to-image generation models like SD3, DALL$\cdot$E3, FLUX, and Ideogram 2.0 demonstrate unprecedented capabilities in following complex prompts and rendering accurate visual text. Graphic design encompasses a diverse array of media, including posters, flyers, diagrams, charts, infographics, and slides, each tailored for distinct contexts. In this study, we concentrate on two of the most complex forms of graphic design: infographics and slides. These particular media types, especially infographics, demand advanced visual planning and sophisticated reasoning skills, as they are crafted to communicate complex and article-level messages effectively to their audiences. Our empirical analysis reveals that even the latest models, such as FLUX and Ideogram 2.0, continue to struggle with generating usable infographics due to inaccurate visual text rendering. Figure~\ref{fig:intro_compare_infographics} shows representative infographic examples generated by the latest SD3 Large. We observe that these models frequently exhibit significant artifacts, including disorganized layout, numerous visual text spelling errors, and generally poor aesthetics, when compared to the visual results generated with our approach in Figure~\ref{fig:teaser}.

\begin{figure}[t]
\centering
\includegraphics[width=0.9\columnwidth]{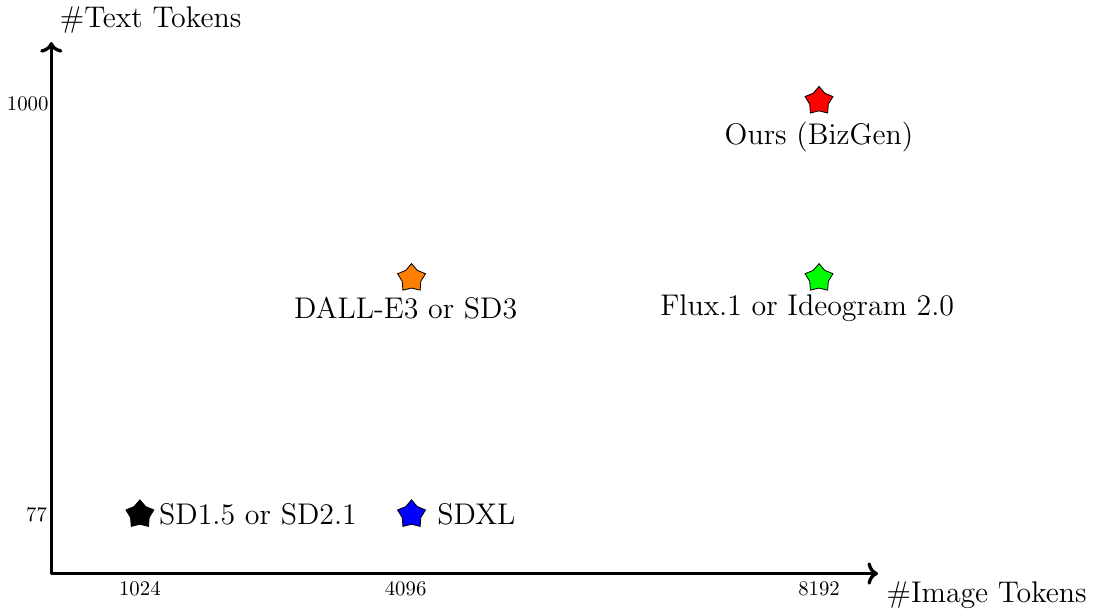}
\caption{ \footnotesize{
\textbf{Significant challenge of business content generation}: the context lengths increase by more than $10\times$ compared to the previous text-to-image generation models.
}}
\label{fig:context_length}
\vspace{-3mm}
\end{figure}

We analyze the fundamental challenges of infographics generation from the following two perspectives:
(i) \emph{context length}: Infographics require a significantly longer context, such as article-level prompt, to be accurately described. Typically, more than $500$ words are needed to describe an infographic in detail, whereas conventional image generation tasks often require fewer than $50$ words. Furthermore, a typical infographic has a resolution of $2240\times896$ to ensure the accurate and dense rendering of visual text, which is nearly twice the size of the conventional $1024\times1024$ resolution adopted by existing text-to-image generation systems. Figure~\ref{fig:context_length} illustrates the comparisons of the context length with previous image generation tasks.
(ii) \emph{data scarcity}: High-quality infographics data is much more challenging to collect. The availability of infographics data on the internet is scarce, primarily due to the substantial human effort and commercial licensing required to obtain it. Consequently, amassing a large dataset of infographics is considerably more difficult compared to other types of graphic design images.

To tackle these two fundamental challenges, we make two key contributions: \underline{First}, we develop an innovative and scalable infographics data engine. This engine generates a comprehensive dataset \ourinfo of high-quality, high-resolution infographics, effectively addressing the issue of data scarcity. This dataset comprises over $\sim650,000$ samples, with each sample containing not only detailed global captions and images but also ultra-dense layouts and region-specific captions. \underline{Second}, unlike the previous Glyph-ByT5~\cite{liu2024glyph}, which only supports layer-wise control for visual text, we introduce a more flexible layout-guided cross attention design that supports complicated layer-wise control for any visual elements rather than only visual text. This pipeline can transform the original extremely long article-level prompt to image generation into a series of shorter, more manageable sentence-level prompts to region generation tasks. We further devise a layout conditional CFG method during inference to eliminate artifacts in every sub-region. By leveraging the relationship between ultra-dense layouts and their corresponding shorter region-specific captions, our approach effectively addresses the challenge of much longer context lengths.

\begin{figure}[t]
\begin{minipage}[t]{1\linewidth}
\centering
\begin{subfigure}[b]{0.99\textwidth}
\includegraphics[width=\textwidth]{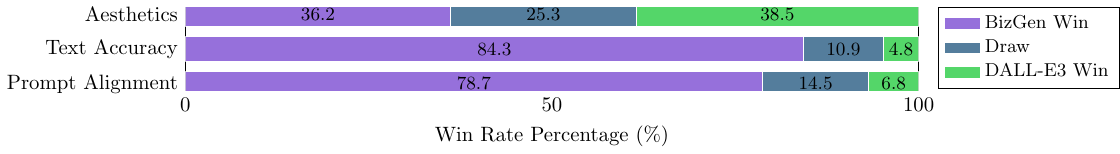}
\vspace{-5mm}
\caption{\scriptsize{\ourapproach v.s. DALL$\cdot$E3}}
\end{subfigure}
\begin{subfigure}[b]{0.99\textwidth}
{\includegraphics[width=\textwidth]{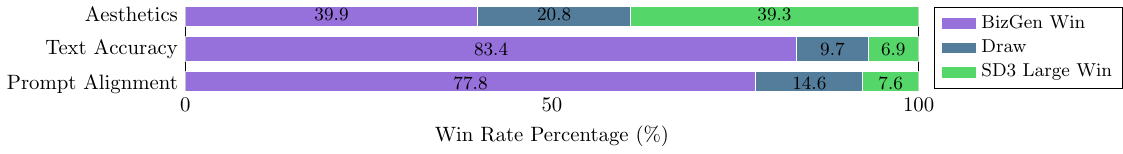}}
\vspace{-5mm}
\caption{\scriptsize{\ourapproach v.s. SD3 Large}}
\end{subfigure}
\begin{subfigure}[b]{0.99\textwidth}
{\includegraphics[width=\textwidth]{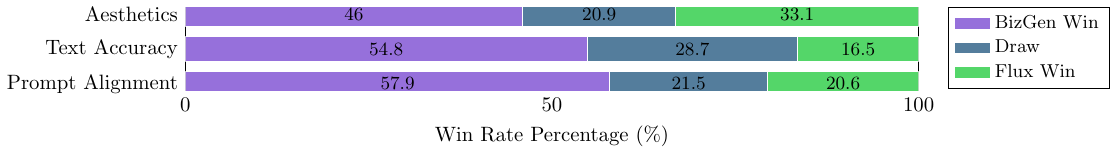}}
\vspace{-5mm}
\caption{\scriptsize{\ourapproach v.s. FLUX}}
\end{subfigure}
\vspace{-3mm}
\caption{\footnotesize{
\textbf{\ourapproach v.s. SOTAs on \ourbenchmark}: Illustrating the win-lose rates based on our user study, which collected feedback from $\sim10$ users.
}}
\label{fig:user_study}
\end{minipage}
\vspace{-5mm}
\end{figure}

To validate the effectiveness of our approach, we build a challenging \textsc{BizEval} benchmark that requires generating accurate article-level visual text in infographics and slides. We empirically show that our approach significantly outperforms the latest state-of-the-art methods like FLUX in terms of visual text rendering and accurate region-wise control. Additionally, we find that previous approaches suffer from severe layer missing issues, whereas our approach can ensure the generation of almost all the requested layers. We also conduct a thorough user study to assess the generated business content from three aspects: visual aesthetics, visual text spelling accuracy, and prompt following. We demonstrate the advantages of our approach and report the detailed user study results in Figure~\ref{fig:user_study}.

\section{Related Work}
\label{sec:related_work}

\subsection{Typography Generation}
Typography generation is known to be a fundamental limitation for the latest text-to-image generation models. Several efforts \cite{yang2024glyphcontrol, ma2023glyphdraw, zhao2023udifftext, Liu2022CharacterAwareMI, chen2024textdiffuser, chen2023textdiffuser2, Paliwal2024CustomTextCT, Lakhanpal2024RefiningTG, Tuo2023AnyTextMV, Li2024EmpoweringBM, Zhu2024VisualTG} have been made to improve the visual text spelling accuracy, but most of them still only handle visual texts with a limited length of tens of characters. State-of-the-art text-to-image generation models, such as FLUX \cite{flux}, Ideogram 2.0\footnote{\textcolor{magenta}{https://about.ideogram.ai/2.0}}, and Stable Diffusion 3 \cite{Esser2024ScalingRF}, have made significant improvements in sentence-level visual text rendering, but spelling accuracy can drop dramatically when the textual content becomes longer. The Glyph-ByT5 \cite{liu2024glyph, liu2024glyphv2} series pretrains a ByT5 model to encode visual text and fine-tunes a LoRA to empower the SDXL model with the capacity to handle textual content longer than 100 characters. These methods still cannot tackle article-level visual text rendering tasks, which typically contain tens of sentence-level text boxes in one image—quite common in infographics and slides. In this study, we focus on the challenging article-level visual text rendering scenario.

\subsection{Layout Guided Image Generation}
Due to the limited ability of representing spatial layout in text prompt, prior works add fine grained spatial controls in the generation process either through encoding layout information\cite{li2023gligen,avrahami2023spatext,yang2023reco,zhang2023adding, wang2024instancediffusion, zheng2023layoutdiffusion}, or through introducing plug-and-play modules that inject visual guidance\cite{zhang2023adding,mou2024t2i}.
In addition to introducing auxiliary modules and extra training, some research manipulates latent or cross-attention maps according to spatial or semantic constraints during inference\cite{chen2024training, xie2023boxdiff, phung2024grounded}.
Currently, some work incorporates LLMs as visual planmner to construct a reasonable layout, guiding the next generation of high-faithfulness images\cite{lian2023llm, yang2024mastering, qu2023layoutllm, Zala2023DiagrammerGPT}
However,these methods struggle to generate satisfactory results in our scenario because of the ultra-dense distribution and heavily overlapping of elements in infographics and slides. Besides, their training pipelines rely on annotating for existing data to get layout-image pairs, which is not scalable. In contrast, we propose a scalable data engine to obtain high-quality, high-resolution infographic data, and introduce a flexible layout-guided latent fusion method to model the complex layout information.
\section{Our Approach}
\label{sec:approach}

In Section~\ref{sec:approach_1}, we illustrate the formulation of the business content generation task by taking infographics as an example and analyze the fundamental challenges of building a business content generation model. In Section~\ref{sec:approach_2}, we explain the first key contribution of this work: a scalable infographics data engine. In Section~\ref{sec:approach_3}, we further introduce a flexible layout-guided layer-wise latent fusion method that supports ultra-dense region-wise control for business content.
Last, we explain the implementation of the \textsc{BizGen} framework based on the aforementioned designs and introduce a layout conditional CFG scheme for flexible control of the layer-wise generation quality.

\subsection{Business Content Generation}
\label{sec:approach_1}
\begin{figure}[t]
\begin{minipage}{0.5\textwidth}
\centering
\includegraphics[width=1\linewidth]{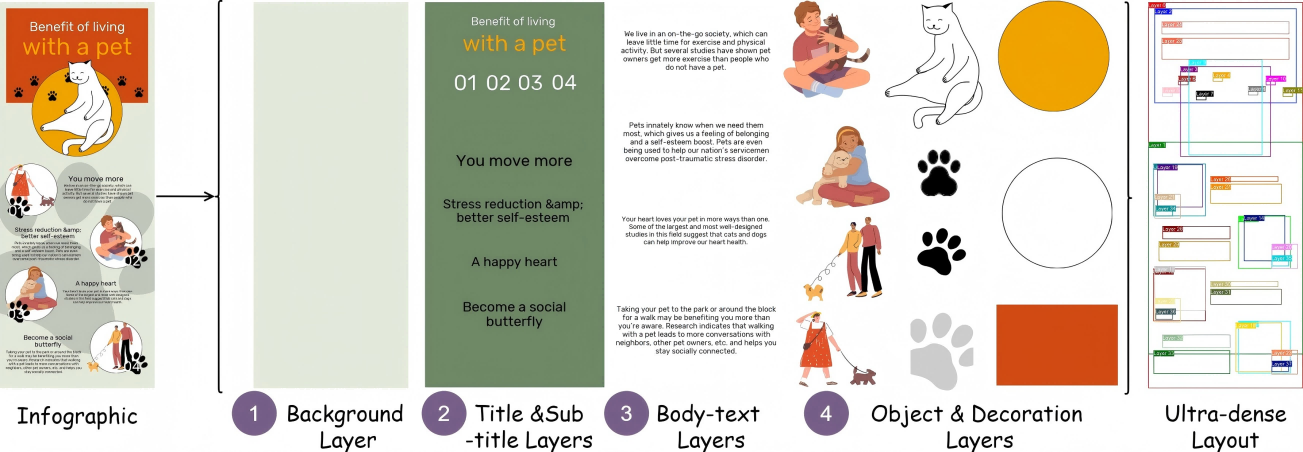}
\caption{ \footnotesize{\textbf{Decomposed visual elements (layers) that form a representative infographic sample}. The right side displays the ultra-dense layout that specifies the spatial arrangements of each visual element.
}}
\label{fig:infographic_example}
\end{minipage}
\end{figure}

In this paper, we study the generation of complex business content, including infographics and slides, from article-level text prompts. Figure~\ref{fig:infographic_example} shows a representative example of an infographic. It consists of multiple layers, including a background layer, several visual text layers, object layers, and decorative layers. The main distinctions between business content generation and previous poster generation lie in (i) precise spatial arrangement of tens of visual design elements and grouping relevant layers together, and (ii) the accurate rendering of article-level visual text formed by title, subtitle, and body text comprising hundreds of words.

\vspace{1mm}
\noindent\textbf{Key Challenges} As shown in Figure~\ref{fig:intro_compare_infographics}, when generating infographics, it is evident that the SOTA generative models exhibit significant artifacts, such as disorganized layout arrangements, numerous visual text spelling errors, and generally poor aesthetics. We attribute these challenges to two primary factors: \emph{the scarcity of high-quality business content data} and \emph{the significantly longer context length}. We address these two fundamental challenges through the following key contributions.

\subsection{Scalable Infographics Data Engine}
\label{sec:approach_2}
We collect an internal dataset consisting of over 5,000 infographics, obtaining raw data that includes all the layers and their layout information. We then propose a scalable infographics data engine comprising two stages: the construction of a high-quality transparent layer database and the implementation of retrieval-augmented infographic generation, resulting in the generation of over 1 million infographics, as detailed below.

\vspace{1mm}
\noindent\textbf{High-Quality Transparent Layer Generation.}
First, we create over 250K design-oriented layer-wise prompts and apply the latest LayerDiffuse\citep{zhang2024transparent} to generate high-resolution transparent graphic design layers based on the recently post-trained SDXL \cite{liang2024step}. To further increase the number of layers and support diverse style control, we integrate four popular LoRAs to generate transparent layers in various styles. We filter the data by eliminating non-transparent layers based on their alpha channel to ensure the visual harmony of synthesized data. As a result, we construct a high-quality, high-resolution transparent layer database consisting of over 1M transparent layers.

\begin{figure}[t]
\begin{minipage}{0.5\textwidth}
\centering
\includegraphics[width=\linewidth]{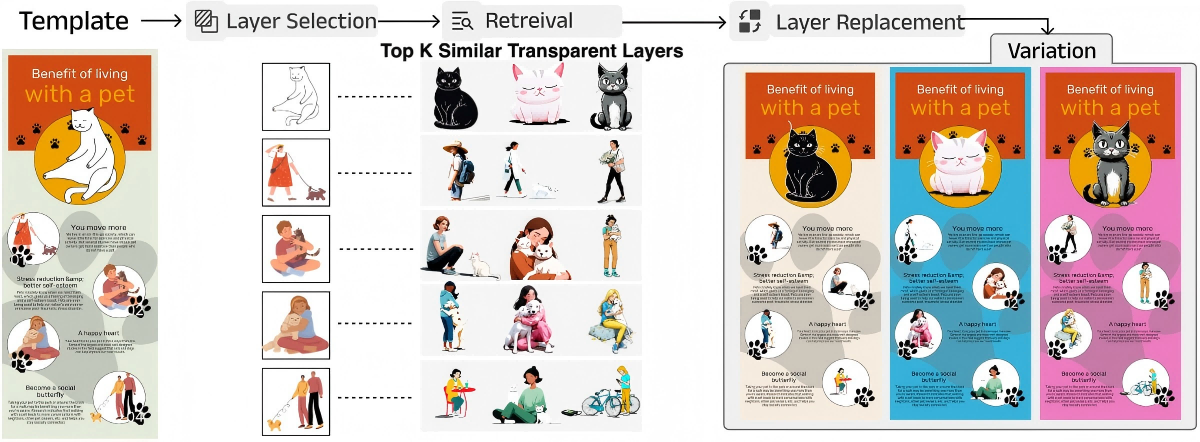}
\caption{ \footnotesize{\textbf{Overall pipeline of retrieval-augmented infographics generation}: we begin with a template infographic and perform layer selection to identify the most important visual layers. Then we conduct layer-wise retrieval from a constructed database of high-quality transparent layers and execute layer-wise replacement to generate various variants of the infographics.}}
\label{fig:infographic_data_engine}
\end{minipage}
\vspace{-3mm}
\end{figure}

\vspace{1mm}
\noindent\textbf{Layer-wise Retrieval-Augmented Infographics Generation.}
As shown in Figure~\ref{fig:infographic_data_engine}, we propose a retrieval-augmented pipeline to transform a given reference infographic into multiple new infographics with different visual elements. First, we use GPT-4o to identify the most important or visually dominant transparent object layers for each infographic separately. 
Next, we treat these selected layers as queries and search for the top 10 relevant layers with the highest CLIP similarity, discarding layers with significant discrepancies in aspect ratio, and use the remaining layers to replace the original layers.

To further diversify our synthesized infographics database, we randomly replace the solid-colored background layers with their counterparts from the retrieval database. Consequently, we generate over 650,000 high-quality multilingual infographics of various styles, collectively referred to as \textsc{Infographics-650K}. Figure~\ref{fig:infographic_data_engine} illustrates the entire pipeline of our layer-wise retrieval-augmented generation approach. Therefore, these generated high-quality multilingual infographics are of significant value to the community for studying the fundamental challenges of business content generation.

\begin{figure}[t]
\begin{minipage}[t]{1\linewidth}
\centering
\begin{subfigure}[b]{0.24\textwidth}
\includegraphics[width=\textwidth]{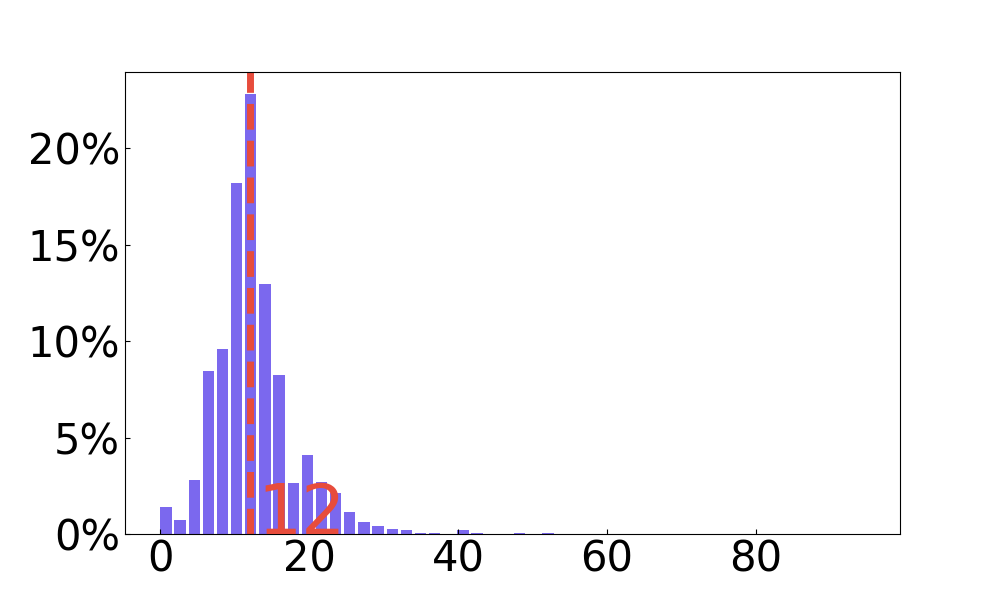}
\vspace{-5mm}
\caption{\scriptsize{\#text layers}}
\end{subfigure}
\begin{subfigure}[b]{0.24\textwidth}
{\includegraphics[width=\textwidth]{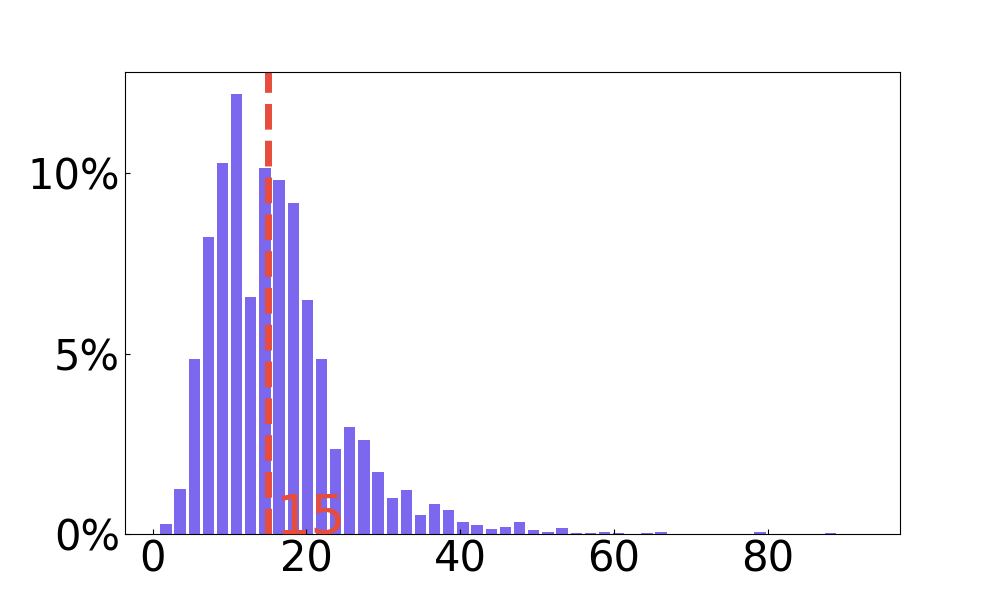}}
\vspace{-5mm}
\caption{\scriptsize{\#non-text layers}}
\end{subfigure}
\begin{subfigure}[b]{0.24\textwidth}
{\includegraphics[width=\textwidth]{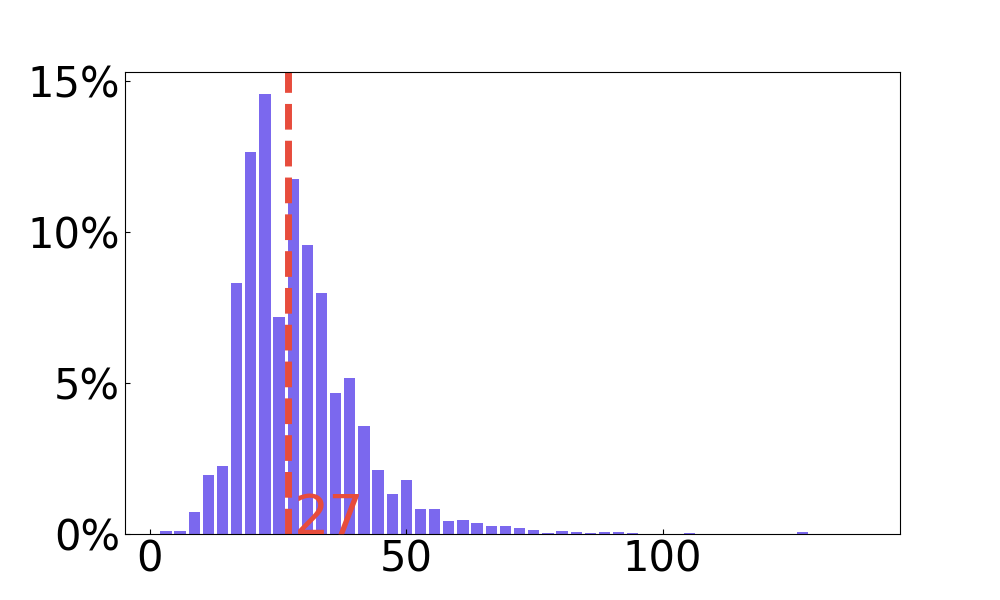}}
\vspace{-5mm}
\caption{\scriptsize{\#total layers}}
\end{subfigure}
\begin{subfigure}[b]{0.24\textwidth}
{\includegraphics[width=\textwidth]{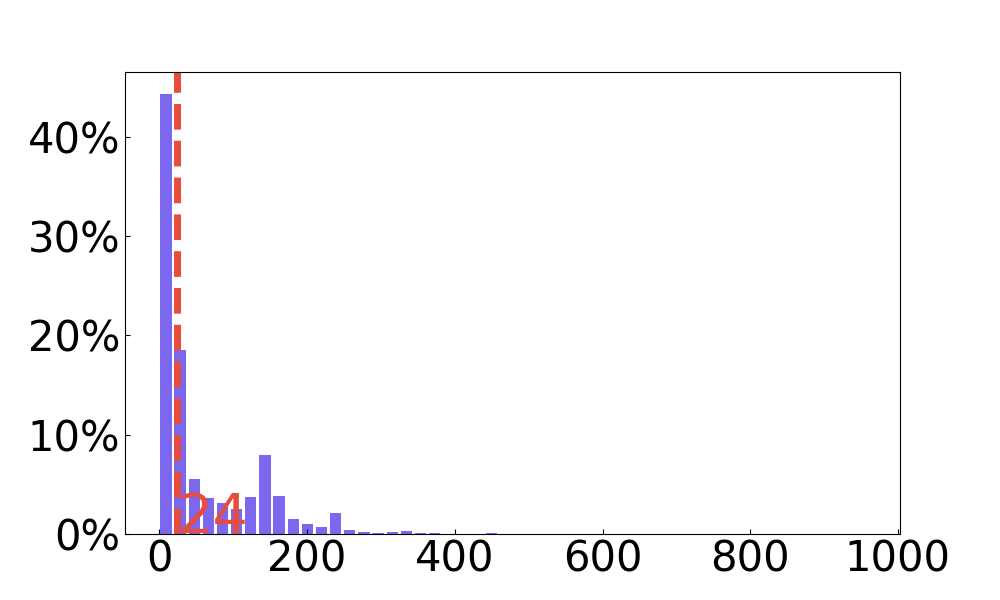}}
\vspace{-5mm}
\caption{\scriptsize{\#chars/layer}}
\end{subfigure}
\vspace{-2mm}
\caption{\footnotesize{\textbf{Distribution of the number of layers}, visual text layers, non-text layers, and the number of characters in the visual text layers. We mark the median values with red dashed lines.}}
\label{fig:data_histogram}
\end{minipage}
\vspace{-4mm}
\end{figure}

\vspace{1mm}
\noindent\textbf{Statistics of \textsc{Infographics-650K}}
The constructed dataset comprises over 650K high-quality infographic samples, evenly distributed across ten different languages: English, French, German, Spanish, Italian, Portuguese, Russian, Chinese, Japanese, and Korean. All the samples have a resolution of $2240\times896$ thanks to our collected original data and generated high-quality transparent layers, which is critical for improving training results. Figure~\ref{fig:data_histogram} provides main statistical information about the extracted layouts across the entire dataset, including the number of text layers, non-text layers, total layers, and characters. We observe that the infographic samples are highly layout-intensive, with an average of approximately 30 layers.

\vspace{1mm}
\noindent\textbf{Layer-wise Caption for \textsc{Infographics-650K}}
Given our access to both ground-truth element layers and visual textual data, we propose using the state-of-the-art LLaVA-1.6-34B model~\citep{liu2023llava,liu2023improved} to generate comprehensive layer-wise captions and global captions corresponding to the collected infographics. To generate the multilingual infographics samples, we utilize the latest GPT-4o to translate the original English text into other languages and then render it back to obtain captioned multilingual business content data.

\subsection{Layout-Guided Cross Attention}
\label{sec:approach_3}

\begin{figure*}[t]
\begin{minipage}[t]{1\linewidth}
\centering
\begin{subfigure}[b]{0.265\textwidth}
\includegraphics[width=1\linewidth]{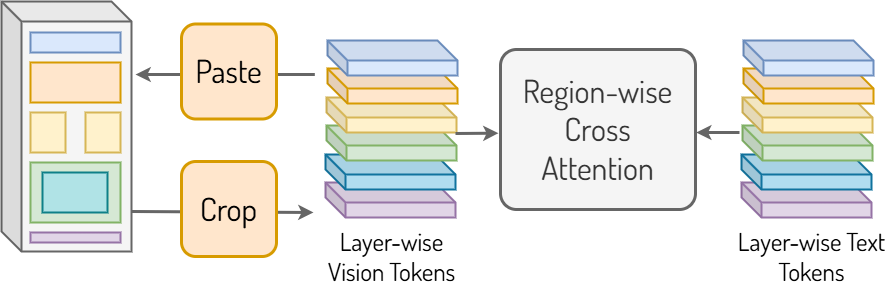}
\vspace{-3mm}
\caption{\scriptsize{Layout Guided Cross Attention}}
\end{subfigure}\hfill
\begin{subfigure}[b]{0.365\textwidth}
\includegraphics[width=\textwidth]{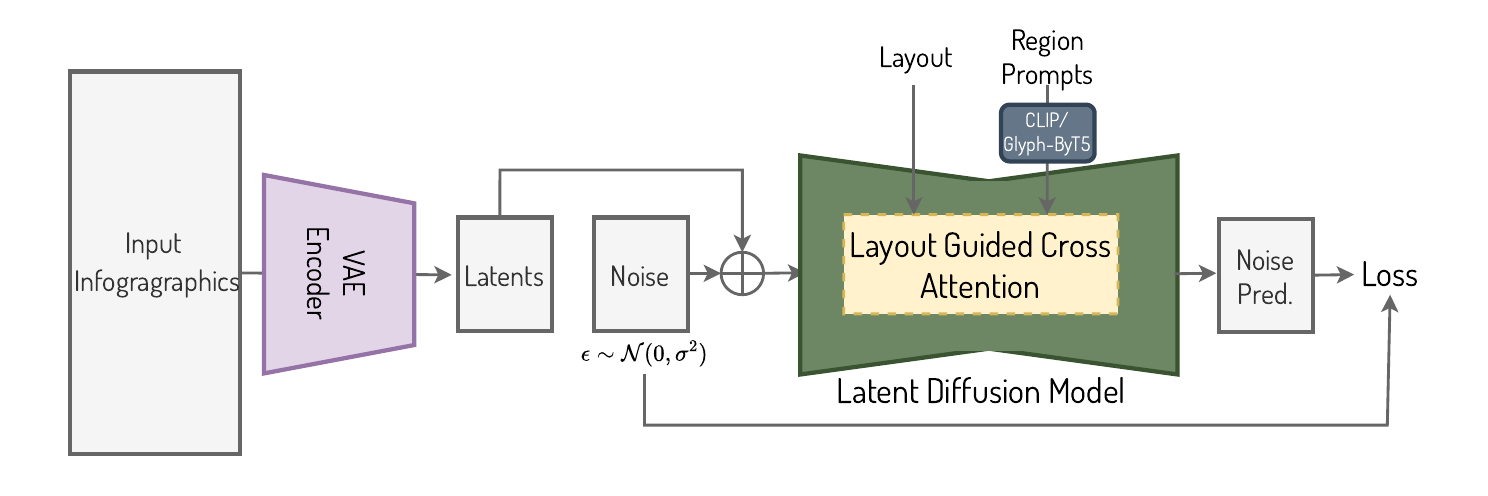}
\vspace{-3mm}
\caption{\scriptsize{Training pipeline of \ourapproach}}
\end{subfigure}\hfill
\begin{subfigure}[b]{0.365\textwidth}
{\includegraphics[width=\textwidth]{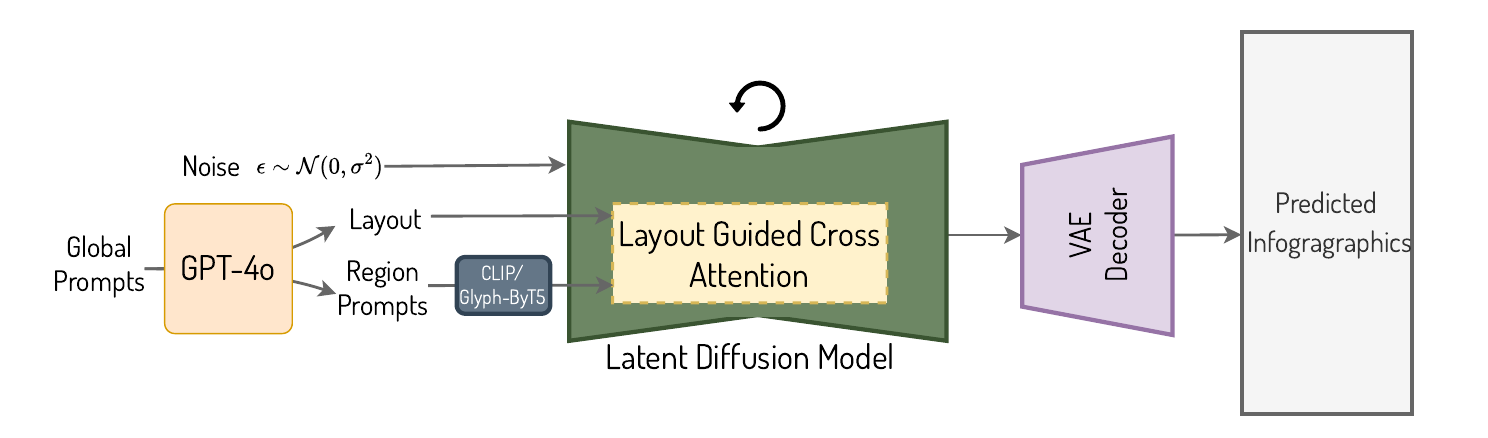}}
\vspace{-3mm}
\caption{\scriptsize{Evaluation pipeline of \ourapproach}}
\end{subfigure}
\caption{\textbf{Illustrating the Framework of Our Approach}: \footnotesize{(a) \textbf{Layout Guided Cross Attention}: we first crop the latent representations of the entire image into multiple groups of layer-wise visual tokens and extract the layer-wise text tokens according to the region prompts. Then, we perform region-wise cross attention over the corresponding layer-wise visual tokens and layer-wise text tokens to control the visual content generation within each region. (b) \textbf{\ourapproach training pipeline}: the inputs to the LDM include a noisy latent feature map, a layout, and the regional prompts, and the output of the LDM is a noise prediction. (c) \textbf{\ourapproach evaluation pipeline}: the inputs to the LDM include a noise map, a layout, and the regional prompts, and the outputs of the LDM (the combination of multiple iterative refinement) form a predicted clean latent that can be decoded into an infographic image that follows the given layout and regional prompts, which are provided by either users or an LLM like GPT-4o.}}
\label{fig:layout_guided_cross_attention}
\end{minipage}
\vspace{-5mm}
\end{figure*}

\vspace{1mm}
\noindent\textbf{Key Idea.} As illustrated in Figure~\ref{fig:layout_guided_cross_attention}, the key idea of the layout guided cross attention is to decompose the long context modeling task (10 times longer than for standard image generation tasks) into multiple shorter context modeling task by explicitly divide the vision tokens and text tokens into multiple groups according to the layout prior.
Instead of letting each vision token interact with nearly 1,000 text tokens, we only need to perform cross-attention for each pair of the grouped vision tokens and text tokens, where the grouped tokens belong to the same rectangle area according to the layout prior and the text tokens are based on the region-wise prompt with only within 100 text tokens. Therefore, the context length is decreased by nearly 10$\times$ with our layout-guided cross-attention design.

\vspace{1mm}
\noindent\textbf{Formulation.} We denote the ultra dense layout prior as \(\{\mathbf{b}^{\text{Non-Text}}_i, \text{prompt}_i\}_{i=1}^N \cup \{\mathbf{b}^{\text{Text}}_j, \text{prompt}_j\}_{j=1}^M\), where \(\mathbf{b}^{\text{Non-Text}}_i\) and \(\mathbf{b}^{\text{Text}}_j\) represent the bounding boxes for the non-text object layer and visual text layer, respectively. The terms \(\text{prompt}_i\) and \(\text{prompt}_j\) denote the prompts for each layer, while \(N\) and \(M\) are the numbers of non-text visual object layers and visual text layers, respectively. 

Then we crop the input latent representation $\mathbf{f}$ before sent into the $l$-th Cross-Attention block and according to:
\begin{align}
\label{eq-1}
\mathbf{x}_i &=\operatorname{Crop}(\mathbf{f}, \mathbf{b}^{\text{Non-Text}}_i),\\
\mathbf{x}_j &=\operatorname{Crop}(\mathbf{f}, \mathbf{b}^{\text{Text}}_j),\\
\mathbf{z}_i &=\operatorname{CrossAttention}(\mathbf{x}_i, \operatorname{CLIP}(\text{prompt}_i)),\\
\mathbf{z}_j &=\operatorname{CrossAttention}(\mathbf{x}_j, \operatorname{GlyphByT5}(\text{prompt}_j)),
\end{align}
where $\mathbf{x}_i$ and $\mathbf{x}_j$ represent the layer-wise vision tokens for the $i$-th non-text visual layer and $j$-th visual text layer, respectively. We use $\operatorname{CLIP}(\cdot)$ to denote the CLIP text encoder and $\operatorname{GlyphByT5}(\cdot)$ to denote the Glyph-ByT5 text encoder~\cite{liu2024glyph,liu2024glyphv2}.
Last, we integrate the updated region-wise representations into a combined whole image representation as follows:

\begin{align}
\label{eq-2}
\mathbf{z}=\sum_{i=1}^{N} \operatorname{Paste}(\mathbf{z}_i, \mathbf{b}^{\text{Non-Text}}_i) + \sum_{j=1}^{M} \operatorname{Paste}(\mathbf{z}_j, \mathbf{b}^{\text{Text}}_j),
\end{align}
where each $\operatorname{Paste}(\cdot)$ operation inserts the updated output from the previous cross-attention into zero-valued tensors of the same shape as the input feature map $\mathbf{f}$. We then combine them following their z-order to maintain their original spatial arrangement along the layer dimension. To ensure efficient implementation, we batch the cropped visual tokens and the region-wise text tokens across multiple layers.

\vspace{1mm}
\noindent\textbf{\textsc{BizGen}.}
We implement our business content generative model (\textsc{BizGen}) based on the latest Glyph-SDXL-v2~\citep{liu2024glyph,liu2024glyphv2} as it supports accurate multilingual visual text generation.
The key modification involves replacing the original region-wise multi-head cross-attention scheme with our layout-guided cross-attention scheme, which explicitly selects the visual tokens and text tokens for each element within a layout prior. This approach allows for more flexible control not only over text layers but also over visual object layers.

The model weights are initialized from the pretrained multilingual Glyph-SDXL~\citep{liu2024glyph,liu2024glyphv2}, to which we add a set of LoRA weights. We encode the prompts for all non-text layers using the CLIP text encoders and the prompts for the text layers with the Glyph-ByT5 text encoder. Since most layer-wise prompts exceed CLIP's default maximum length of 77, we divide the prompts into smaller chunks before sending them through the CLIP encoder and then concatenate the results. We also apply a lightweight mapper to align the text embeddings from Glyph-ByT5 with the original SDXL latent space.
We directly fine-tune the LoRA weights and the mapper weights with the constructed infographics dataset with the following hybrid loss:

{\footnotesize
\begin{multline}
\mathcal{L} = \mathbb{E}_{t, \mathbf{z}_0, \boldsymbol{\epsilon}} \Bigg[
    (1 - \mathsf{M}_{\text{text}}) \odot \left\| \boldsymbol{\epsilon} - \epsilon_\theta\left(\mathbf{z}_t, t, \{\mathbf{b}_i^{\textrm{Non-Text}}, \text{prompt}_i\}\right) \right\|_2^2 \\
    + \beta \cdot \mathsf{M}_{\text{text}} \odot \left\| \boldsymbol{\epsilon} - \epsilon_\theta\left(\mathbf{z}_t, t, \{\mathbf{b}_j^{\textrm{Text}}, \text{prompt}_j\}\right) \right\|_2^2
\Bigg] ,\label{eq-6}
\end{multline}
}
where $\mathsf{M}_\text{text}$ is a binary mask that encodes all visual text layer regions as 1 and 0 otherwise, and $\beta$ is a hyperparameter that controls the loss weight on the visual text regions.

\vspace{1mm}
\noindent\textbf{Layout Conditional CFG.}
To enable flexible control for each layer separately, we further propose a layout-conditional CFG strategy that applies a dense guidance scale map with different guidance values for positions from different layers. During training, we randomly drop the layer-wise captions and replace them with a null prompt. During inference, we first generate a set of binary masks $\{\mathsf{M}_i\}_{i=1}^{N+M}$ that record the positions belonging to the bounding box region, e.g., \(\{\mathbf{b}^{\text{Non-Text}}_i\}_{i=1}^N \cup \{\mathbf{b}^{\text{Text}}_j\}_{j=1}^M\) , in the given layer with 1, and 0 otherwise.
Then, we compute the dense guidance scale map $\mathsf{M}$ by adding the above rescaled binary masks following their layer order:
\begin{align}
\mathsf{M} &= \sum_{i=1}^{N+M} \gamma_i\mathsf{M}_i, \\
\hat{\mathbf{x}}_t^{\text{LCFG}} &= \mathbf{x}_t + \mathsf{M} \odot(\mathbf{x}^{\text{cond}}_t - \mathbf{x}_t),
\end{align}
where $\mathsf{M}$ is a 2D matrix of the same shape as the latent representation $\mathbf{x}_t$, $\gamma_i$ is the guidance weight used to control the CFG strength for the $i$-th layer guidance mask, and we perform the above process for certain timesteps $(0 \leq t \leq \alpha T)$. Here, we assume these layers already follow the correct order from index $i=1$ to index $i=(M+N)$ for simplicity. We empirically demonstrate that this layout-conditional CFG scheme is important for addressing artifacts and present the qualitative results in Section~\ref{sec:ablation}.
The ablation experiments on the effect of applying such layout-conditional CFG strategy across different timestep ranges are presented in the Appendix~\ref{sec:Appendix_lcfg_timestep}.
\section{Experiment}
\label{experiment}
First, we explain the training settings and construct a \ourbenchmark (Business Content Generation Benchmark) to evaluate business content generation results conditioned on various complex business-oriented layouts.
Second, we introduce a novel evaluation metric, the layer-wise generation success rate, to reflect whether the generation model can follow the ultra-dense layout and reliably generate each layer. We also report metrics for the entire infographics or slides from different perspectives, such as prompt following, aesthetics, and OCR.
Third, we present comparison results, including the aforementioned metrics and a thorough user study, with state-of-the-art image generation models, including SD3 and DALL$\cdot$E3, on our \ourbenchmark.
Finally, we conduct a comprehensive ablation study to examine the influence of different factors within our system.

\subsection{Training Settings}
We follow the default training settings of Glyph-ByT5~\citep{liu2024glyph,liu2024glyphv2} to fine-tune the SDXL using LoRA on our constructed \textsc{Infographics-650K}. We also construct a \textsc{Slides-500K} for multi-page slides generation. We detail the hyperparameter choices for different types of business content in the Appendix~\ref{sec:Appendix_train_set}. All these models are trained with 
$16\times$A100 GPUs for $280$ hours (w/o InfiniBand). Additionally, we follow Glyph-ByT5-v2~\citep{liu2024glyphv2} to opt for Glyph-SDXL+SPO~\citep{liang2024step} to improve visual aesthetics.
 
\subsection{Business Content Generation Benchmark}
We have constructed a \ourbenchmark benchmark to evaluate the business content generation task, containing detailed article-level prompts and ultra-dense layouts for approximately $\thicksim10\times100$ infographic, $\thicksim10\times90$ sets of slides (approximately $\thicksim10\times450$ pages) in ten languages. We cover a varying number of visual text layers and non-visual text layers of different generation difficulties, as detailed in Figure~\ref{fig:data_histogram}  and Appendix~\ref{sec:Appendix_lcfg_timestep}. All experiments are conducted on \ourbenchmark with a resolution of $2240\times896$ by default. During testing, all methods are provided with global prompts, while only methods supporting layout-guided cross attention are fed with layer-wise captions and layouts. 

\subsection{Evaluation Metrics}

\vspace{1mm}
\noindent\textbf{Global Image Quality.} We do not adopt the CLIP score as assessment, as our article-level global prompts are typically longer than 77 tokens. Instead, we utilize the latest GPT-4o to assess the generated infographics and slides from two perspectives: visual aesthetics and (global) prompt following. Additionally, we evaluate style consistency across different pages for slides. The GPT-4o prompt is provided in Appendix~\ref{sec:Appendix_global_qua}. 
To evaluate the visual text spelling accuracy, we employ character-level precision for character-based languages, specifically Chinese, Japanese, and Korean, while using word-level precision for the other seven alphabetic languages.

\begin{figure}[t]
\centering
\includegraphics[width=1\linewidth]{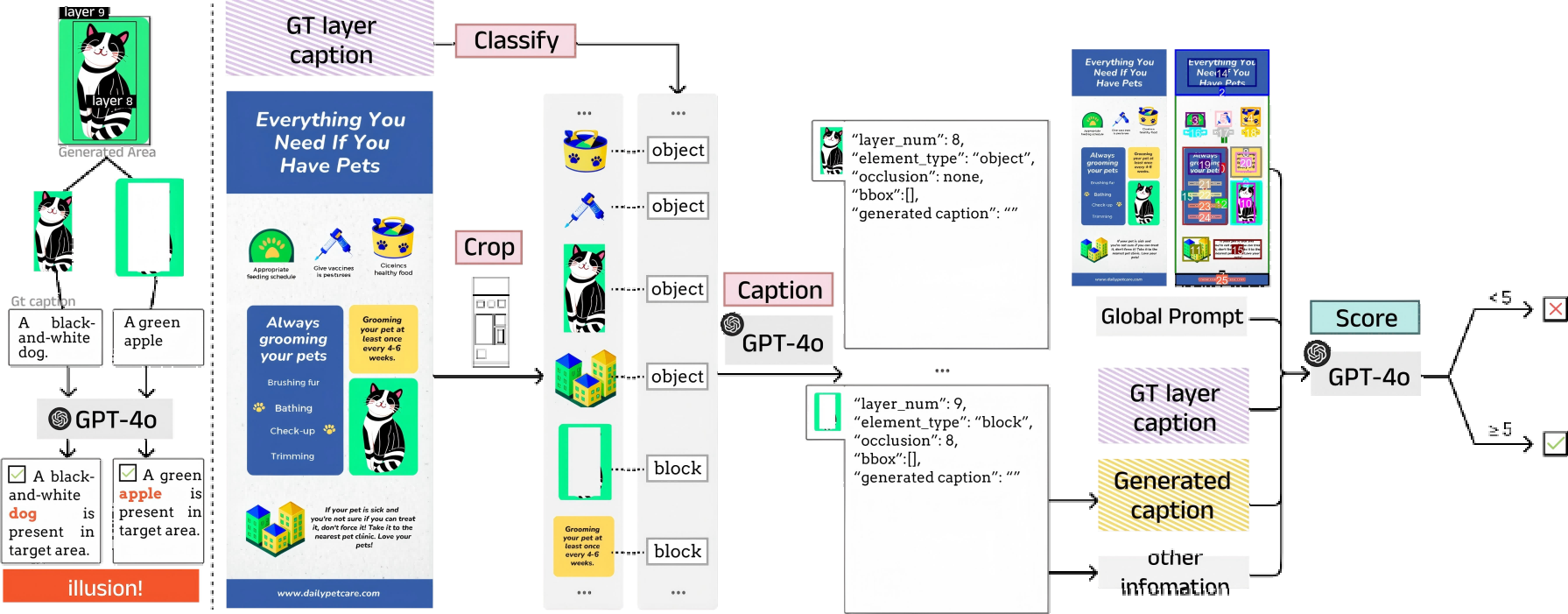}
\caption{\footnotesize{\textbf{Local region quality accessment pipeline}: this pipeline estimates the layer generation success rate based on the reference ultra-dense layout and region-wise prompts as conditions. We find that given only the GT layer caption and corresponding cropped out area of the generated image, GPT-4o tends to give conservative answers. For example, GPT-4o may mistaken the black-and-white cat for a black-and-white dog, or imagine there to be a green apple in the occluded area as shown in the left side of the dashed line. So we design the Classify-Caption-Score pipeline on the right to eliminate the illusion.}}
\label{fig:evaluation_metric}
\end{figure}

\vspace{1mm}
\noindent\textbf{Local Region Quality.} Considering that an infographic image typically consists of numerous elements, including visual text layers and non-text layers, it is crucial to assess whether the generated infographics adhere to the region-wise prompts and accurately generate each layer, particularly the quality of the non-text layers. To this end, we propose a novel layout-conditional region-wise evaluation scheme using GPT-4o to rate the local region quality of all non-text layers.

As shown in Figure~\ref{fig:evaluation_metric}, we first use GPT-4o to classify all layer prompts into two categories: ``block'', which typically refers to substrate layers, and ``object'', which usually describes specific objects. This classification is important because ``block'' layers are often occluded by visual text layers or other elements, so we focus on their color and texture. In contrast, for an ``object'' layer, we concentrate on its semantic content.
Next, we crop each layer from the generated image (with occlusions from higher layers masked out) and ask GPT-4o to provide a detailed description of the layers without supplying their ground-truth captions. Following this, we provide GPT-4o with the following information as outlined in Set-of-Mark~\citep{yang2023setofmark}: (i) the global caption, (ii) the generated image and its annotated replica, (iii) detailed information for each layer, including the layer caption, bounding box, occlusion relationships, and element type from the first step, and (iv) the layer descriptions from the second step, which serve as auxiliary information to avoid confusion.
We then ask GPT-4o to assign a layer-wise prompt alignment score ranging from 0 to 10 for each layer, with a score of 5 set as the threshold to calculate the LGSR (Layer Generation Success Rate) scores. The LGSR score is reported only when regional prompts and layouts are provided as conditions. The detailed prompt is included in the supplementary material.

\begin{figure}[!t]
\begin{minipage}[!t]{0.95\linewidth}
\begin{subfigure}[b]{0.19\textwidth}
\centering
\includegraphics[width=1\textwidth]{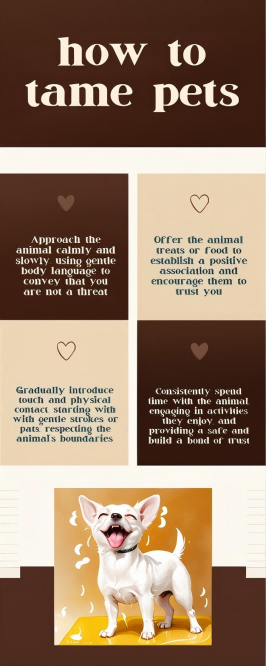}
\vspace{-3mm}
\caption*{\scriptsize{OCR: $\underline{97\%}$}}
\end{subfigure}
\hfill
\begin{subfigure}[b]{0.19\textwidth}
\centering
\includegraphics[width=1\textwidth]{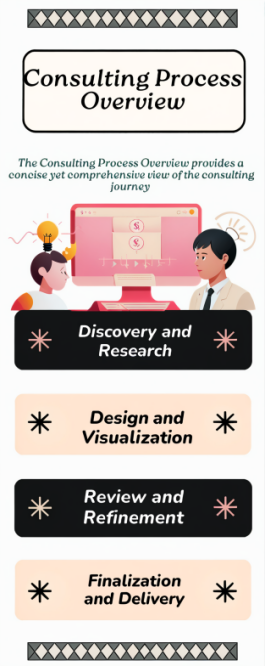}
\vspace{-3mm}
\caption*{\scriptsize{OCR: $\underline{100\%}$}}
\end{subfigure}
\hfill
\begin{subfigure}[b]{0.19\textwidth}
\centering
\includegraphics[width=1\textwidth]{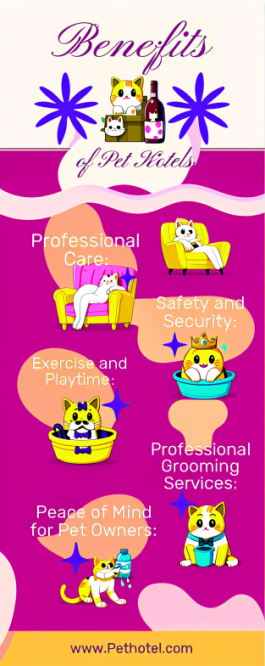}
\vspace{-3mm}
\caption*{\scriptsize{OCR: $\underline{100\%}$}}
\end{subfigure}
\hfill
\begin{subfigure}[b]{0.19\textwidth}
\centering
\includegraphics[width=1\textwidth]{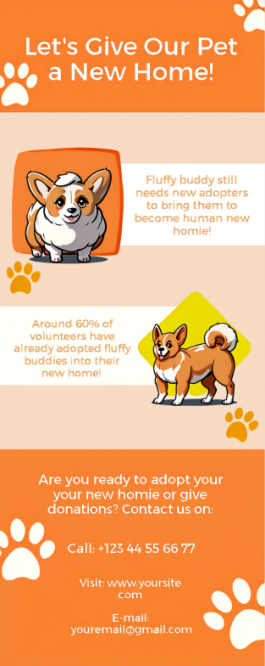}
\vspace{-3mm}
\caption*{\scriptsize{OCR: $\underline{98\%}$}}
\end{subfigure}
\begin{subfigure}[b]{0.19\textwidth}
\centering
\includegraphics[width=1\textwidth]{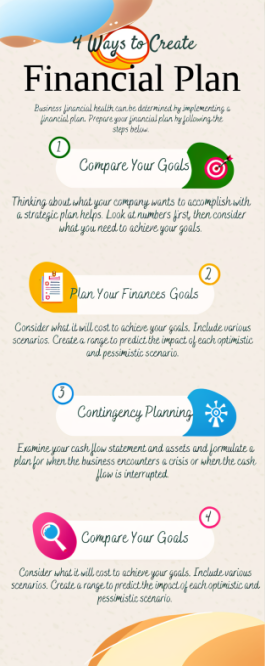}
\vspace{-3mm}
\caption*{\scriptsize{OCR: $\underline{98\%}$}}
\end{subfigure}
\hfill
\begin{subfigure}[b]{0.19\textwidth}
\centering
\includegraphics[width=1\textwidth]{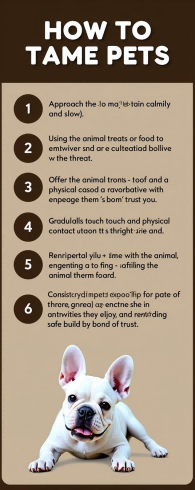}
\vspace{-3mm}
\caption*{\scriptsize{OCR: $\underline{37\%}$}}
\end{subfigure}
\hfill
\begin{subfigure}[b]{0.19\textwidth}
\centering
\includegraphics[width=1\textwidth]{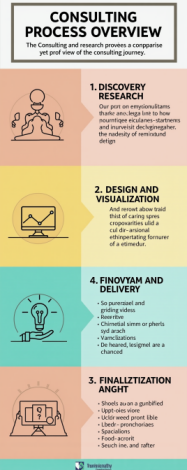}
\vspace{-3mm}
\caption*{\scriptsize{OCR: $\underline{52\%}$}}
\end{subfigure}
\hfill
\begin{subfigure}[b]{0.19\textwidth}
\centering
\includegraphics[width=1\textwidth]{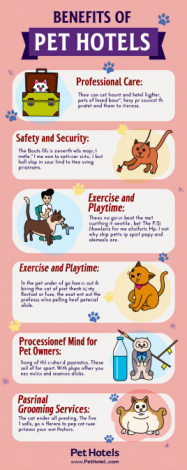}
\vspace{-3mm}
\caption*{\scriptsize{OCR: $\underline{15\%}$}}
\end{subfigure}
\hfill
\begin{subfigure}[b]{0.19\textwidth}
\centering
\includegraphics[width=1\textwidth]{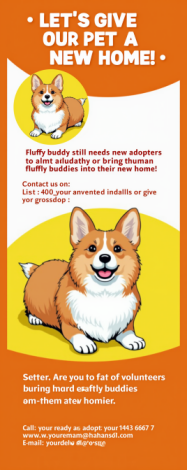}
\vspace{-3mm}
\caption*{\scriptsize{OCR: $\underline{65\%}$}}
\end{subfigure}
\hfill
\begin{subfigure}[b]{0.19\textwidth}
\centering
\includegraphics[width=1\textwidth]{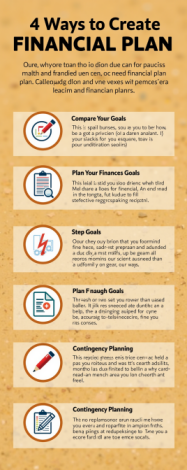}
\vspace{-3mm}
\caption*{\scriptsize{OCR: $\underline{19\%}$}}
\end{subfigure}
\hfill
\begin{subfigure}[b]{0.19\textwidth}
\centering
\includegraphics[width=1\textwidth]{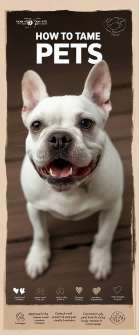}
\vspace{-3mm}
\caption*{\scriptsize{OCR: $\underline{21\%}$}}
\end{subfigure}
\hfill
\begin{subfigure}[b]{0.19\textwidth}
\centering
\includegraphics[width=1\textwidth]{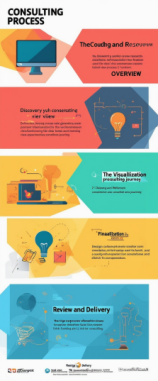}
\vspace{-3mm}
\caption*{\scriptsize{OCR: $\underline{14\%}$}}
\end{subfigure}
\hfill
\begin{subfigure}[b]{0.19\textwidth}
\centering
\includegraphics[width=1\textwidth]{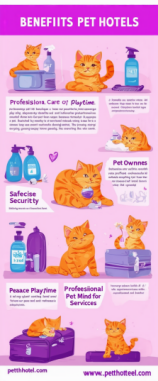}
\vspace{-3mm}
\caption*{\scriptsize{OCR: $\underline{10\%}$}}
\end{subfigure}
\hfill
\begin{subfigure}[b]{0.19\textwidth}
\centering
\includegraphics[width=1\textwidth]{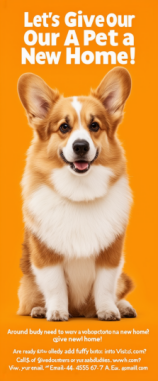}
\vspace{-3mm}
\caption*{\scriptsize{OCR: $\underline{38\%}$}}
\end{subfigure}
\hfill
\begin{subfigure}[b]{0.19\textwidth}
\centering
\includegraphics[width=1\textwidth]{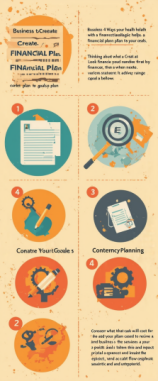}
\vspace{-3mm}
\caption*{\scriptsize{OCR: $\underline{43\%}$}}
\end{subfigure}
\hfill
\begin{subfigure}[b]{0.19\textwidth}
\centering
\includegraphics[width=1\textwidth]{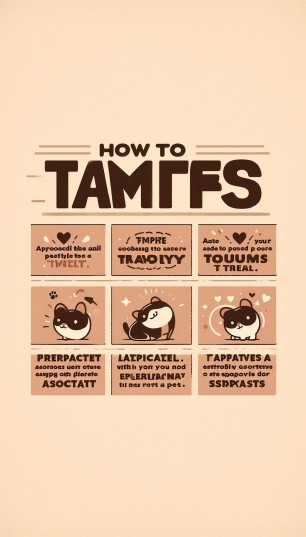}
\vspace{-3mm}
\caption*{\scriptsize{OCR: $\underline{16\%}$}}
\end{subfigure}
\hfill
\begin{subfigure}[b]{0.19\textwidth}
\centering
\includegraphics[width=1\textwidth]{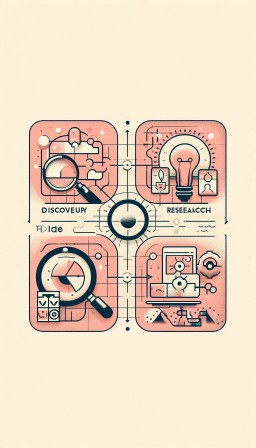}
\vspace{-3mm}
\caption*{\scriptsize{OCR: $\underline{0\%}$}}
\end{subfigure}
\hfill
\begin{subfigure}[b]{0.19\textwidth}
\centering
\includegraphics[width=1\textwidth]{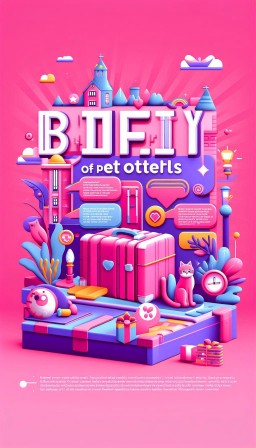}
\vspace{-3mm}
\caption*{\scriptsize{OCR: $\underline{40\%}$}}
\end{subfigure}
\hfill
\begin{subfigure}[b]{0.19\textwidth}
\centering
\includegraphics[width=1\textwidth]{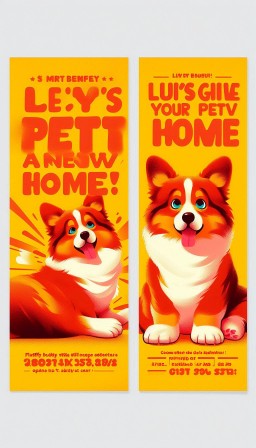}
\vspace{-3mm}
\caption*{\scriptsize{OCR: $\underline{14\%}$}}
\end{subfigure}
\hfill
\begin{subfigure}[b]{0.19\textwidth}
\centering
\includegraphics[width=1\textwidth]{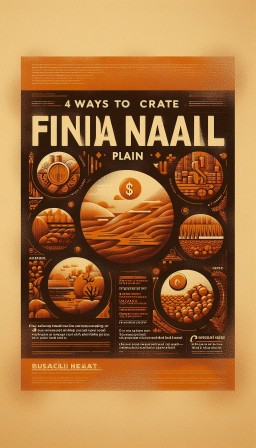}
\vspace{-3mm}
\caption*{\scriptsize{OCR: $\underline{27\%}$}}
\end{subfigure}
\end{minipage}
\caption{\footnotesize{
\textbf{Qualitative comparison results with SOTAs}. The 1st, 2nd, 3rd, and 4th rows correspond to the results generated with our \ourapproach, FLUX, SD3 Large, and \dalle.
}}
\label{fig:sota}
\vspace{-5mm}
\end{figure}

\subsection{Comparison with State-of-the-art}
We compare our \ourapproach with DALL$\cdot$E3, SD3 Large~\cite{Esser2024ScalingRF}, and the latest FLUX~\cite{flux} on our \ourbenchmark benchmark. The detailed comparison results are summarized in Table~\ref{tab:ocr_comparison_results} and Table~\ref{tab:numerical_quality_comparison_results}. Our approach significantly outperforms these strong models in terms of visual text spelling accuracy and prompt following, even though it is based solely on SDXL. We believe that combining our approach with the latest FLUX will yield even stronger results.

According to Table~\ref{tab:ocr_comparison_results}, we observe a substantial drop in spelling accuracy across all approaches when handling a greater number of visual layers. Nevertheless, we achieve over a $25\%\uparrow$ absolute performance gain over FLUX when generating infographics with more than 20 layers. Figure~\ref{fig:sota} illustrates some qualitative comparison results, showing that our \ourapproach achieves better quality than the other models. Additionally, since \ourapproach is based on SDXL, it significantly outmatches FLUX.1-dev in inference efficiency (40 seconds v.s. 68 seconds per image tested on a subset of our \ourbenchmark).
We also conducted a user study involving 10 infographic users with design backgrounds, asking them to choose their preferred results. The user study results are summarized in Figure~\ref{fig:user_study}. We found that \ourapproach was preferred in \(57.9\%\), \(83.4\%\), and \(84.3\%\) of cases compared to DALL$\cdot$E3, SD3 Large, and FLUX, respectively, in terms of typography quality. An interesting observation is that FLUX demonstrated unprecedented capability in generating relatively better infographics than both DALL$\cdot$E3 and SD3 Large. A reasonable hypothesis is that FLUX may have been trained on an infographic dataset.

\begin{table}[!t]
\begin{minipage}[t]{1\linewidth}
\centering
\tablestyle{1pt}{1.25}
\resizebox{1.0\linewidth}{!}
{
\begin{tabular}{l|cccc|cccc}
\multirow{2}{*}{Method} & \multicolumn{4}{c|}{Infographics Visual Text Spelling Precision ($\%$)}  & \multicolumn{4}{c}{Slides Visual Text Spelling Precision ($\%$)} \\\cline{2-9}
& $\le$10 layers & $\le$10-15 layers & $\le$15-20 layers & $\ge$20 layers & $\le$10 layers & $\le$10-20 layers & $\le$20-30 layers & $\ge$30 layers  \\
\shline
DALL$\cdot$E3 & ${16.42}$ & ${14.94}$ & ${21.64}$ & ${24.29}$ & ${24.55}$ & ${24.88}$ & ${30.59}$ & ${22.52}$ \\
SD3 Large& ${30.92}$ & ${32.38}$ & ${24.10}$ & ${20.52}$ & ${38.09}$ & ${42.96}$ & ${43.46}$ & ${31.39}$ \\
FLUX & ${41.33}$ & ${29.06}$ & ${28.47}$ & ${27.12}$ & ${62.30}$ & ${70.54}$ & ${67.57}$ & ${55.37}$ \\
Glyph-SDXL-v2 & ${78.18}$ & ${68.83}$ & ${51.73}$ & ${34.46}$  & ${80.18}$ & ${85.34}$ & ${79.89}$ & ${65.07}$ \\
\ourapproach & \pmb{${92.98}$} & \pmb{${84.25}$} & \pmb{${66.42}$} & \pmb{${55.48}$} & \pmb{${88.13}$} & \pmb{${88.41}$} & \pmb{${80.29}$} & \pmb{${66.81}$} \\
\end{tabular}
}
\vspace{-3mm}
\caption{
\footnotesize{\textbf{Comparison with SOTAs on visual text spelling precision} for infographic and slide generation.
}}
\label{tab:ocr_comparison_results}
\end{minipage}
\vspace{2mm}
\begin{minipage}[t]{\linewidth}
\centering
\tablestyle{15pt}{1.25}
\resizebox{1.0\linewidth}{!}
{
\begin{tabular}{l|cc|ccc}
\multirow{2}{*}{Method} & \multicolumn{2}{c|}{Infographics}  & \multicolumn{3}{c}{Slides} \\\cline{2-6}
& Aesthetics & Prompt Fo. & Aesthetics & Prompt Fo. & Style Cons. \\
\shline
DALL$\cdot$E3 & ${7.73}$ & ${5.32}$ & \pmb{${8.07}$} & ${6.80}$ & ${8.23}$ \\
SD3 Large& ${6.04}$ & ${3.88}$ & ${6.49}$ & ${5.29}$ & ${6.67}$  \\
FLUX & ${6.20}$ & ${4.11}$ & ${7.16}$ & ${6.32}$ & ${7.84}$  \\
\ourapproach & \pmb{${7.74}$} & \pmb{${8.09}$} & ${7.64}$ & \pmb{${7.61}$} & \pmb{${8.37}$} \\
\end{tabular}
}
\vspace{-3mm}
\caption{
\footnotesize{\textbf{Comparison with SOTAs on GPT-4o assessment scores} for infographic and slide generation.
}}
\label{tab:numerical_quality_comparison_results}
\end{minipage}
\vspace{-5mm}
\end{table}

\begin{figure}[!t]
\begin{minipage}[!t]{0.95\linewidth}
\begin{subfigure}[b]{0.32\textwidth}
\centering
\includegraphics[width=1\textwidth]{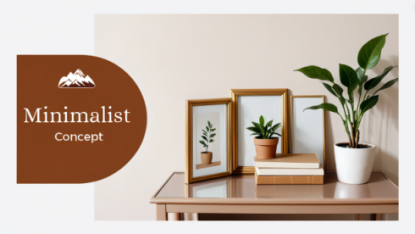}
\vspace{-3mm}
\caption*{\scriptsize{OCR: $\underline{100\%}$}}
\end{subfigure}
\hfill
\begin{subfigure}[b]{0.32\textwidth}
\centering
\includegraphics[width=1\textwidth]{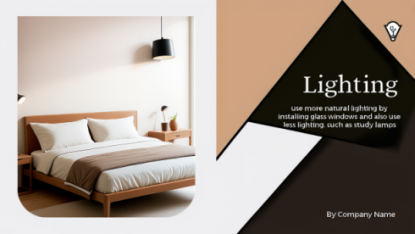}
\vspace{-3mm}
\caption*{\scriptsize{OCR: $\underline{100\%}$}}
\end{subfigure}
\hfill
\begin{subfigure}[b]{0.32\textwidth}
\centering
\includegraphics[width=1\textwidth]{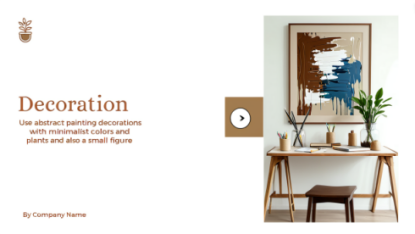}
\vspace{-3mm}
\caption*{\scriptsize{OCR: $\underline{100\%}$}}
\end{subfigure}
\begin{subfigure}[b]{0.32\textwidth}
\centering
\includegraphics[width=1\textwidth]{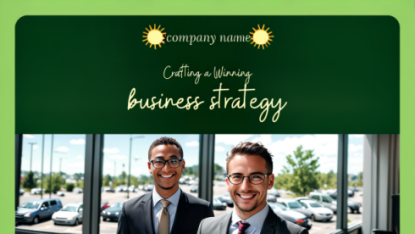}
\vspace{-3mm}
\caption*{\scriptsize{OCR: $\underline{100\%}$}}
\end{subfigure}
\hfill
\begin{subfigure}[b]{0.32\textwidth}
\centering
\includegraphics[width=1\textwidth]{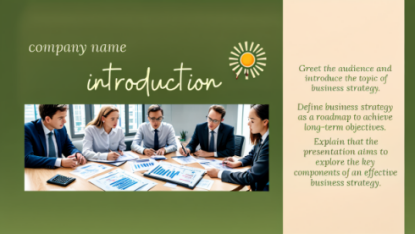}
\vspace{-3mm}
\caption*{\scriptsize{OCR: $\underline{100\%}$}}
\end{subfigure}
\hfill
\begin{subfigure}[b]{0.32\textwidth}
\centering
\includegraphics[width=1\textwidth]{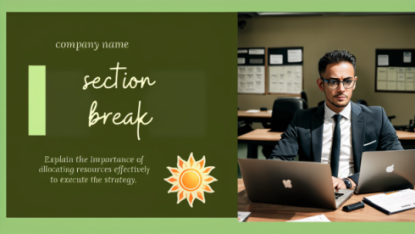}
\vspace{-3mm}
\caption*{\scriptsize{OCR: $\underline{93\%}$}}
\end{subfigure}
\hfill
\begin{subfigure}[b]{0.32\textwidth}
\centering
\includegraphics[width=1\textwidth]{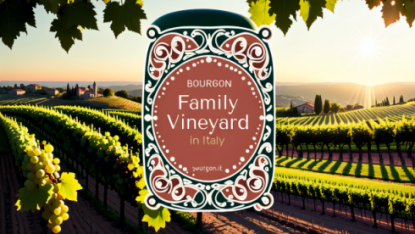}
\vspace{-3mm}
\caption*{\scriptsize{OCR: $\underline{83\%}$}}
\end{subfigure}
\hfill
\begin{subfigure}[b]{0.32\textwidth}
\centering
\includegraphics[width=1\textwidth]{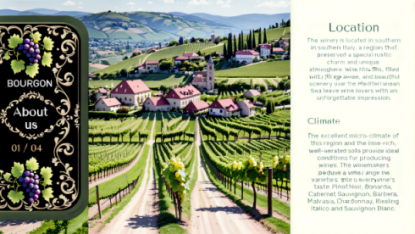}
\vspace{-3mm}
\caption*{\scriptsize{OCR: $\underline{81\%}$}}
\end{subfigure}
\hfill
\begin{subfigure}[b]{0.32\textwidth}
\centering
\includegraphics[width=1\textwidth]{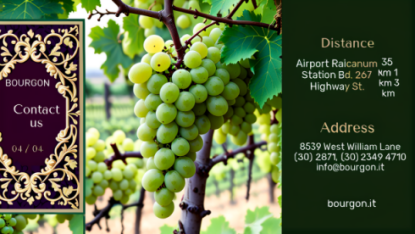}
\vspace{-3mm}
\caption*{\scriptsize{OCR: $\underline{91\%}$}}
\end{subfigure}
\end{minipage}
\vspace{-2mm}
\caption{\footnotesize{
\textbf{Qualitative results of generated slides}. Each row displays three selected pages of the generated slides.
}}
\label{fig:slides}
\vspace{-2mm}
\end{figure}

\subsection{Ablation Study}
\label{sec:ablation}
\vspace{1mm}
\noindent \textbf{Scaling Infographics Training Data} We study the effect of scaling the infographics dataset size as illustrated in Tasble~\ref{tab:ablation_dataset_scale}. Here baseline means directly using our layer-guided cross attention in inference without tuning the model. We observe significantly improved performance with more data across all metrics.  For the trade-off between efficiency and performance, we choose \textsc{Infographics-65K} to conduct all other ablation experiments by default.

\vspace{1mm}
\noindent \textbf{Layer-Guided Cross Attention}
Table~\ref{tab:ablation_layout_guided_ca} ablates the benefits of applying layout-guided cross attention compared to the original region-wise cross attention used in Glyph-ByT5~\cite{liu2024glyph,liu2024glyphv2}. According to the results, we find that our layout-guided cross attention primarily improves visual aesthetics and prompt-following capability, while also ensuring better visual text spelling accuracy when the number of layers are less than $20$.

\vspace{1mm}
\noindent \textbf{High Resolution Matters}
Table~\ref{tab:ablation_resolution} summarizes the results of choosing different resolutions, including \(1120 \times 448\), \(1600 \times 640\), and \(2240 \times 896\), which correspond to approximately \(0.5\times\), \(1\times\), and \(2\times\) the size of the universal \(1024 \times 1024\) image resolution. According to the results, we find that high resolution is critical for improving visual text spelling precision, as there can be many small-sized visual texts in an infographic. An interesting observation is that fine-tuning with lower resolution can actually hurt performance compared to the baseline reported in Table~\ref{tab:ablation_dataset_scale}.

\begin{figure}[t]
\centering
\includegraphics[width=.85\linewidth]{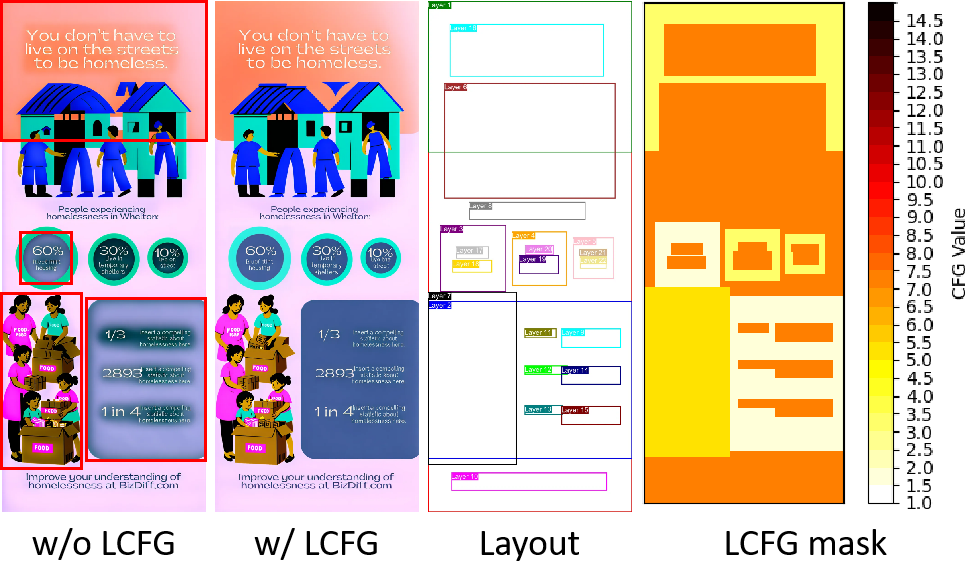}
\vspace{-2mm}
\caption{\footnotesize{\textbf{LCFG removes layer artifacts}: the image on the left is generated with a global CFG of 7, with some local artifacts marked by red boxes. By adjusting the CFG values of different layers, these artifacts are eliminated in the second image.}}
\label{fig:LCFG}
\vspace{-5mm}
\end{figure}

\vspace{1mm}
\noindent \textbf{Layout Conditional Classifier-free Guidance}
Figure~\ref{fig:LCFG} illustrates the benefits of applying our LCFG to refine the artifacts in specific layers. For example, we observe that choosing the CFG values as 3, 1.5, 1.5, 3, 3 and 5 for layer 1, layer 2, layer 3, layer 4, layer 5, and layer 7 helps remove the observable artifacts. We also visualize the final layout-conditional CFG mask on the right. We set the timestep-related hyperparameter $\alpha$ as $0.5$ in this case.

\vspace{1mm}
\noindent \textbf{More Experiments}
We provide additional experiments and discussion on the support for multilingual infographic and slide generation, the application of our \ourinfo to train multi-layer transparent infographic generation models, and more ablation experiments in the supplementary material.

\begin{table}[!t]
\begin{minipage}[t]{1\linewidth}
\centering
\tablestyle{1pt}{1.25}
\resizebox{1.0\linewidth}{!}
{
\begin{tabular}{l|cccc|cccc|cc}
\multirow{2}{*}{Method} & \multicolumn{4}{c|}{Text Spelling Precision ($\%$)}  & \multicolumn{4}{c|}{Non-text Layer Precision/LGSR ($\%$)} & \multicolumn{2}{c}{GTP-4o Score (1-10)}\\\cline{2-11}
& $\le$10 L. & $\le$10-15 L. & $\le$15-20 L. & $\ge$20 L. & $\le$10 L. & $\le$10-20 L. & $\le$20-30 L. & $\ge$30 L. & Aesthetics & Prompt Fo. \\
\shline
Baseline & ${78.18}$ & ${68.83}$ & ${51.73}$ & ${34.46}$ & ${61.38}$ & ${55.85}$ & ${44.33}$ & ${48.01}$ & ${5.98}$ & ${5.39}$\\
+ \textsc{Infographics-6.5K} & ${83.69}$ & ${75.68}$ & ${55.30}$ & ${46.96}$ & ${90.00}$ & ${84.10}$ & ${78.47}$ & ${83.32}$ & ${7.03}$ & ${7.42}$\\
+ \textsc{Infographics-65K} & ${86.47}$ & ${79.04}$ & ${58.48}$ & ${48.24}$ & ${94.71}$ & ${89.38}$ & ${82.33}$ & ${86.07}$ & ${7.20}$ & ${7.78}$\\
+ \ourinfo & \pmb{${92.98}$} & \pmb{${84.25}$} & \pmb{${66.42}$} & \pmb{${55.48}$} & \pmb{${95.12}$} & \pmb{${90.39}$} & \pmb{${88.34}$} & \pmb{${91.54}$} & \pmb{${7.74}$} & \pmb{${8.09}$}\\
\end{tabular}
}
\vspace{-3mm}
\caption{
\footnotesize{\textbf{Scaling the infographics dataset improves performance.}
}}
\label{tab:ablation_dataset_scale}
\vspace{2mm}
\end{minipage}
\begin{minipage}[t]{1\linewidth}
\centering
\tablestyle{10pt}{1.25}
\resizebox{1.0\linewidth}{!}
{
\begin{tabular}{l|cccc|cc}
\multirow{2}{*}{Method} & \multicolumn{4}{c|}{Text Spelling Precision ($\%$)} & \multicolumn{2}{c}{GTP-4o Score (1-10)}\\\cline{2-7}
& $\le$10 layers & $\le$10-15 layers & $\le$15-20 layers & $\ge$20 layers & Aesthetics & Prompt Fo. \\
\shline
Region-wise CA & ${84.91}$ & ${77.28}$ & ${57.18}$ & \pmb{${50.56}$} & ${6.82}$ & ${6.99}$\\
Layout-Guided CA  & \pmb{${86.47}$} & \pmb{${79.04}$} & \pmb{${58.48}$} & ${48.24}$  & \pmb{${7.20}$} & \pmb{${7.78}$}\\
\end{tabular}
}
\vspace{-3mm}
\caption{
\footnotesize{\textbf{Layout-guided cross attention improves performance.}
}}
\label{tab:ablation_layout_guided_ca}
\vspace{2mm}
\end{minipage}
\begin{minipage}[t]{1\linewidth}
\centering
\tablestyle{1pt}{1.25}
\resizebox{1.0\linewidth}{!}
{
\begin{tabular}{l|cccc|cccc|cc}
\multirow{2}{*}{Resolution} & \multicolumn{4}{c|}{Text Spelling Precision ($\%$)}  & \multicolumn{4}{c|}{Non-text Layer Precision/LGSR ($\%$)} & \multicolumn{2}{c}{GTP-4o Score (1-10)}\\\cline{2-11}
& $\le$10 L. & $\le$10-15 L. & $\le$15-20 L. & $\ge$20 L. & $\le$10 L. & $\le$10-20 L. & $\le$20-30 L. & $\ge$30 L. & Aesthetics & Prompt Fo. \\
\shline
$1120\times448$ & ${44.68}$ & ${32.47}$ & ${30.55}$ & ${24.39}$ & ${86.46}$ & ${83.21}$ & ${75.17}$ & ${77.11}$ & ${6.56}$ & ${6.73}$\\
$1600\times640$ & ${76.07}$ & ${53.47}$ & ${38.26}$ & ${29.64}$ & ${87.50}$ & ${86.37}$ & ${80.56}$ & ${76.62}$ & ${6.51}$ & ${7.05}$\\
$2240\times896$ & \pmb{${86.47}$} & \pmb{${79.04}$} & \pmb{${58.48}$} & \pmb{${48.24}$} & \pmb{${94.71}$} & \pmb{${89.38}$} & \pmb{${82.33}$} & \pmb{${86.07}$} & \pmb{${7.20}$} & \pmb{${7.78}$}\\
\end{tabular}
}
\vspace{-3mm}
\caption{
\footnotesize{\textbf{High-resolution is critical for the improvements.}
}}
\label{tab:ablation_resolution}
\end{minipage}
\vspace{-3mm}
\end{table}

\section{Conclusion}
\label{sec:conclusion}
This paper presents a novel generation framework, \ourapproach, to tackle the challenging task of business content generation with ultra-dense layouts and region-wise prompts. By leveraging our scalable infographics generation engine and the layout-guided cross-attention scheme, our system addresses two fundamental challenges: data scarcity and extremely long context length. We empirically demonstrate that our approach can generate high-quality multilingual and multi-style infographics and slides. Additionally, we show that \ourapproach outperforms DALL$\cdot$E3, SD3, and FLUX by a significant margin on our \ourbenchmark benchmark.

{
    \small
    \bibliographystyle{ieeenat_fullname}
    \bibliography{main}
}

\newpage
\clearpage

\maketitlesupplementary
\setcounter{table}{5}
\setcounter{figure}{12}

\appendix
\renewcommand{\thesection}{\Alph{section}} 
\renewcommand{\thesubsection}{\Alph{section}.\arabic{subsection}} 

\section{Dataset Statistics of \ourslide} 
\label{sec:Appendix_data_sta}
We present statistics on the number of text layers, non-text layers, total layers, characters, and pages across the entire \ourslide dataset in Figure~\ref{fig:slide_data_histogram}. Each set consists of multiple slide pages with the same style. Compared to Figure~\ref{fig:data_histogram} in the main paper, we find that a single-page slide typically has a sparser layout than an infographic; however, the total number of layers accumulated across a set of slides can far exceed that of an infographic.

\section{Training Settings}
\label{sec:Appendix_train_set}
Table~\ref{table:diffusion_hparam} details the training hyperparameter choices for our infographic generation model and slides generation model, respectively. We use a smaller glyph loss weight for infographics to avoid artifacts, such as shadows, after extensive tuning on data with denser layouts.

\section{Layer-wise Retrieval-Augmented Infographics Data Engine}
\label{sec:Appendix_data_eng}
\noindent\textbf{Dominant Layer Selection.}
We present the GPT-4o prompt used to identify the main layers in our originally collected infographic data. These layers will be replaced by generated high-quality transparent layers in our data engine.
\begin{mainlayerprompt}
You are an experienced visual designer. You are given 1+N images. The first one is the whole image of an infographic. Assume the infographic has N layers. The following N images are different layers of the first image. The caption of the layer images are also given. You are required to classify the layer images into two categories: "Main element" or "Others". "Main element" means the layer should contain at least one specific object, the object(s) should be intact in the layer, and the object(s) should be aesthetically beautiful and the layer shouldn't be too small. Others include solid-colored blocks, decoration layers and any other layers that do not meet the requirements of "Main element". Your answer should be in list format, containing only "Others" or "Main element" and nothing else. The length of the returned list should be equal to N.
\end{mainlayerprompt}
\vspace{1mm}
\noindent\textbf{Filtering of the Generated Transparent Layers.}
Figure~\ref{fig:transparent_layer_example} presents examples of both good and bad layers generated with our scalable data engine. The bad layers are filtered out because: (i) objects fill the entire canvas (1st and 2nd in the second row), or (ii) the image has a solid-colored background with a non-zero alpha channel (3rd, 4th, 5th, and 6th in the second row). The remaining layers after the filtering are transparent and of high aesthetic quality.

\begin{figure}[!t]
\vspace{-5mm}
\centering
\includegraphics[width=0.45\textwidth]{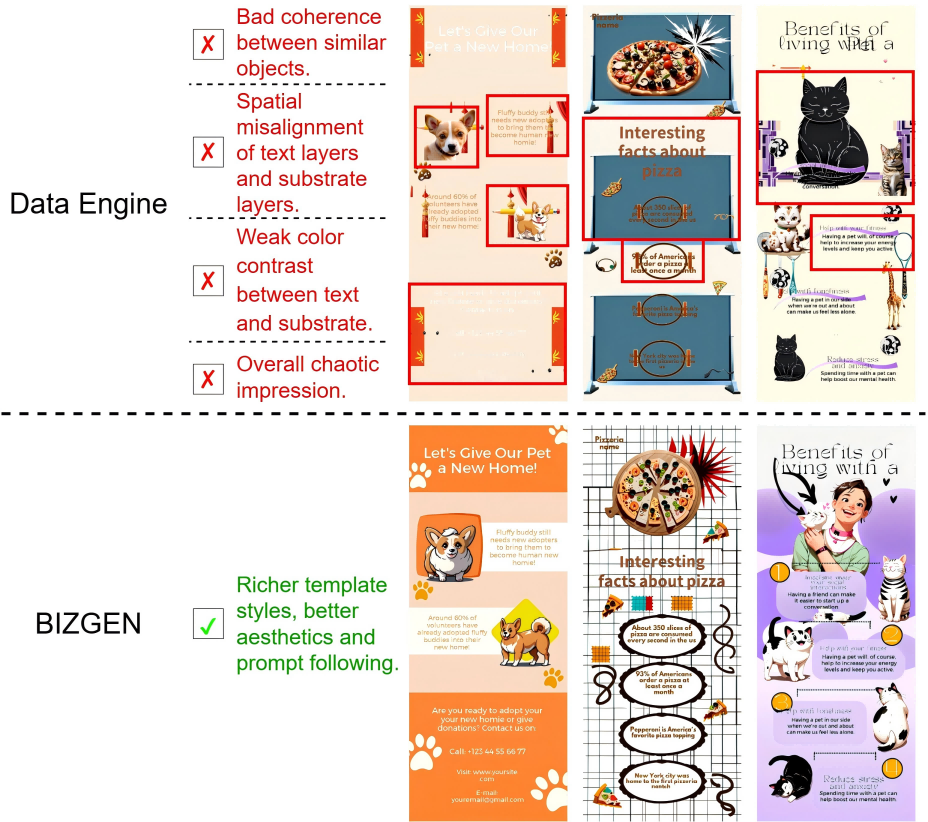}
\vspace{-3mm}
\caption{\footnotesize{
\textbf{Comparison of assembling through data engine against \ourapproach }}}
\label{fig:assembled_against_generated}
\vspace{-6mm}
\end{figure}

\vspace{1mm}
\noindent\textbf{Discussion on directly assembling through data engine v.s. generating through \ourapproach.}Since we already construct the automatic infographic data engine, it seems like we can directly assemble infographics through layer-wise retrieval and text rendering, which ensures 100\% OCR accuracy. However, to generate infographics with high diversity, such approach requires retrieving all the non-text layers (different from just doing augmentation where we only select dominant layers), and assembling them. The assembled results lag behind \ourapproach in terms of limited template styles, poorer aesthetics, and weaker prompt following. We visualize the quantitative comparison results in Table~\ref{tab:assembled_against_generated} and qualitative comparison results in Figure~\ref{fig:assembled_against_generated}. 

We argue that the poor performance of the assembled infographics lies in the complex relevance between text and non-text layers within an image. \underline{First}, the explanatory text is usually spatially adjacent to the corresponding non-text elements, enhancing the mapping between visuals and semantics.
\underline{Second}, rather than relying on post-hoc combinations of predefined transparent layers from a database, \ourapproach jointly generates both elements end-to-end within a unified generative model ensures intrinsic contextual alignment.
\underline{Third}, texts can have spatially overlapped substrate layers, requiring spatial alignment and contrast in colors, as illustrated in Figure~\ref{fig:assembled_against_generated}.

\begin{table}
\begin{minipage}[t]{\linewidth}
\centering
\tablestyle{15pt}{1.25}
\resizebox{1.0\linewidth}{!}
{
\begin{tabular}{l|cc|ccc}
\multirow{2}{*}{Method} & \multicolumn{2}{c|}{Infographics}  & \multicolumn{3}{c}{Slides} \\\cline{2-6}
& Aesthetics & Prompt Fo. & Aesthetics & Prompt Fo. & Style Cons. \\
\shline
DataEngine & ${6.33}$ & ${6.50}$ & ${7.00}$ & ${6.87}$ & ${7.73}$ \\
\ourapproach & \pmb{${7.74}$} & \pmb{${8.09}$} & \pmb{${7.64}$} & \pmb{${7.61}$} & \pmb{${8.37}$} \\
\end{tabular}
}
\vspace{-3mm}
\caption{
\footnotesize{Comparison with the assembled results produced by data engine.}
}
\label{tab:assembled_against_generated}
\end{minipage}
\vspace{-6mm}
\end{table}

\begin{table}[t]
\footnotesize
\centering
\begin{minipage}[t]{1\linewidth}
\tablestyle{10pt}{1.15}
\resizebox{1.0\linewidth}{!}{
\begin{tabular}{l|c|c}
Hyperparameter     &        \ourapproach-Infographics &                \ourapproach-Slides \\
\shline
Backbone  &    Glyph-SDXL &               Glyph-SDXL \\
UNet Learning-rate &          1.00E-04 &            1.00E-04  \\
Batch Size &                128 &                128 \\
Epochs &                  5 &                  10  \\
Weight Decay &                0.01 &                0.01  \\
Text-Encoder Dropout &                     0.1 &                   0.1 \\
Gradient Clipping &                      1.0 &                   1.0 \\
Resolution &                          [2240,896] &                   [864,1536] \\ 
UNet LoRA Rank &                       128 &                   128 \\ 
Text Feature Length &                 2048 &                   2048 \\ 
Glyph Loss Weight &                         1 &                   5 \\
Dataset &                 \ourinfo &                   \ourslide \\
\end{tabular}
}
\vspace{-3mm}
\caption{\footnotesize \ourapproach \textbf{Training hyper-parameter choices.}}
\label{table:diffusion_hparam}
\end{minipage}
\vspace{2mm}
\end{table}

\begin{figure}[!t]
\vspace{-5mm}
\centering
\includegraphics[width=0.45\textwidth]{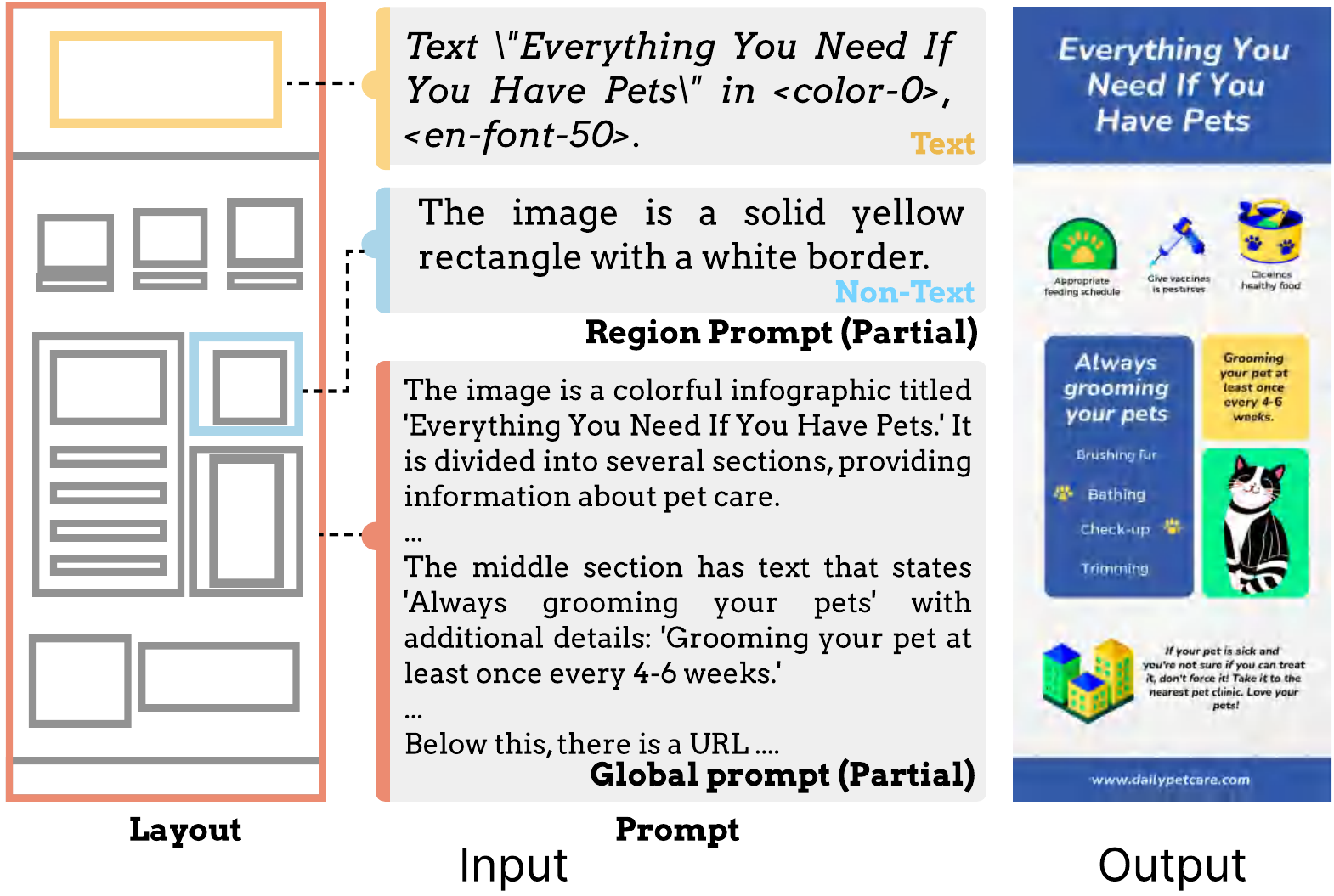}
\vspace{-3mm}
\caption{\footnotesize{
\textbf{Input and output examples of \ourapproach }}}
\label{fig:input_output}
\vspace{-3mm}
\end{figure}

\begin{infoqualityprompt}
You are an autonomous AI Assistant who aids designers by providing insightful, objective, and constructive critiques of graphic design projects. Your goals are: Deliver comprehensive and unbiased evaluations of graphic designs based on established design principles and industry standards. Maintain a consistent and high standard of critique. 

You will be given an image of graphic design (poster, infographic, slide, etc.) and the caption that describes the content in the image. Please abide by the following rules: Strive to score as objectively as possible. Grade seriously. A flawless design can earn 10 points, a mediocre design can only earn 7 points, a design with obvious shortcomings can only earn 4 points, and a very poor design can only earn 1-2 points. Keep your reasoning concise when rating, and describe it as briefly as possible. If the output is too long, it will be truncated. Only respond in JSON format, no other information. 

Grading criteria: 

Aesthetics (1-10): The graphic design should be aesthetically beautiful, with a pleasing color palette, harmonious composition, and appropriate use of space. There should not be artifacts, visual noise, blurriness, wrong words, incomplete text, or any other issues that detract from the overall visual experience. A score of 10 indicates a design that is visually stunning, while a score of 1 indicates a design that is visually unappealing. 

Prompt Following (1-10): The image should be well aligned with the prompt given. Every object mentioned in the prompt should be present in the image. The object's attribute such as color, size, and position should also match the description in the prompt if assigned. If the prompt mentions text in the image, it should be present and legible, and the texts should also match the logical flow of the prompt. The overall style of the image should also align with the tone of the prompt. A score of 10 indicates perfect alignment, while a score of 1 indicates a complete mismatch. 

Your answer should be a dictionary with two keys: 'Aesthetics' and 'Prompt Following', each containing two keys: 'score' and 'reason'. The 'score' should be an integer between 1 and 10, and the 'reason' should be a string explaining the score.
\end{infoqualityprompt}

\section{Model Architecture}
\label{sec:Appendix_model_architecture}

\vspace{1mm}
Figure~\ref{fig:input_output} shows an example of the imput layout and regional prompts and the output infographic of \ourapproach. We discuss the architecture difference with Glyph-SDXL as follows: While we adopt Glyph-SDXL as our backbone, the key difference lies in the implementation of the region-wise cross-attention scheme. The original GlyphByT5 computes full attention between all text tokens and visual tokens, using a pre-computed attention mask to determine what to generate and where. To deal with far more text and visual tokens in infographic setting, our \ourapproach explicitly crops the visual tokens based on the layout and partitions the text tokens within each region. Attention is then computed only between each set of cropped visual tokens and the corresponding regional text tokens, which is critical for higher efficiency. Additionally, we introduce a layout-conditional CFG scheme during inference to further enhance the results.

\section{Global Quality Assurance Prompt}
\label{sec:Appendix_global_qua}
\begin{slidequalityprompt}
You are an autonomous AI Assistant who aids designers by providing insightful, objective, and constructive critiques of graphic design projects. Your goals are: Deliver comprehensive and unbiased evaluations of graphic designs based on established design principles and industry standards. Maintain a consistent and high standard of critique. 

You will be given a set of slides and the caption that describes the content in each of the single page. Your score should be based on the overall quality of the slides, not a single page. Please abide by the following rules: Strive to score as objectively as possible. Grade seriously. A flawless design can earn 10 points, a mediocre design can only earn 7 points, a design with obvious shortcomings can only earn 4 points, and a very poor design can only earn 1-2 points. Keep your reasoning concise when rating, and describe it as briefly as possible. If the output is too long, it will be truncated. Only respond in JSON format, no other information. 

Grading criteria: 

Aesthetics (1-10): The slides should be aesthetically beautiful, with a pleasing color palette, harmonious composition, and appropriate use of space. There should not be artifacts, visual noise, blurriness, wrong words, incomplete text, or any other issues that detract from the overall visual experience. A score of 10 indicates a design that is visually stunning, while a score of 1 indicates a design that is visually unappealing. 

Prompt Following (1-10): Every page of the slide should be well aligned with the corresponding prompt given. Every object mentioned in the prompt should be present in the corresponding page. The object's attribute such as color, size, and position should also match the description in the prompt if assigned. If the prompt mentions text in a slide, it should be present and legible, and the texts should also match the logical flow of the prompt. The overall style of the slides should also align with the tone of the prompt. A score of 10 indicates perfect alignment, while a score of 1 indicates a complete mismatch. 

Style Consistency (1-10): The style of different pages in a set of slide should be consisitent. The color palette should be harmonious, the styles of elements in different pages should match, and the font type of text in different pages should also be consistent. A score of 10 indicates a set of slides that are perfectly consistent, while a score of 1 indicates a set of slides that are completely inconsistent. 

Your answer should be a dictionary with three keys: 'Aesthetics', 'Prompt Following' and 'Style Consistency' , each containing two keys: 'score' and 'reason'. The 'score' should be an integer between 1 and 10, and the 'reason' should be a string explaining the score.
\end{slidequalityprompt}

\begin{layerprompt}
You are an objective, strict, cold-blooded, experienced visual art designer. You are given an image and detailed information about it. The information includes:\\
1. The caption of the full image, which describes the overall content of the image.\\
    2. Two images. The first one is the original image, and the second one is exactly the same, while bounding boxes of the layers and indexes are added to the second one. The layers are given in z-order, from bottom to top, so higher layers may occlude lower layers. Each of the bounding boxes has its index labeled on it, in the same color of the box.\\
    3. The detailed information of each layer in the image, including the caption of the layer and the bounding box of the layer. Bounding boxes are in the format of [top\_left\_x, top\_left\_y, bottom\_right\_x, bottom\_right\_y]. The bbox coordinates are normalized to [0,1]. The occlusion relationship between layers is also provided, which can also be told from the second image with bounding boxes.\\
    4. For each layer, two auxiliary information is provided: "element\_type" and "layer\_description". "element\_type" can either be "block" or "object" for non-background layers. "block" means the layer doesn't contain specific objects, while "object" means the layer contains specific objects. For "block" layers, you should focus on the color and texture of the layer. For "object" layers, you should focus on the specific objects in the layer. "layer\_description" (if given) is a description of the content of the target layer without occlusions. You should decide the score based on the two images given, while the "layer\_description" is only for your reference.\\
    Now for the specified layer, you have to decide whether the region within the bounding box is consistent with the caption of the layer. Your answer should be a score ranging from 0 to 10. 0 means the region is completely irrelevant with the caption, while 10 means the region is perfectly aligned with the caption. The text in the image should be neglected. The key point is to determine whether the object described in the caption appear in the region. You should also provide a detailed reason for your decision. There are some rules for your decision:\\
    (1). You should consider only the bounded region and the caption of the layer. \\
    (2). If the caption of the layer is too vague to determine whether the object appears in the region, you should give a low score.\\
    (3). Most attention should be paid to the main object in the caption. If the main object is missing in the region, you should give a very low score. The main object must be clearly visible, different from the background. This is the most important rule and you must always follow it. If the main object is missing, or the region has no difference from the background, zero score. Specifically, for "object" layers, you should focus on the specific objects in the layer. You can also refer to the "layer\_description" to judge whether it's consistent with the caption.\\
    (4). If the caption specifies multiple objects, all of them should appear in the region. If some objects are missing, you should deduct points accordingly.\\
    (5). If the caption specifies a specific attribute of the object, the attribute should be visible in the region and consistent with the caption. For example, if the caption specifies the color of the object, the object should be in the specified color, or you should give a low score. Specifically, for "block" layers, you only have to consider the color and texture of the layer, and "layer\_description" only describes the color. If the color is close to the caption, such as beige compared to white, you shouldn't give low score.\\
    (6). If the caption described that the main object is on a colored background, the mismatch of color of the background could be tolerated. But if the caption only describes the color of the object or the background, the object or background should be in the specified color.\\
    (7). Remember the higher layers may occlude lower layers, so you should consider the occlusion relationship between layers. Specifically, higher layer occlude lower layer, and something not described in the caption of the lower layer appears in the region, this should be tolerated and you shouldn't deduct scores because of such reasons. Only layers with larger index occlude lower layers. To judge occlusion, you have to refer to the second image with bounding boxes, and also the detailed bounding box information of each layer if necessary.\\
    (8). Text in the image region should be neglected. If the caption specifies no text in the region but there is text, you don't have to deduct points.\\
    (9). If there are artifacts or implausible objects in the region, you should deduct points accordingly.\\
    (10). You have to be strict and objective in your evaluation and shouldn't easily give a very high score unless the bounded region is consistent with the caption.\\
    Important: Your answer should be in dict format following the example given below, and no other answer needed. 
\end{layerprompt}

We design a Global Quality Assurance Prompt for GPT-4o to evaluate the overall quality of generated business content from two aspects: visual aesthetics and adherence to the global prompt. Additionally, style consistency is assessed across different pages in a slide set. The detailed system prompts are shown in the brown and sky blue blocks, respectively.

\section{Details about Layer Generation Success Rate}
\label{sec:Appendix_LGSR}
We demonstrate the scoring prompt in our LGSR assessment pipeline in the section titled ``Layer-wise Quality Assurance Prompt for GPT-4o." It consists of two parts: (i) an introduction to the information and images provided, and (ii) rules to follow when scoring the layers in the generated images. We further provide a detailed example in Figure~\ref{fig:LGSR_example} to give a clearer picture of how our LGSR assessment pipeline works and what information is required at each step.

\section{Multi-Layer Transparent Infographic Generation}
\label{sec:Appendix_transparent}
\begin{figure}[!t]
\begin{minipage}[!t]{1\linewidth}
   \begin{subfigure}[b]{0.13\textwidth}
   \centering
   \includegraphics[width=1\textwidth]{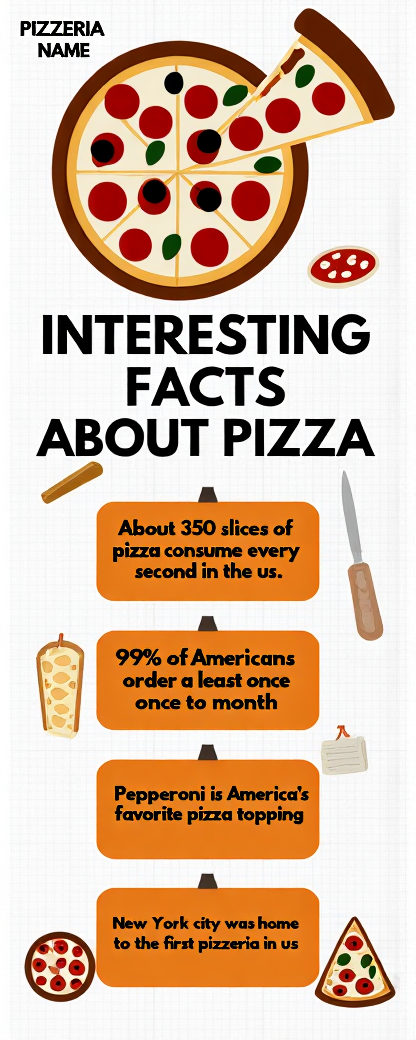}
   \vspace{-3mm}
   \caption*{Comp.}
   \end{subfigure}
   \begin{subfigure}[b]{0.13\textwidth}
   \centering
   \includegraphics[width=1\textwidth]{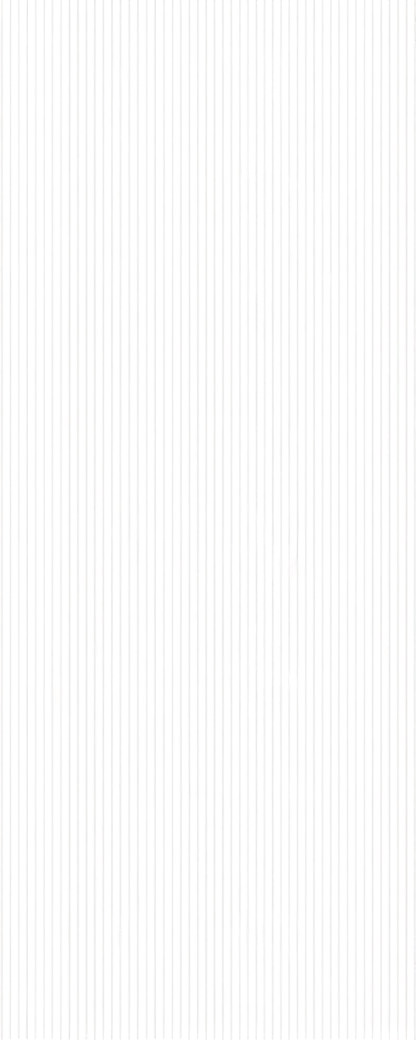}
   \vspace{-3mm}
   \caption*{\#0}
   \end{subfigure}
   \begin{subfigure}[b]{0.13\textwidth}
   \centering
   \includegraphics[width=1\textwidth]{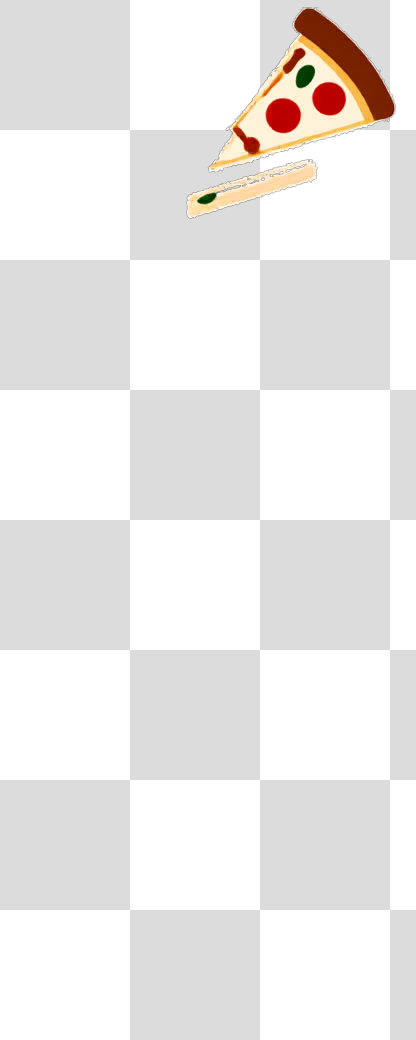}
   \vspace{-3mm}
   \caption*{\#1}
   \end{subfigure}
   \begin{subfigure}[b]{0.13\textwidth}
   \centering
   \includegraphics[width=1\textwidth]{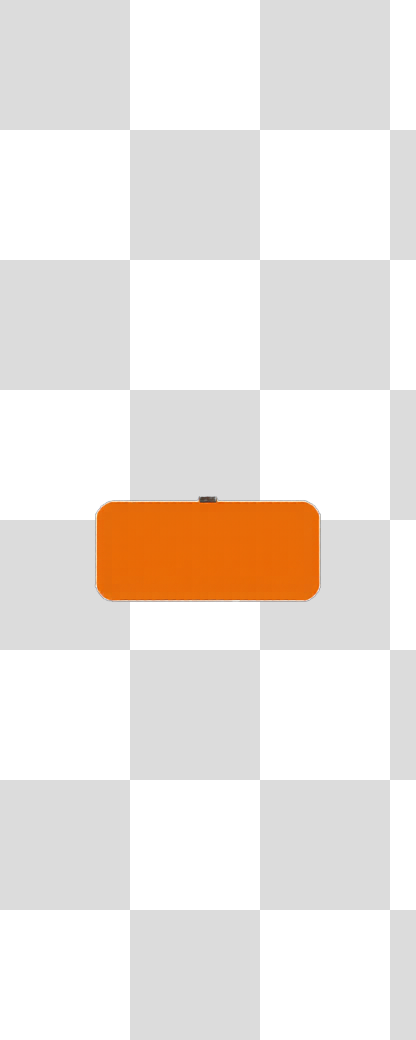}
   \vspace{-3mm}
   \caption*{\#2}
   \end{subfigure}
   \begin{subfigure}[b]{0.13\textwidth}
   \centering
   \includegraphics[width=1\textwidth]{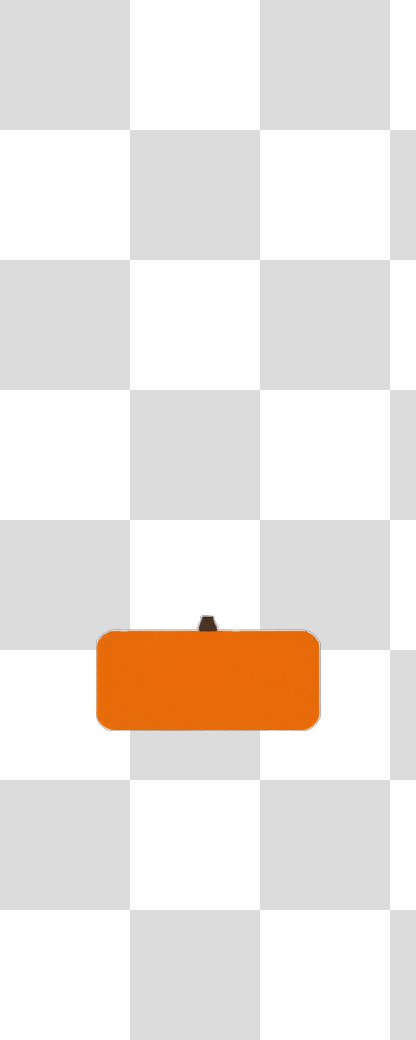}
   \vspace{-3mm}
   \caption*{\#3}
   \end{subfigure}
   \begin{subfigure}[b]{0.13\textwidth}
   \centering
   \includegraphics[width=1\textwidth]{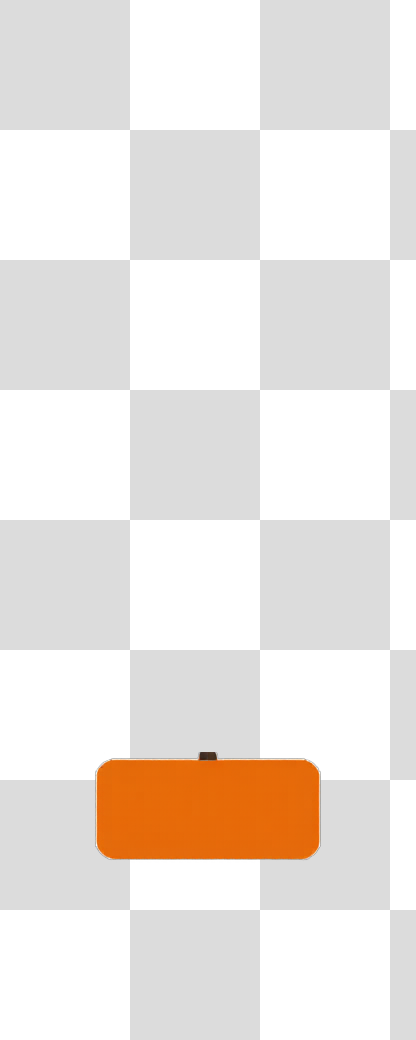}
   \vspace{-3mm}
   \caption*{\#4}
   \end{subfigure}
   \begin{subfigure}[b]{0.13\textwidth}
   \centering
   \includegraphics[width=1\textwidth]{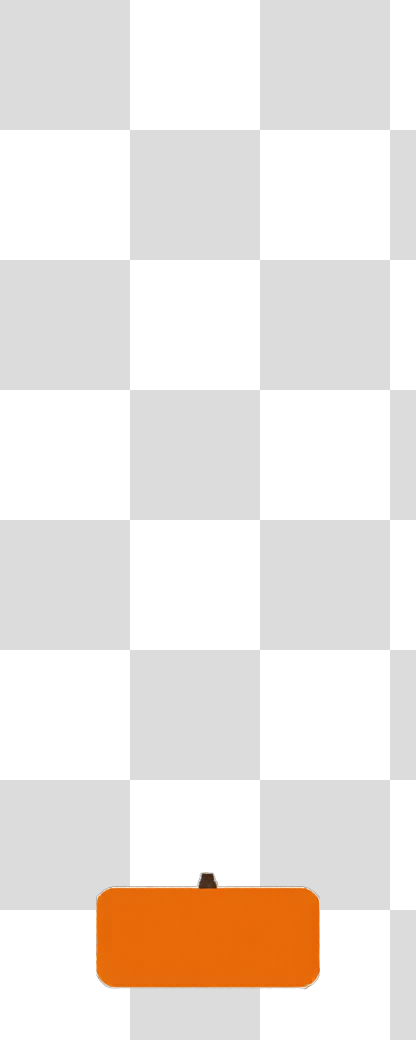}
   \vspace{-3mm}
   \caption*{\#5}
   \end{subfigure}
   \begin{subfigure}[b]{0.13\textwidth}
   \centering
   \includegraphics[width=1\textwidth]{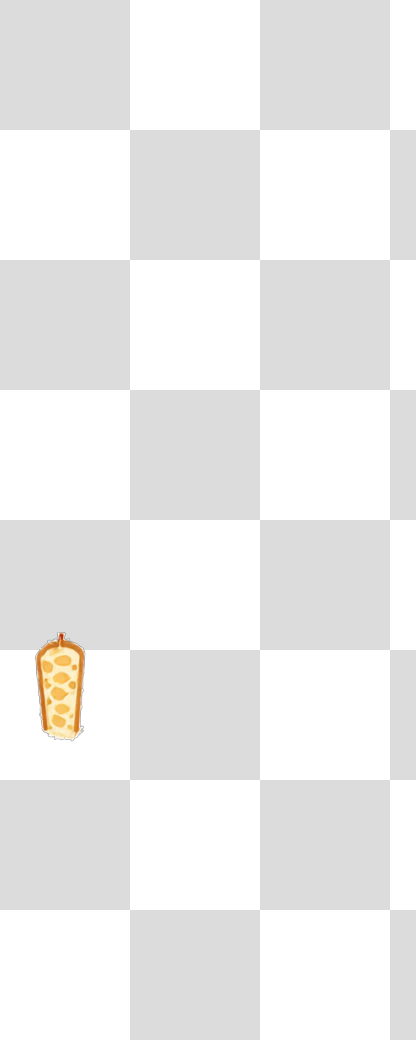}
   \vspace{-3mm}
   \caption*{\#6}
    \end{subfigure}
   \begin{subfigure}[b]{0.13\textwidth}
   \centering
   \includegraphics[width=1\textwidth]{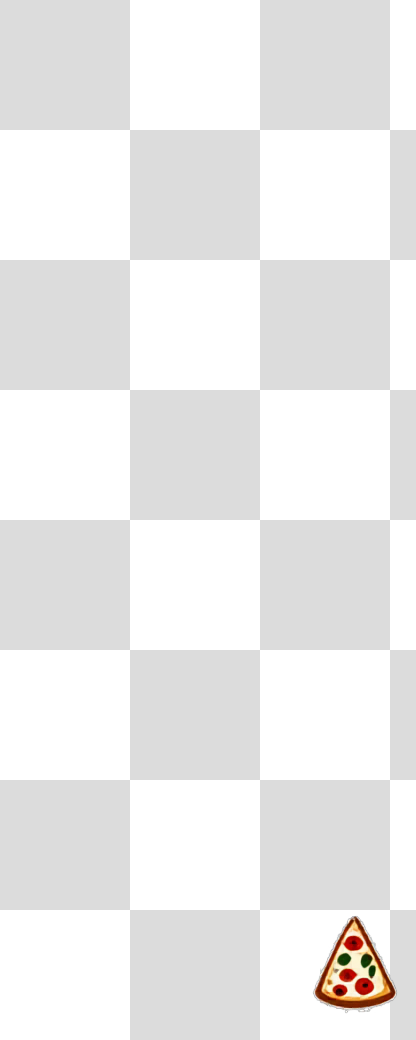}
   \vspace{-3mm}
   \caption*{\#7}
    \end{subfigure}
   \begin{subfigure}[b]{0.13\textwidth}
   \centering
   \includegraphics[width=1\textwidth]{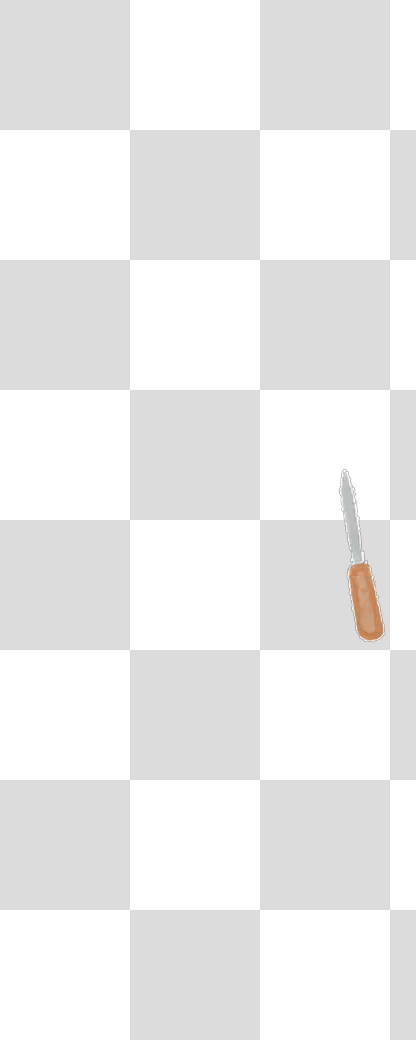}
   \vspace{-3mm}
   \caption*{\#8}
    \end{subfigure}
   \begin{subfigure}[b]{0.13\textwidth}
   \centering
   \includegraphics[width=1\textwidth]{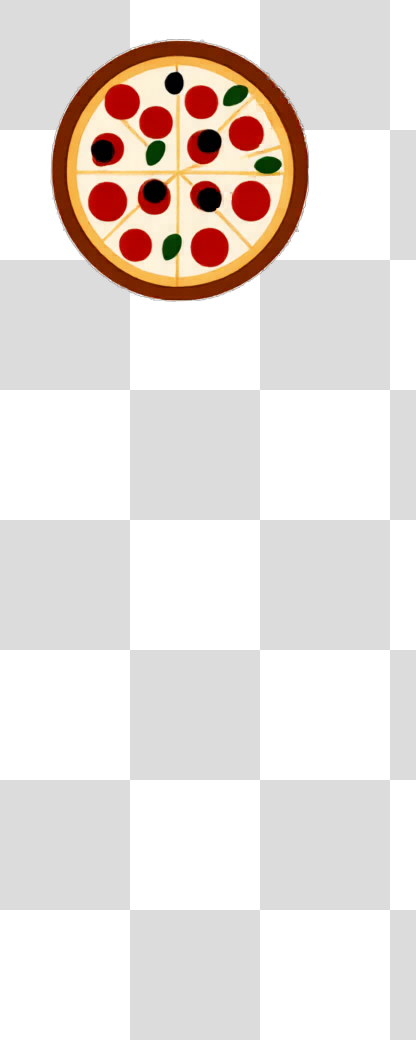}
   \vspace{-3mm}
   \caption*{\#9}
    \end{subfigure}
   \begin{subfigure}[b]{0.13\textwidth}
   \centering
   \includegraphics[width=1\textwidth]{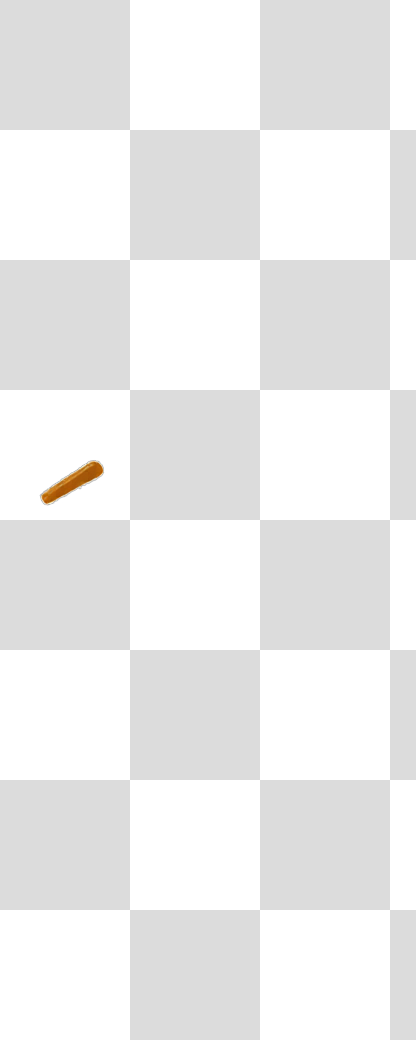}
   \vspace{-3mm}
   \caption*{\#10}
    \end{subfigure}
   \begin{subfigure}[b]{0.13\textwidth}
   \centering
   \includegraphics[width=1\textwidth]{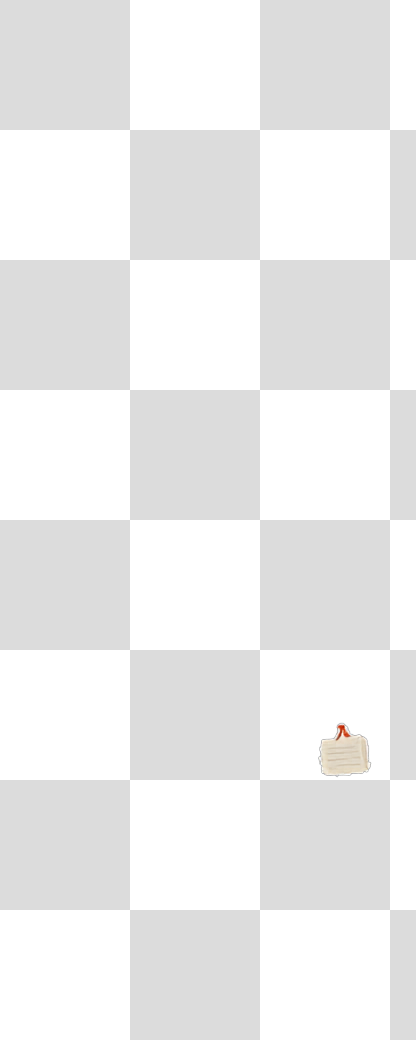}
   \vspace{-3mm}
   \caption*{\#11}
    \end{subfigure}
   \begin{subfigure}[b]{0.13\textwidth}
   \centering
   \includegraphics[width=1\textwidth]{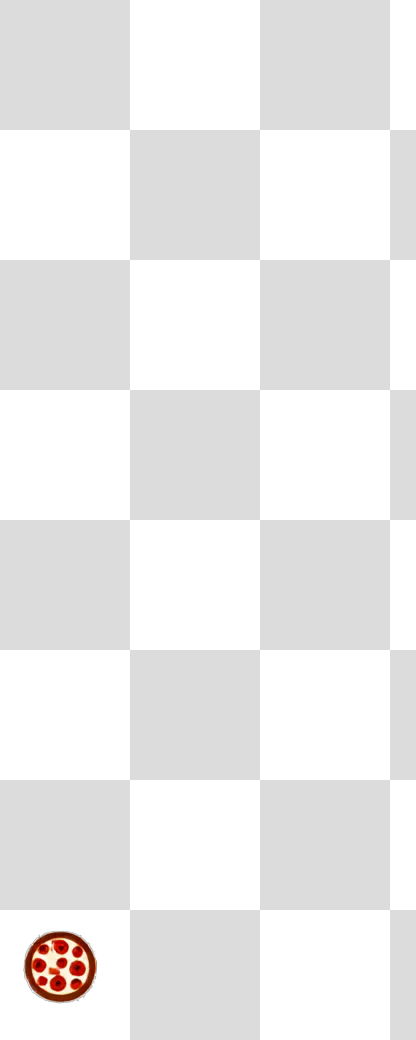}
   \vspace{-3mm}
   \caption*{\#12}
    \end{subfigure}
   \begin{subfigure}[b]{0.13\textwidth}
   \centering
   \includegraphics[width=1\textwidth]{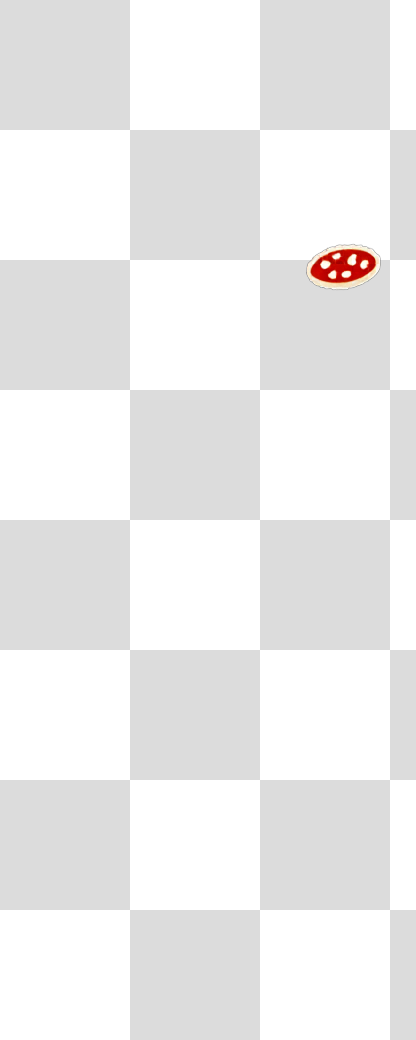}
   \vspace{-3mm}
   \caption*{\#13}
    \end{subfigure}
   \begin{subfigure}[b]{0.13\textwidth}
   \centering
   \includegraphics[width=1\textwidth]{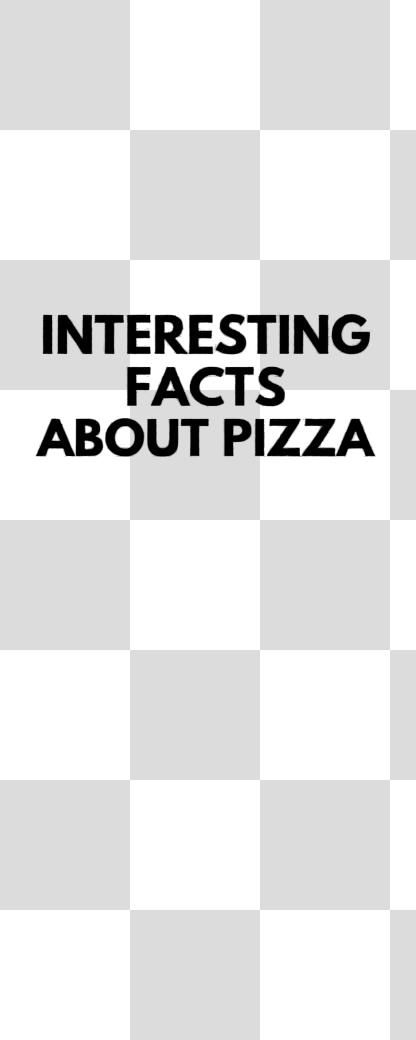}
   \vspace{-3mm}
   \caption*{\#14}
    \end{subfigure}
   \begin{subfigure}[b]{0.13\textwidth}
   \centering
   \includegraphics[width=1\textwidth]{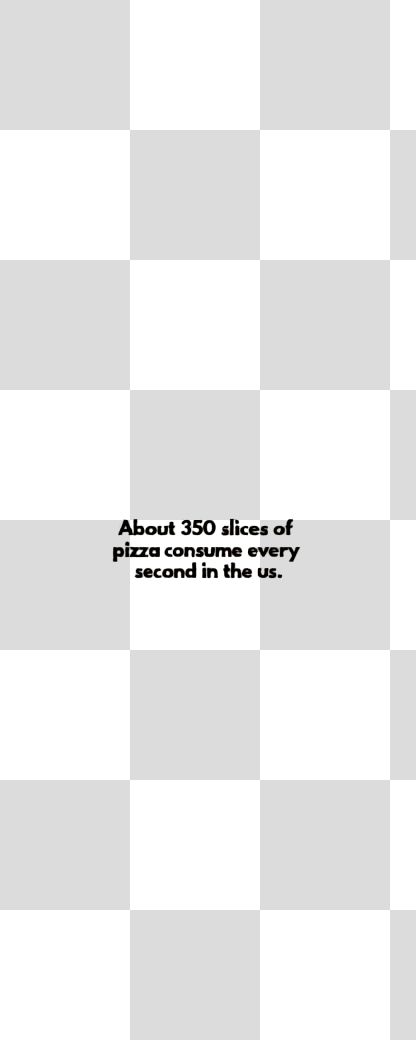}
   \vspace{-3mm}
   \caption*{\#15}
    \end{subfigure}
   \begin{subfigure}[b]{0.13\textwidth}
   \centering
   \includegraphics[width=1\textwidth]{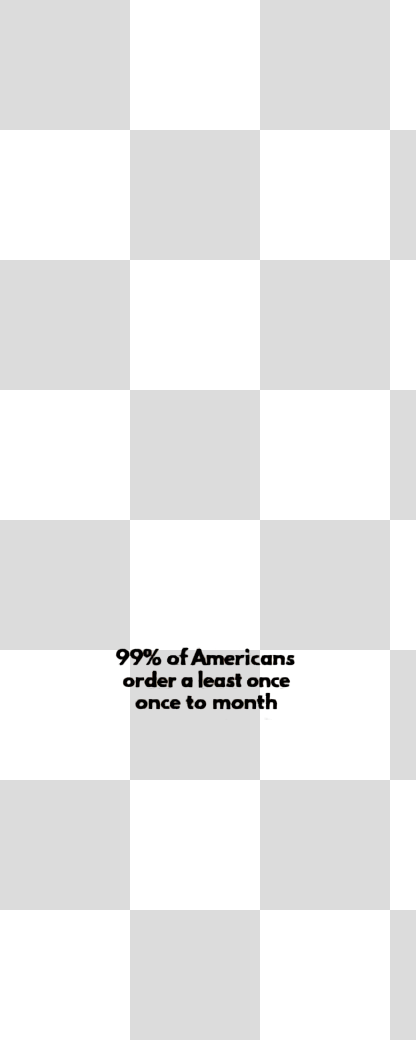}
   \vspace{-3mm}
   \caption*{\#16}
    \end{subfigure}
   \begin{subfigure}[b]{0.13\textwidth}
   \centering
   \includegraphics[width=1\textwidth]{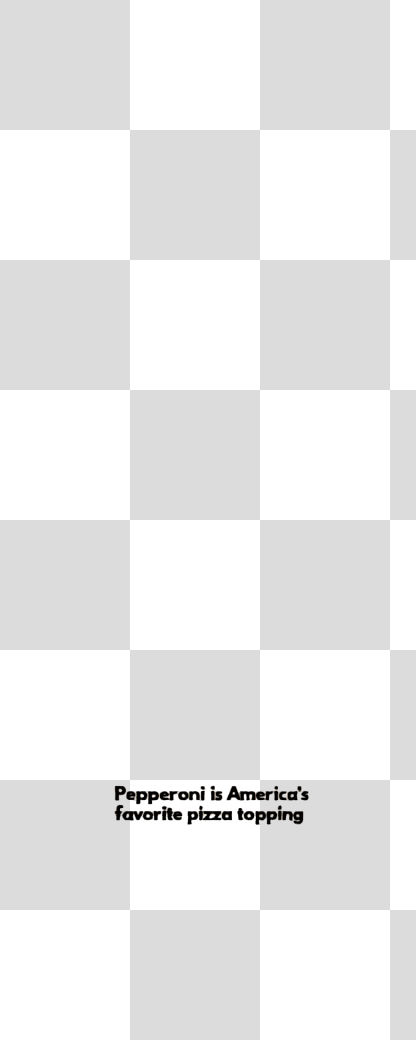}
   \vspace{-3mm}
   \caption*{\#17}
    \end{subfigure}
   \begin{subfigure}[b]{0.13\textwidth}
   \centering
   \includegraphics[width=1\textwidth]{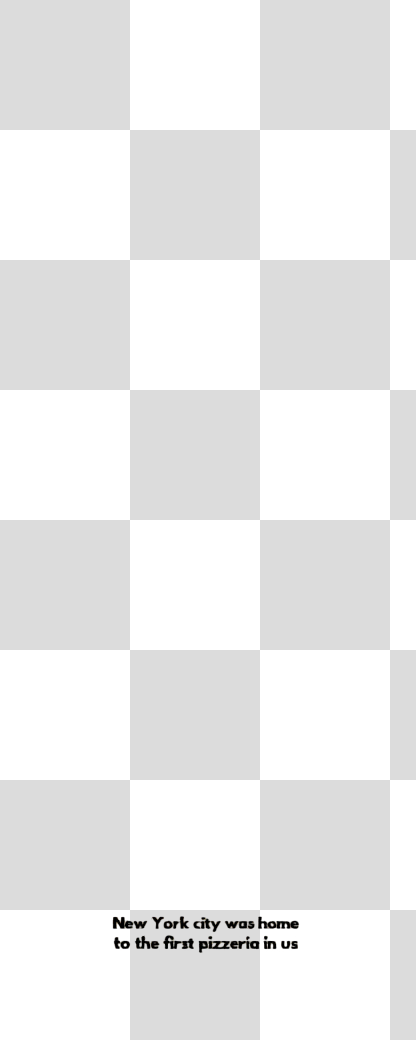}
   \vspace{-3mm}
   \caption*{\#18}
    \end{subfigure}
   \begin{subfigure}[b]{0.13\textwidth}
   \centering
   \includegraphics[width=1\textwidth]{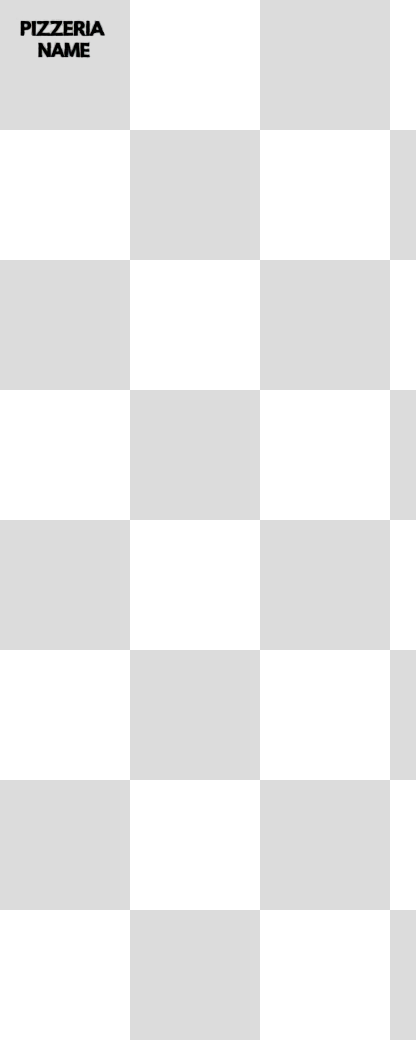}
   \vspace{-3mm}
   \caption*{\#19}
    \end{subfigure}
\end{minipage}
\vspace{-2mm}
\caption{\footnotesize{
Multi-layer Transparent Infographics Generation Results.
}}
\label{fig:multi_layer_infographics}
\vspace{-2mm}
\end{figure}

\begin{figure*}[t]
\begin{minipage}[t]{1\linewidth}
\centering
\begin{subfigure}[b]{0.12\textwidth}
\includegraphics[width=\textwidth]{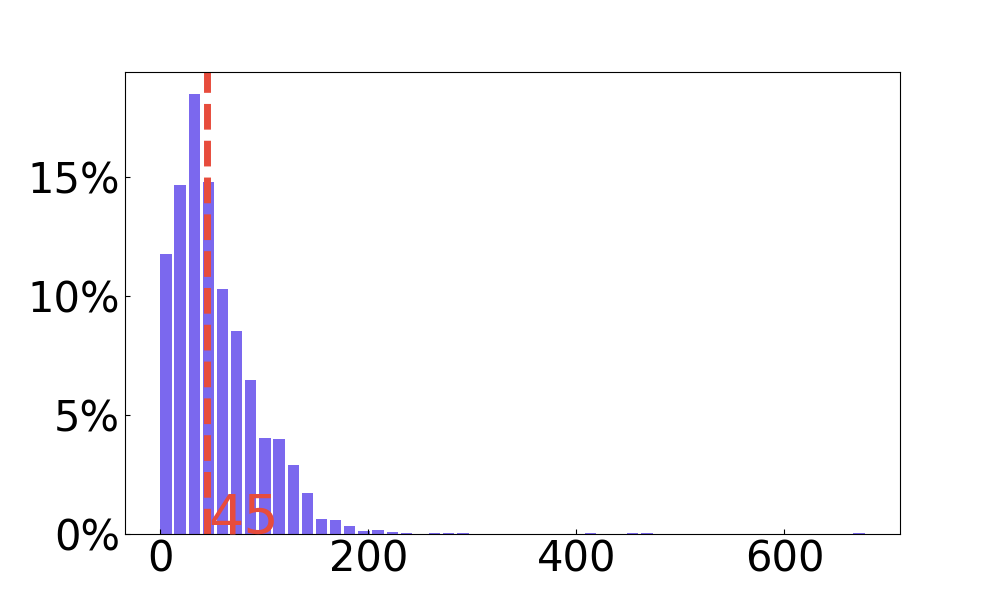}
\vspace{-5mm}
\caption{}
\end{subfigure}
\begin{subfigure}[b]{0.12\textwidth}
{\includegraphics[width=\textwidth]{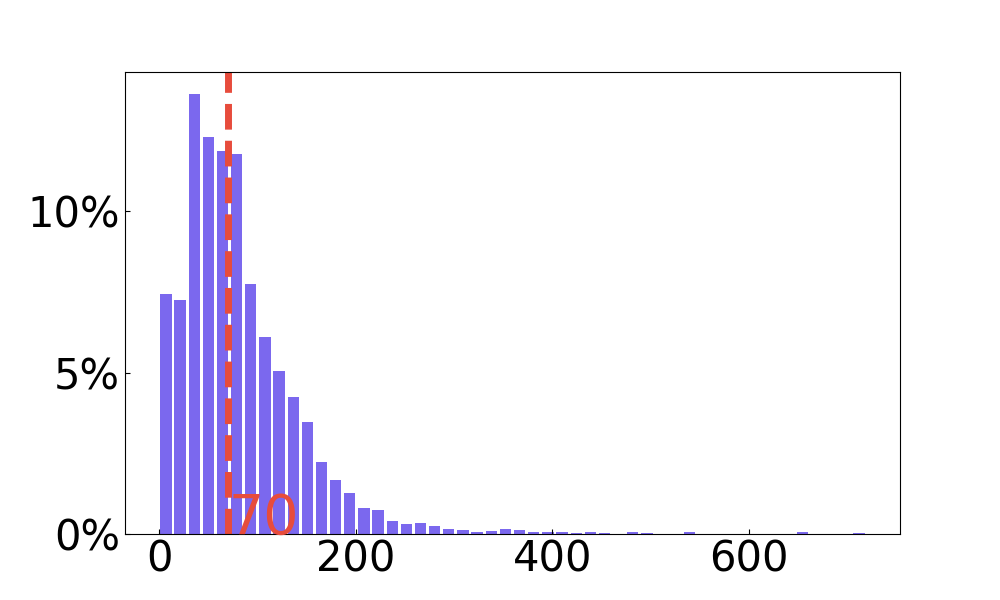}}
\vspace{-5mm}
\caption{}
\end{subfigure}
\begin{subfigure}[b]{0.12\textwidth}
{\includegraphics[width=\textwidth]{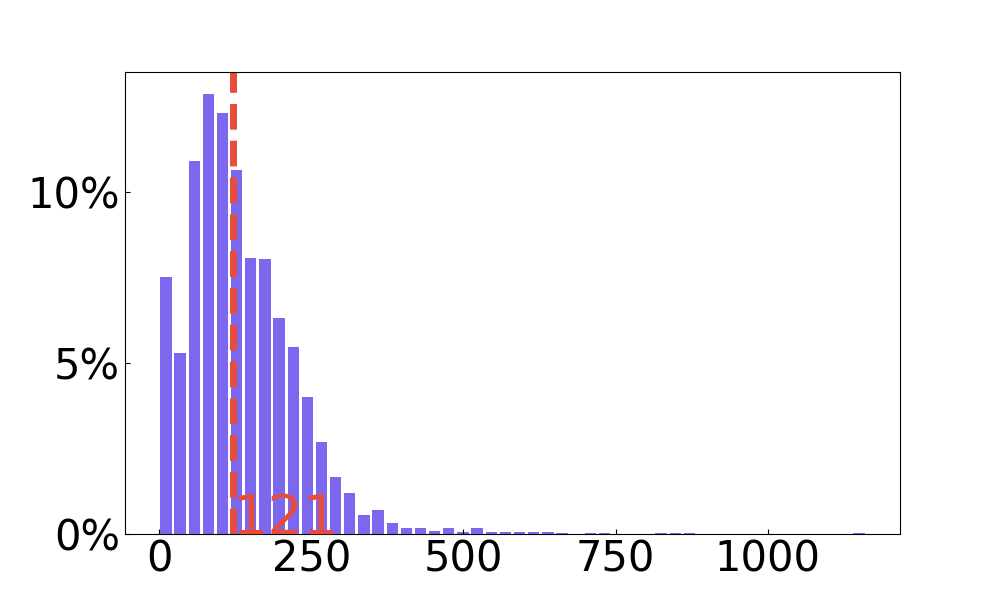}}
\vspace{-5mm}
\caption{}
\end{subfigure}
\begin{subfigure}[b]{0.12\textwidth}
{\includegraphics[width=\textwidth]{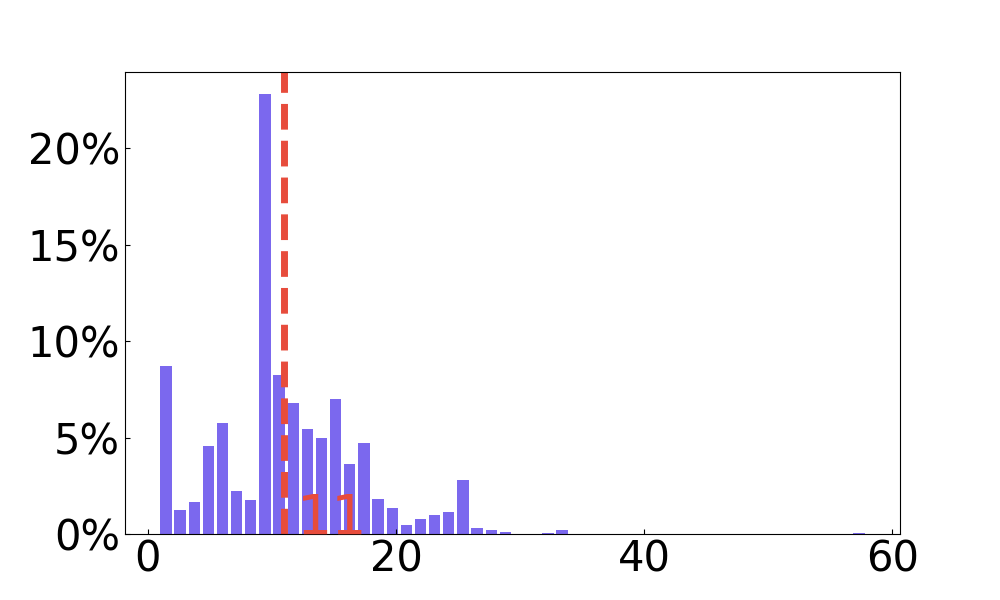}}
\vspace{-5mm}
\caption{}
\end{subfigure}
\begin{subfigure}[b]{0.12\textwidth}
\includegraphics[width=\textwidth]{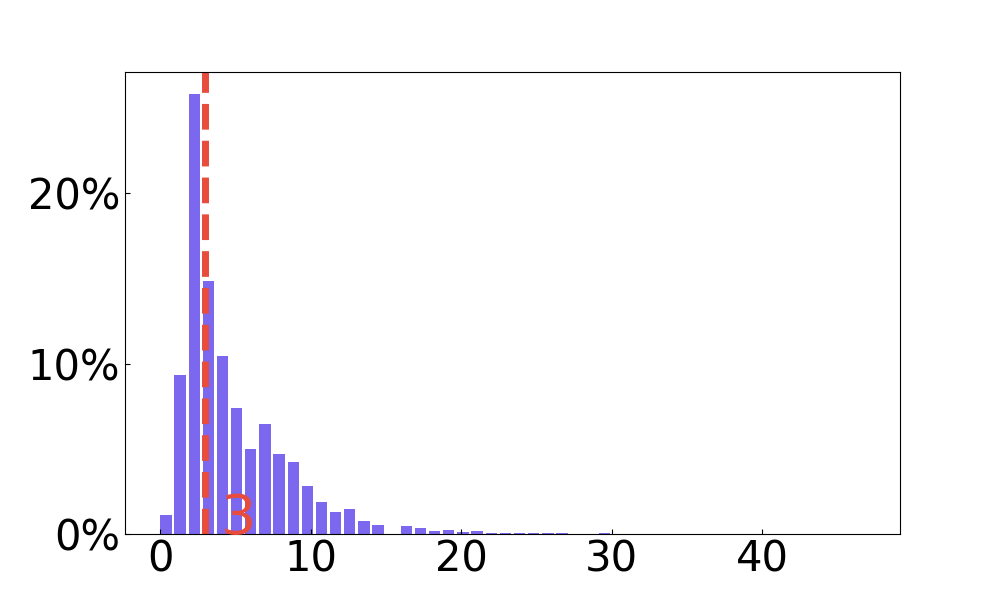}
\vspace{-5mm}
\caption{}
\end{subfigure}
\begin{subfigure}[b]{0.12\textwidth}
{\includegraphics[width=\textwidth]{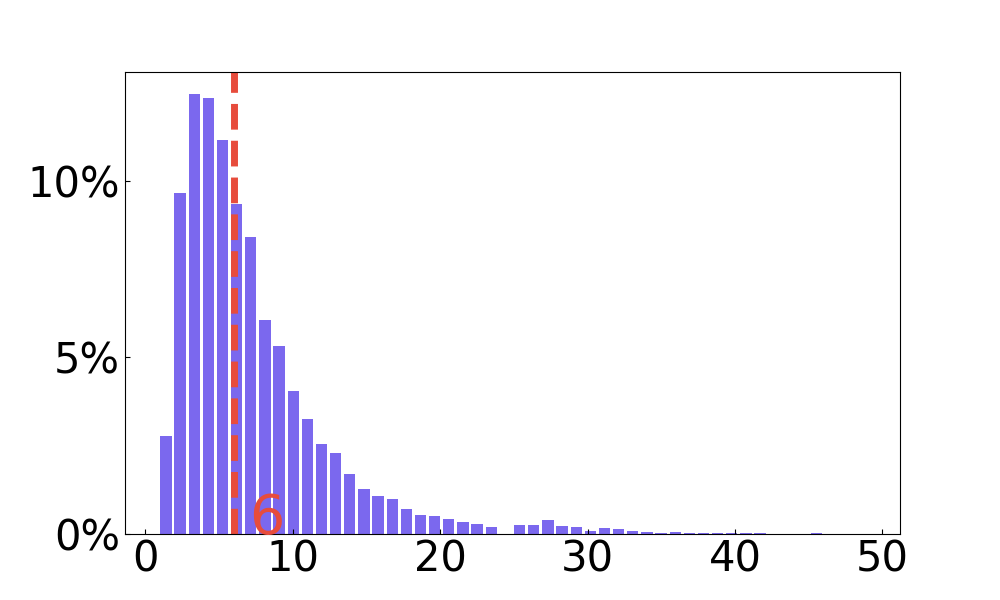}}
\vspace{-5mm}
\caption{}
\end{subfigure}
\begin{subfigure}[b]{0.12\textwidth}
{\includegraphics[width=\textwidth]{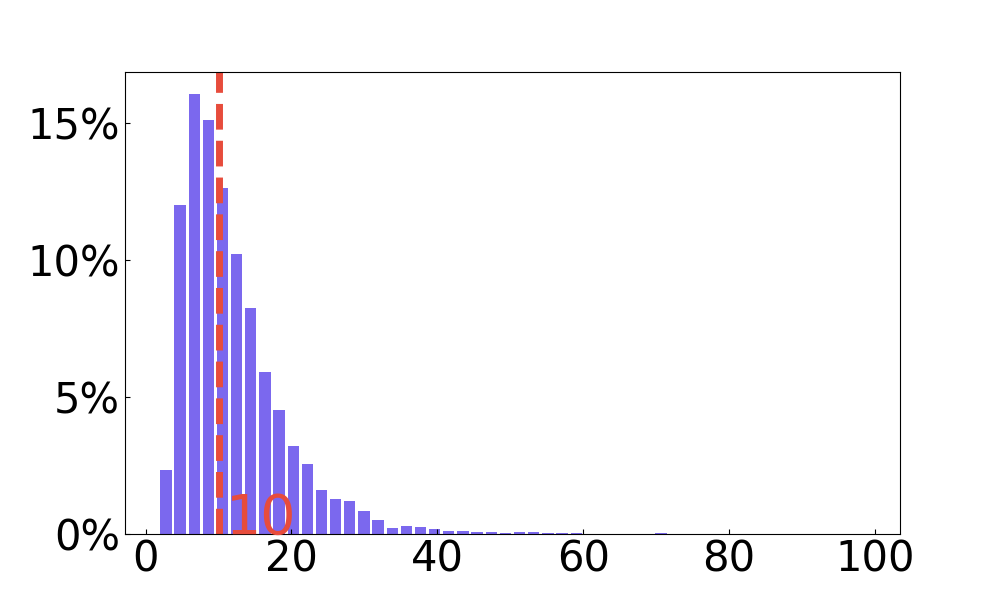}}
\vspace{-5mm}
\caption{}
\end{subfigure}
\begin{subfigure}[b]{0.12\textwidth}
{\includegraphics[width=\textwidth]{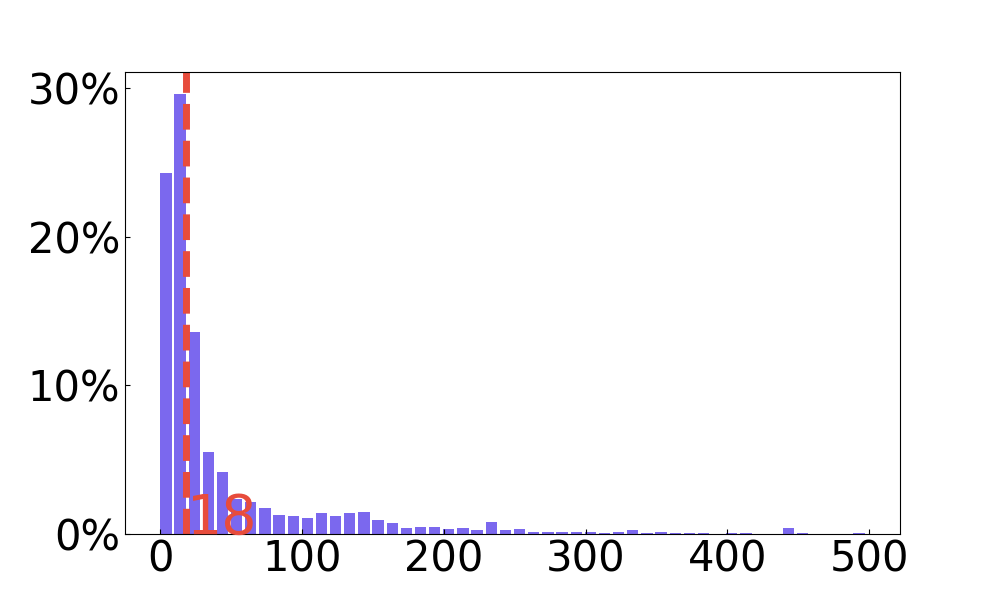}}
\vspace{-5mm}
\caption{}
\end{subfigure}
\vspace{-2mm}
\caption{\footnotesize{\textbf{Illustrating the statistics of our \ourslide:}(a)\# of text layers/set, (b)\# of non-text layers/set, (c) \# of total layers/set, (d) \# of pages/set, (e)\# of text layers/page, (f)\# of non-text layers/page, (g) \# of total layers/page, (h) \# of chars/text layer.  We mark the median values with red dashed lines.}}
\label{fig:slide_data_histogram}
\end{minipage}
\end{figure*}

As our \ourinfo essentially consists of multiple transparent layers, we follow \cite{pu2025artanonymousregiontransformer} and train a multi-layer generation model on \ourinfo and present a representative result in Figure~\ref{fig:multi_layer_infographics}, which consists of more than 20 transparent layers, including various visual element layers and visual text layers. We believe our \ourinfo will be a valuable contribution not only to infographic image generation but also to the multi-layer transparent infographic generation task.

\section{Multilingual Infographics and Slides Generation}
\label{sec:Appendix_multilingual}
Our \ourapproach model can render visual text in ten different languages, including English, German, Spanish, French, Italian, Portuguese, Russian, Chinese, Japanese, and Korean, thanks to our multilingual datasets \ourinfo and \ourslide. Table~\ref{tab:ocr_results} demonstrates the visual text spelling precision of the generated multilingual infographics and slides, while Figures~\ref{fig:multilingual_info} and \ref{fig:multilingual_slide} visualize some qualitative results. We find that our \ourapproach can achieve approximately 90\% visual text spelling precision in all ten languages, except in cases with dense, tiny visual texts, which slightly drag down the overall performance. We do not compare our multilingual generation results with state-of-the-art methods such as FLUX, as these approaches are primarily trained on English data.

\begin{table}[!t]
\begin{minipage}[t]{1\linewidth}
\centering
\tablestyle{1pt}{1.15}
\resizebox{1.0\linewidth}{!}
{
\begin{tabular}{l|cccc|cccc}
\multirow{2}{*}{Language} & \multicolumn{4}{c|}{Infographics Visual Text Spelling Precision ($\%$)}  & \multicolumn{4}{c}{Slides Visual Text Spelling Precision ($\%$)} \\\cline{2-9}
& $\le$10 layers & $\le$10-15 layers & $\le$15-20 layers & $\ge$20 layers & $\le$10 layers & $\le$10-20 layers & $\le$20-30 layers & $\ge$30 layers  \\
\shline
English & ${92.98}$ & ${84.25}$ & ${66.42}$ & ${55.48}$ & ${88.13}$ & ${88.41}$ & ${80.29}$ & ${66.81}$ \\
French & ${89.68}$ & ${81.02}$ & ${60.07}$ & ${50.97}$ & ${88.91}$ & ${85.32}$ & ${78.86}$ & ${60.71}$ \\
Spanish & ${87.31}$ & ${81.68}$ & ${62.37}$ & ${51.89}$ & ${84.38}$ & ${83.17}$ & ${74.19}$ & ${60.38}$ \\
German & ${84.11}$ & ${73.99}$ & ${54.24}$ & ${44.34}$ & ${87.06}$ & ${82.11}$ & ${73.80}$ & ${52.91}$ \\
Portuguese & ${89.15}$ & ${80.74}$ & ${65.73}$ & ${50.58}$ & ${87.40}$ & ${88.69}$ & ${77.04}$ & ${63.07}$ \\
Italian & ${87.23}$ & ${80.40}$ & ${64.12}$ & ${53.70}$ & ${93.53}$ & ${88.41}$ & ${77.43}$ & ${60.87}$ \\ 
Russian & ${83.10}$ & ${69.11}$ & ${52.11}$ & ${42.59}$ & ${83.04}$ & ${81.17}$ & ${70.16}$ & ${49.41}$ \\
Chinese & ${92.50}$ & ${91.71}$ & ${85.43}$ & ${69.18}$ & ${92.56}$ & ${93.22}$ & ${89.39}$ & ${85.68}$ \\
Japanese & ${94.07}$ & ${93.05}$ & ${91.79}$ & ${87.36}$ & ${95.68}$ & ${94.32}$ & ${95.65}$ & ${90.46}$ \\
Korean & ${91.94}$ & ${85.57}$ & ${80.80}$ & ${74.94}$ & ${92.1}$ & ${94.02}$ & ${89.90}$ & ${83.51}$ \\
\end{tabular}
}
\caption{
\footnotesize{{Illustrating the visual text spelling precision of the multilingual infographics and slides generation results.
}}}
\label{tab:ocr_results}
\end{minipage}
\end{table}

\section{Multi-Style Infographics Generation}
\label{sec:Appendix_multistyle}
As mentioned in Section~\ref{sec:approach_2}, four different LoRAs are used to generate diverse transparent layers for the retrieval database in our infographic data engine. During training, we add the corresponding trigger words to the prompts of the replaced layers, allowing the model to learn the style information. This approach enables us to customize our generation by simply adding trigger words to different layer prompts during inference, resulting in aesthetically pleasing infographics in various styles, as shown in Figure~\ref{fig:multi_style}.

\section{Comparison with State-of-the-art on Slides Generation}
\label{sec:Appendix_slides}
\begin{figure*}[!t]
\begin{minipage}[!t]{0.98\linewidth}
\begin{subfigure}[b]{0.16\textwidth}
\centering
\includegraphics[width=1\textwidth]{img/slides/s1p1.png}
\vspace{-3mm}
\end{subfigure}
\hfill
\begin{subfigure}[b]{0.16\textwidth}
\centering
\includegraphics[width=1\textwidth]{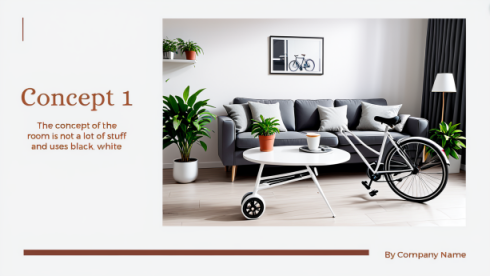}
\vspace{-3mm}
\end{subfigure}
\hfill
\begin{subfigure}[b]{0.16\textwidth}
\centering
\includegraphics[width=1\textwidth]{img/slides/s1p3.png}
\vspace{-3mm}
\end{subfigure}
\hfill
\begin{subfigure}[b]{0.16\textwidth}
\centering
\includegraphics[width=1\textwidth]{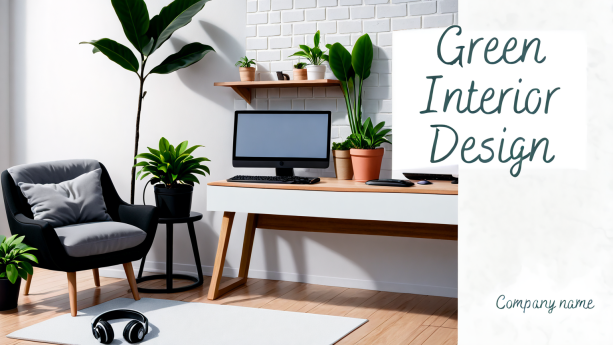}
\vspace{-3mm}
\end{subfigure}
\hfill
\begin{subfigure}[b]{0.16\textwidth}
\centering
\includegraphics[width=1\textwidth]{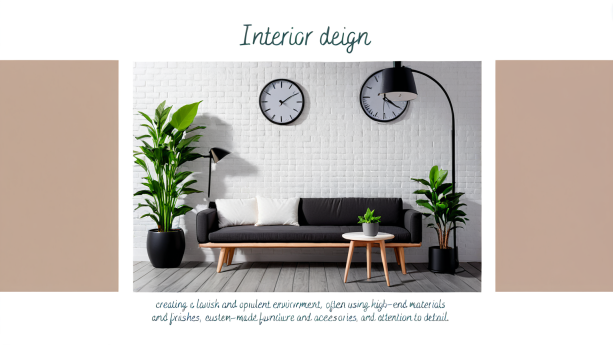}
\vspace{-3mm}
\end{subfigure}
\hfill
\begin{subfigure}[b]{0.16\textwidth}
\centering
\includegraphics[width=1\textwidth]{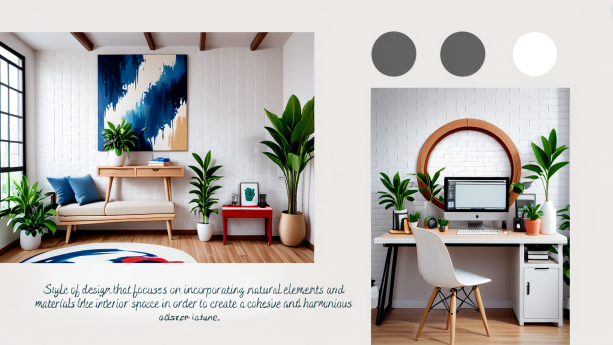}
\vspace{-3mm}
\end{subfigure}
\begin{subfigure}[b]{0.16\textwidth}
\centering
\includegraphics[width=1\textwidth]{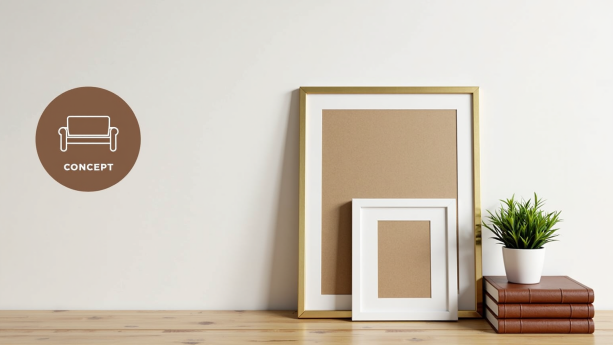}
\vspace{-3mm}
\end{subfigure}
\hfill
\begin{subfigure}[b]{0.16\textwidth}
\centering
\includegraphics[width=1\textwidth]{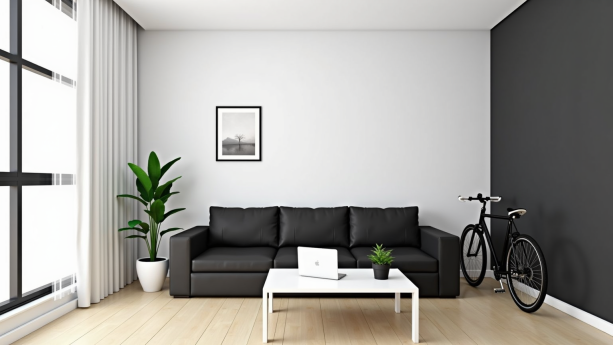}
\vspace{-3mm}
\end{subfigure}
\hfill
\begin{subfigure}[b]{0.16\textwidth}
\centering
\includegraphics[width=1\textwidth]{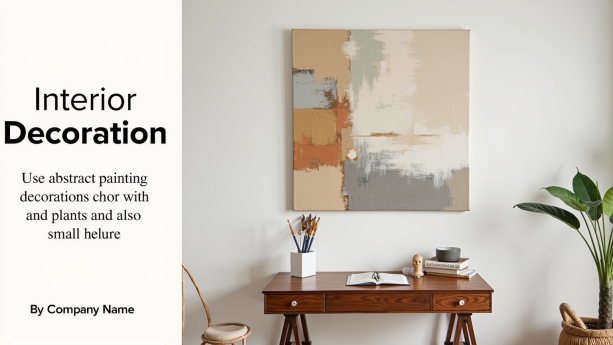}
\vspace{-3mm}
\end{subfigure}
\hfill
\begin{subfigure}[b]{0.16\textwidth}
\centering
\includegraphics[width=1\textwidth]{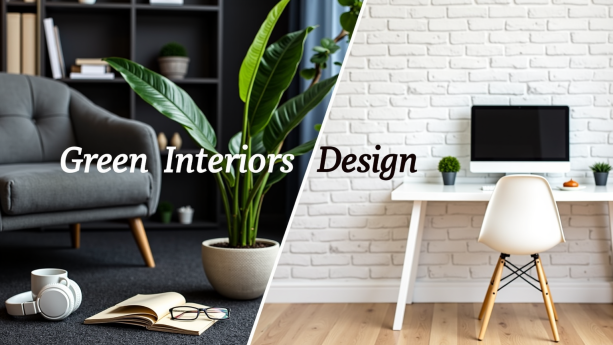}
\vspace{-3mm}
\end{subfigure}
\hfill
\begin{subfigure}[b]{0.16\textwidth}
\centering
\includegraphics[width=1\textwidth]{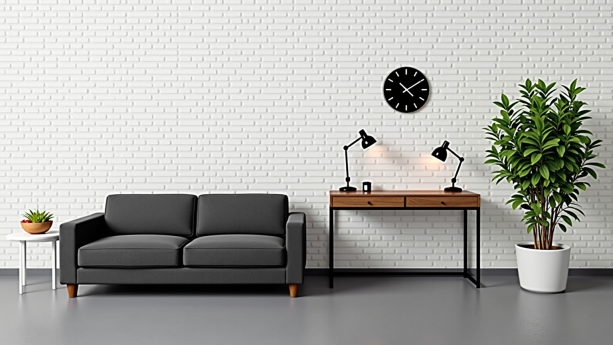}
\vspace{-3mm}
\end{subfigure}
\hfill
\begin{subfigure}[b]{0.16\textwidth}
\centering
\includegraphics[width=1\textwidth]{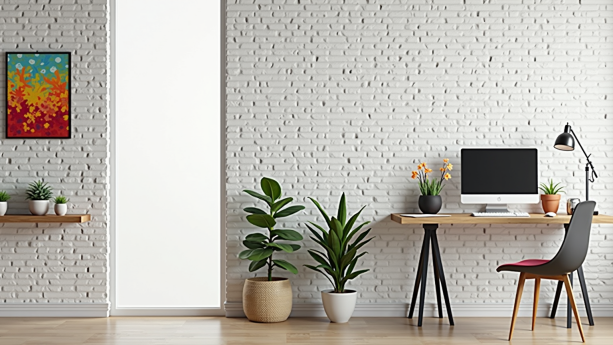}
\vspace{-3mm}
\end{subfigure}
\hfill
\begin{subfigure}[b]{0.16\textwidth}
\centering
\includegraphics[width=1\textwidth]{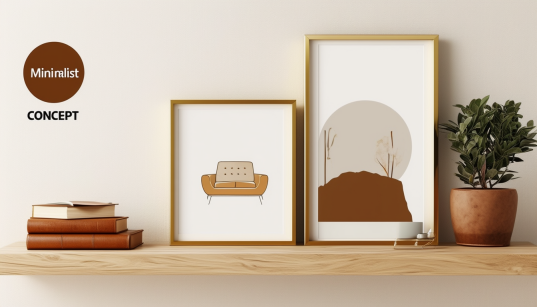}
\vspace{-3mm}
\end{subfigure}
\hfill
\begin{subfigure}[b]{0.16\textwidth}
\centering
\includegraphics[width=1\textwidth]{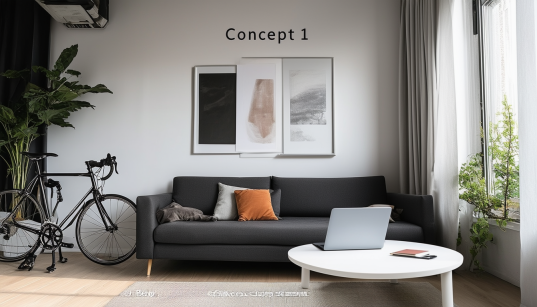}
\vspace{-3mm}
\end{subfigure}
\hfill
\begin{subfigure}[b]{0.16\textwidth}
\centering
\includegraphics[width=1\textwidth]{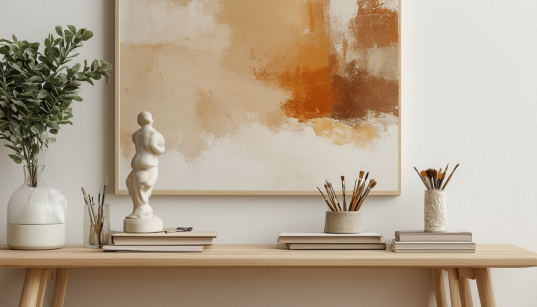}
\vspace{-3mm}
\end{subfigure}
\hfill
\begin{subfigure}[b]{0.16\textwidth}
\centering
\includegraphics[width=1\textwidth]{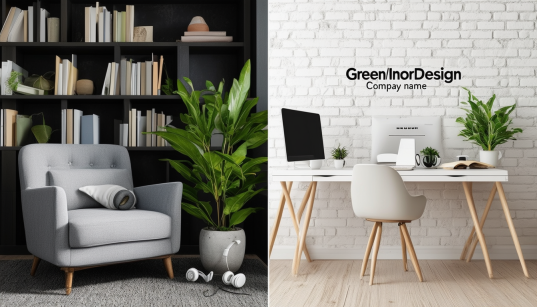}
\vspace{-3mm}
\end{subfigure}
\hfill
\begin{subfigure}[b]{0.16\textwidth}
\centering
\includegraphics[width=1\textwidth]{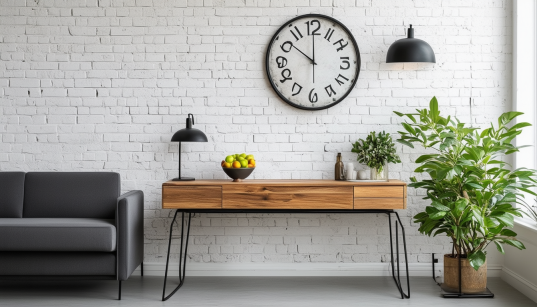}
\vspace{-3mm}
\end{subfigure}
\hfill
\begin{subfigure}[b]{0.16\textwidth}
\centering
\includegraphics[width=1\textwidth]{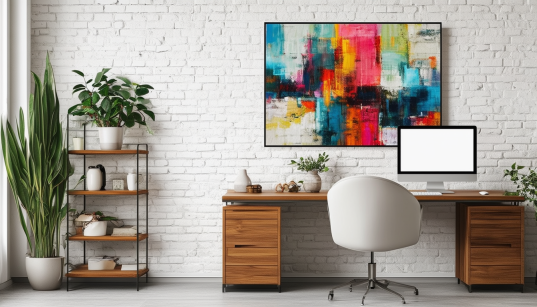}
\vspace{-3mm}
\end{subfigure}
\hfill
\begin{subfigure}[b]{0.16\textwidth}
\centering
\includegraphics[width=1\textwidth]{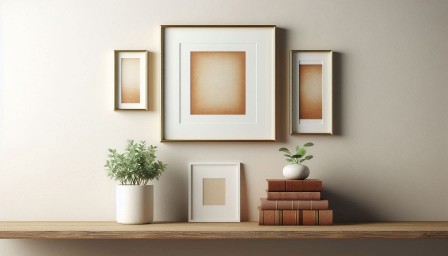}
\vspace{-3mm}
\end{subfigure}
\hfill
\begin{subfigure}[b]{0.16\textwidth}
\centering
\includegraphics[width=1\textwidth]{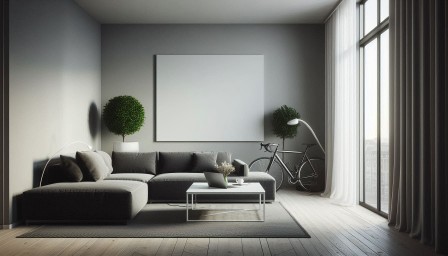}
\vspace{-3mm}
\end{subfigure}
\hfill
\begin{subfigure}[b]{0.16\textwidth}
\centering
\includegraphics[width=1\textwidth]{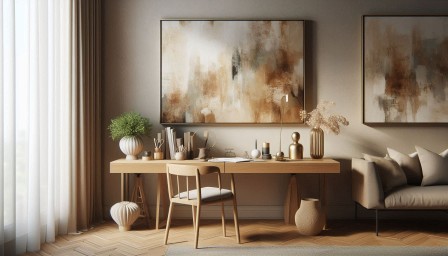}
\vspace{-3mm}
\end{subfigure}
\hfill
\begin{subfigure}[b]{0.16\textwidth}
\centering
\includegraphics[width=1\textwidth]{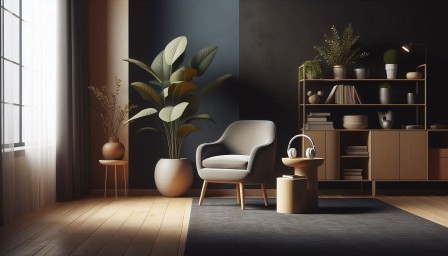}
\vspace{-3mm}
\end{subfigure}
\hfill
\begin{subfigure}[b]{0.16\textwidth}
\centering
\includegraphics[width=1\textwidth]{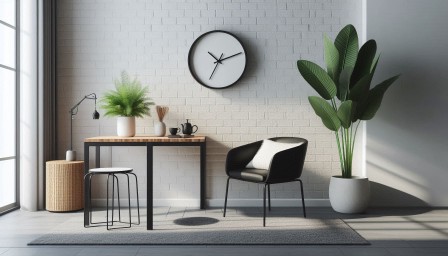}
\vspace{-3mm}

\end{subfigure}
\hfill
\begin{subfigure}[b]{0.16\textwidth}
\centering
\includegraphics[width=1\textwidth]{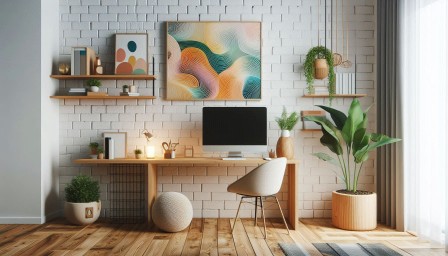}
\vspace{-3mm}
\end{subfigure}
\end{minipage}
\caption{\footnotesize{
\textbf{Qualitative comparison results of slides generation with SOTAs}. The 1st, 2nd, 3rd, and 4th rows correspond to the results generated with our \ourapproach, FLUX, SD3 Large, and DALL$\cdot$E3. The left three columns are in the same set, while the right three columns are in another.
}}
\label{fig:comparison_slide}
\end{figure*}

\begin{figure*}[!t]
\begin{minipage}[!t]{0.98\linewidth}
\begin{subfigure}[b]{0.1\textwidth}
\centering
\includegraphics[width=1\textwidth]{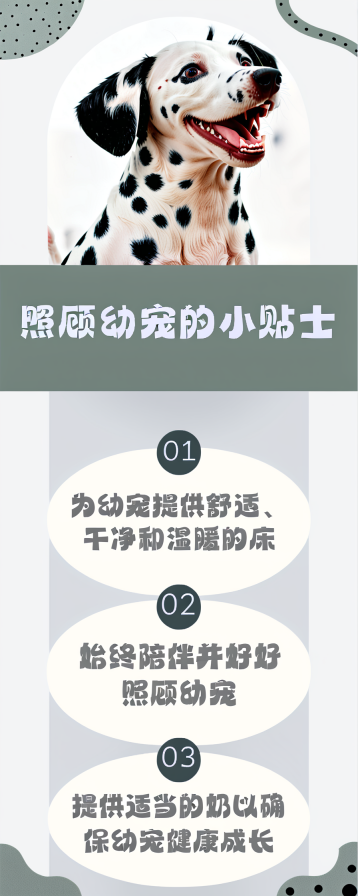}
\vspace{-3mm}
\end{subfigure}
\hfill
\begin{subfigure}[b]{0.1\textwidth}
\centering
\includegraphics[width=1\textwidth]{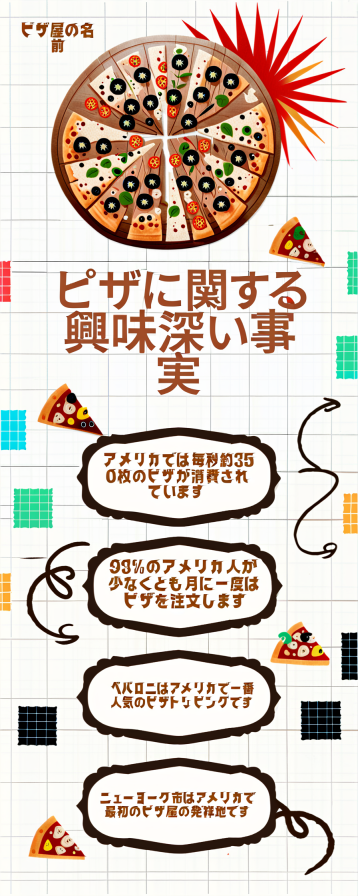}
\vspace{-3mm}
\end{subfigure}
\hfill
\begin{subfigure}[b]{0.1\textwidth}
\centering
\includegraphics[width=1\textwidth]{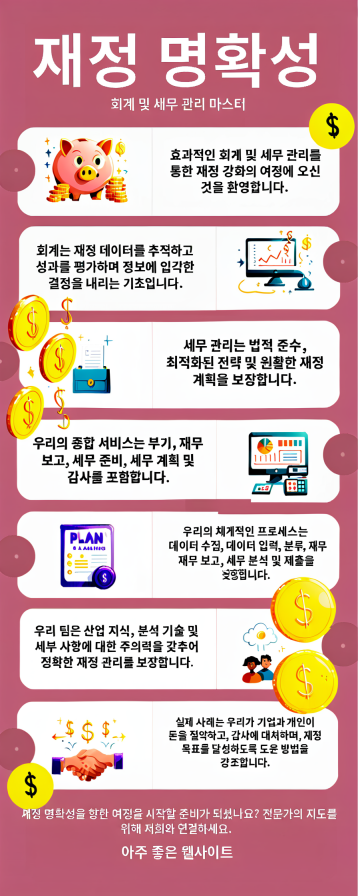}
\vspace{-3mm}
\end{subfigure}
\hfill
\begin{subfigure}[b]{0.1\textwidth}
\centering
\includegraphics[width=1\textwidth]{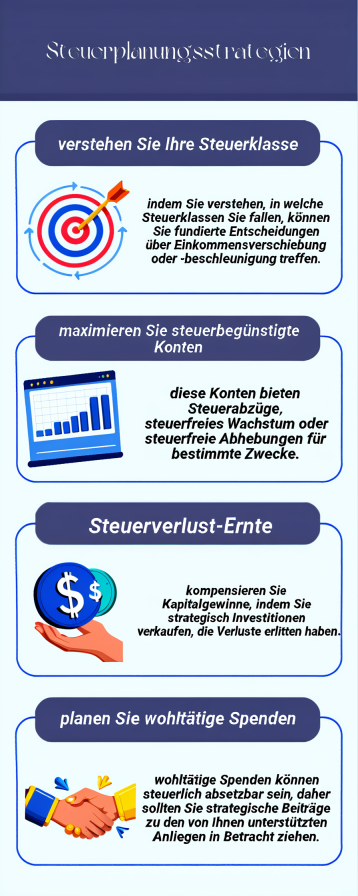}
\vspace{-3mm}
\end{subfigure}
\hfill
\begin{subfigure}[b]{0.1\textwidth}
\centering
\includegraphics[width=1\textwidth]{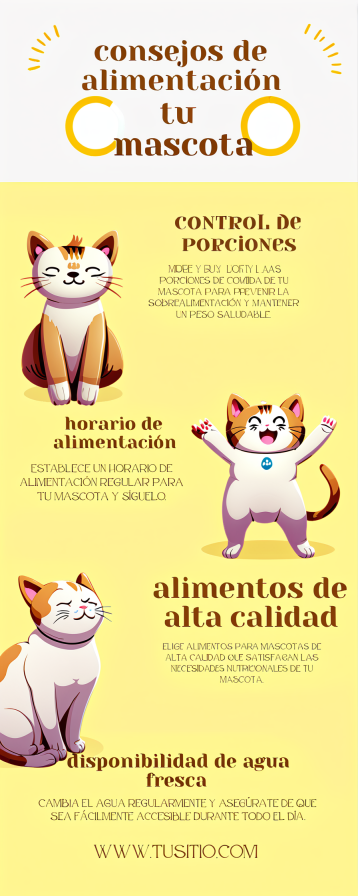}
\vspace{-3mm}
\end{subfigure}
\hfill
\begin{subfigure}[b]{0.1\textwidth}
\centering
\includegraphics[width=1\textwidth]{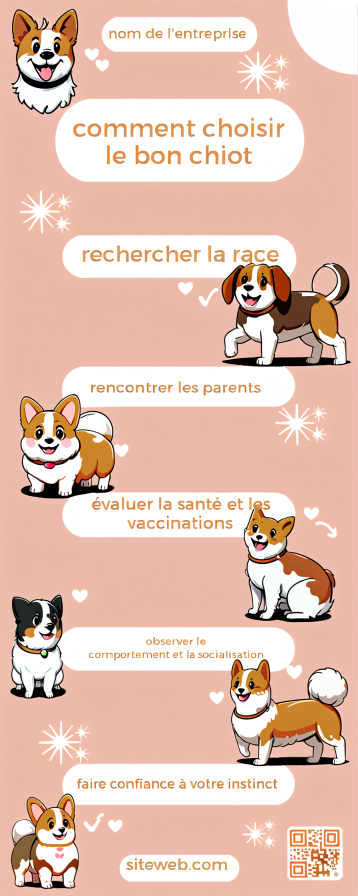}
\vspace{-3mm}
\end{subfigure}
\begin{subfigure}[b]{0.1\textwidth}
\centering
\includegraphics[width=1\textwidth]{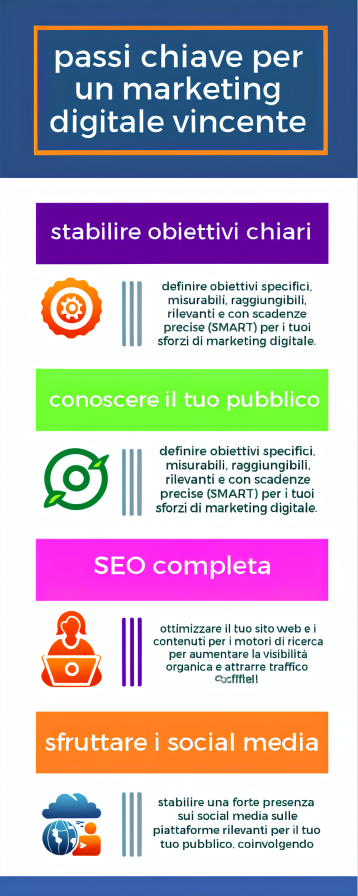}
\vspace{-3mm}
\end{subfigure}
\hfill
\begin{subfigure}[b]{0.1\textwidth}
\centering
\includegraphics[width=1\textwidth]{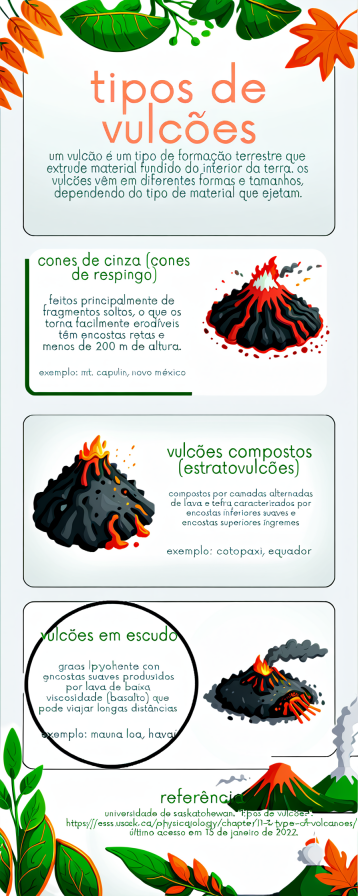}
\vspace{-3mm}
\end{subfigure}
\hfill
\begin{subfigure}[b]{0.1\textwidth}
\centering
\includegraphics[width=1\textwidth]{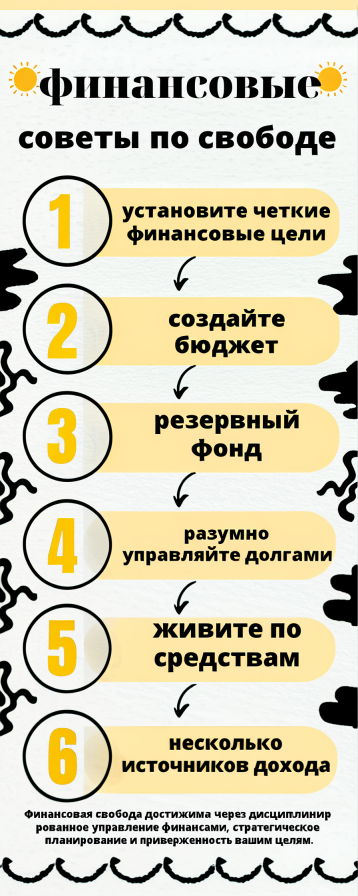}
\vspace{-3mm}
\end{subfigure}
\hfill
\begin{subfigure}[b]{0.1\textwidth}
\centering
\includegraphics[width=1\textwidth]{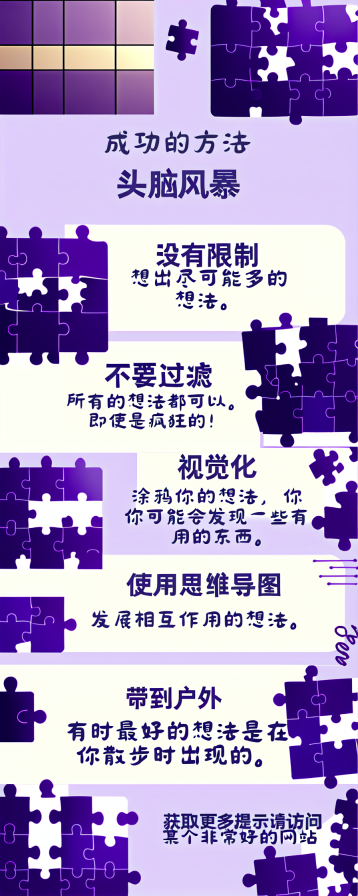}
\vspace{-3mm}
\caption*{\footnotesize{Chinese}}
\end{subfigure}
\hfill
\begin{subfigure}[b]{0.1\textwidth}
\centering
\includegraphics[width=1\textwidth]{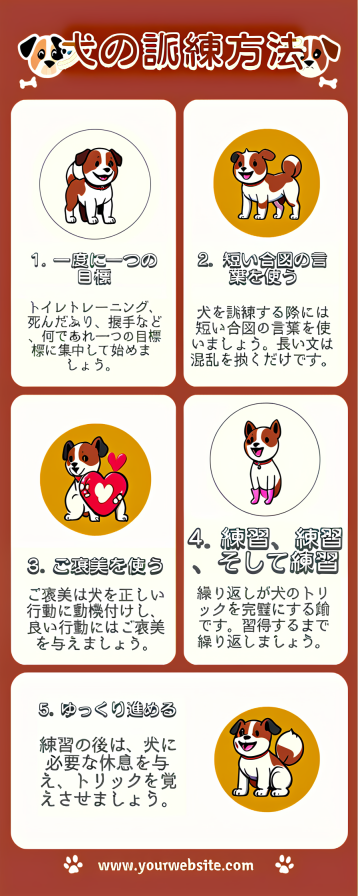}
\vspace{-3mm}
\caption*{\footnotesize{Japanese}}
\end{subfigure}
\hfill
\begin{subfigure}[b]{0.1\textwidth}
\centering
\includegraphics[width=1\textwidth]{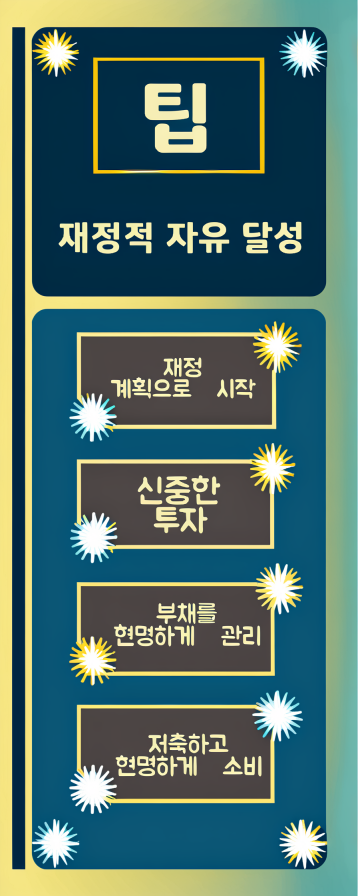}
\vspace{-3mm}
\caption*{\footnotesize{Korean}}
\end{subfigure}
\hfill
\begin{subfigure}[b]{0.1\textwidth}
\centering
\includegraphics[width=1\textwidth]{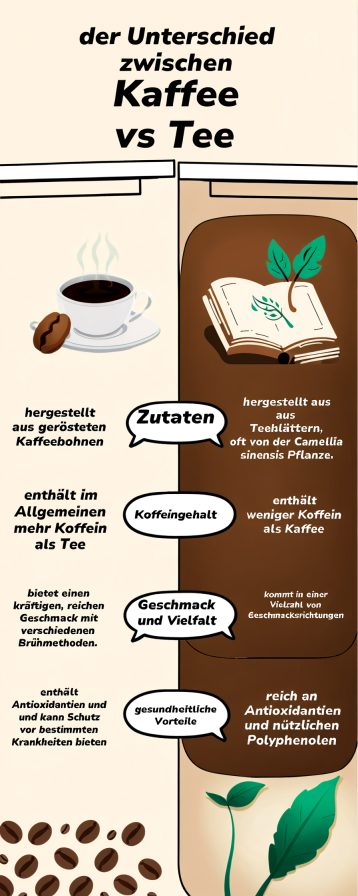}
\vspace{-3mm}
\caption*{\footnotesize{German}}
\end{subfigure}
\hfill
\begin{subfigure}[b]{0.1\textwidth}
\centering
\includegraphics[width=1\textwidth]{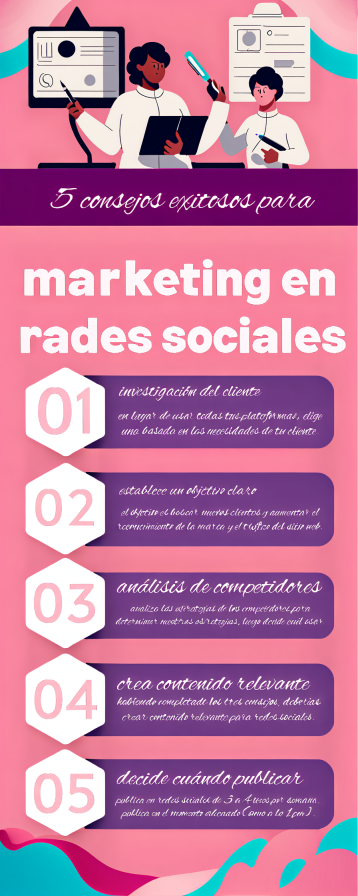}
\vspace{-3mm}
\caption*{\footnotesize{Spanish}}
\end{subfigure}
\hfill
\begin{subfigure}[b]{0.1\textwidth}
\centering
\includegraphics[width=1\textwidth]{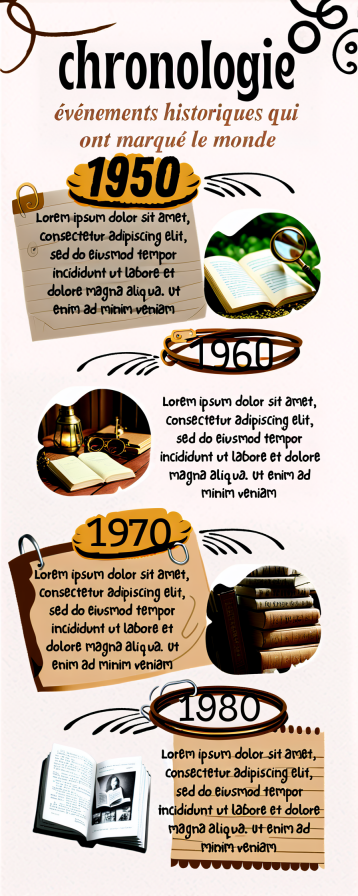}
\vspace{-3mm}
\caption*{\footnotesize{French}}
\end{subfigure}
\hfill
\begin{subfigure}[b]{0.1\textwidth}
\centering
\includegraphics[width=1\textwidth]{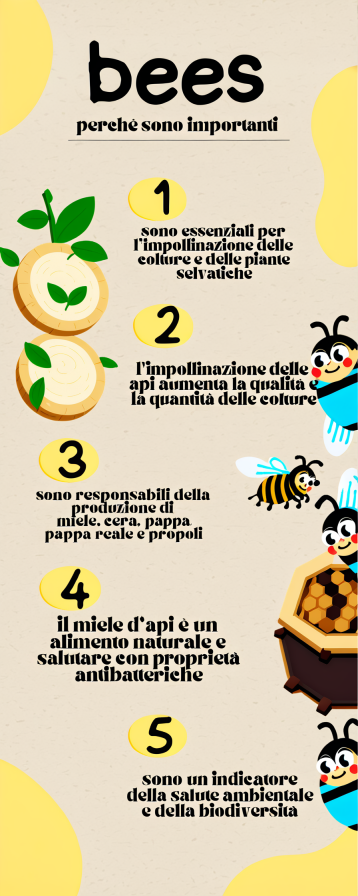}
\vspace{-3mm}
\caption*{\footnotesize{Italian}}
\end{subfigure}
\hfill
\begin{subfigure}[b]{0.1\textwidth}
\centering
\includegraphics[width=1\textwidth]{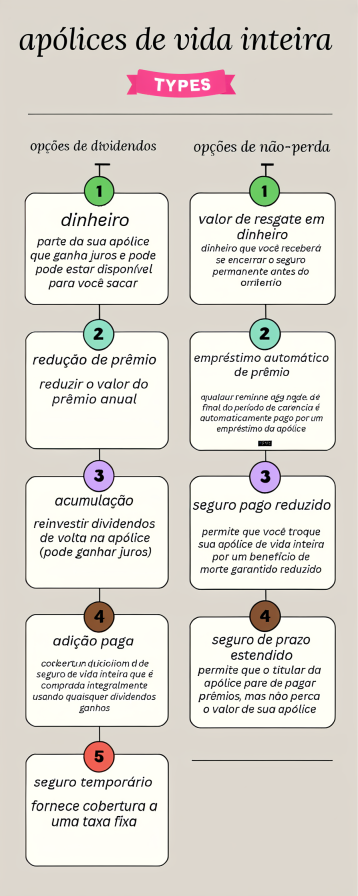}
\vspace{-3mm}
\caption*{\footnotesize{Portuguese}}
\end{subfigure}
\hfill
\begin{subfigure}[b]{0.1\textwidth}
\centering
\includegraphics[width=1\textwidth]{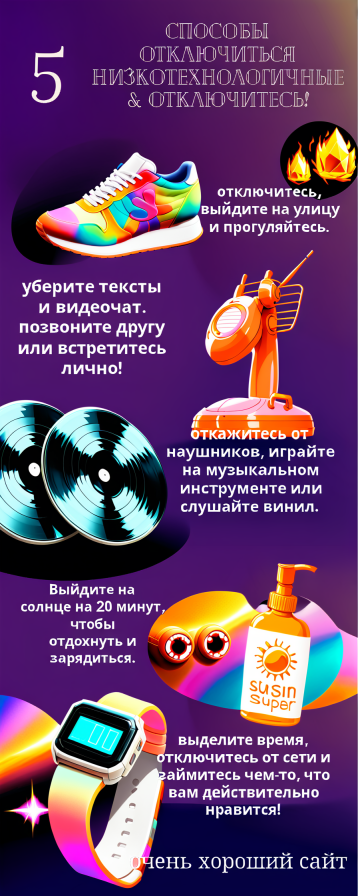}
\vspace{-3mm}
\caption*{\footnotesize{Russian}}
\end{subfigure}
\end{minipage}
\caption{\footnotesize{
\textbf{Qualitative results of multilingual infographics generation}. 
}}
\label{fig:multilingual_info}
\vspace{-5mm}
\end{figure*}

\begin{figure}[!t]
\begin{minipage}[!t]{1\linewidth}
\begin{subfigure}[b]{0.32\textwidth}
\centering
\includegraphics[width=1\textwidth]{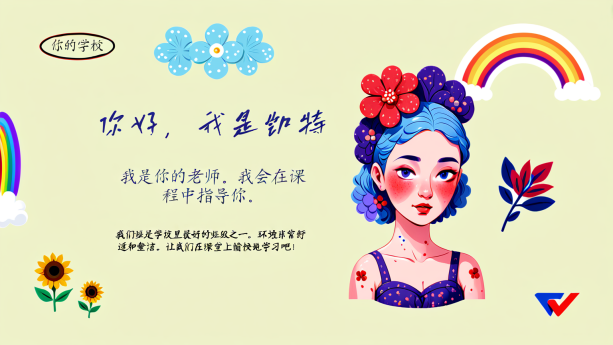}
\vspace{-3mm}
\end{subfigure}
\hfill
\begin{subfigure}[b]{0.32\textwidth}
\centering
\includegraphics[width=1\textwidth]{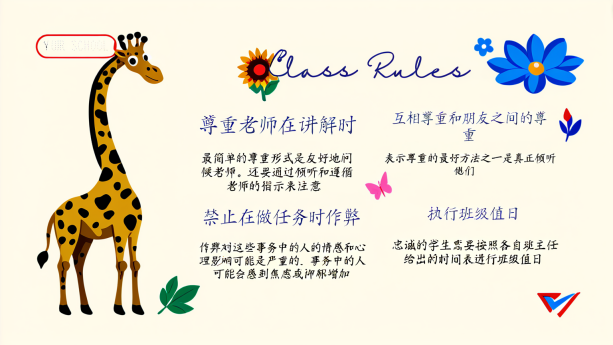}
\vspace{-3mm}
\end{subfigure}
\hfill
\begin{subfigure}[b]{0.32\textwidth}
\centering
\includegraphics[width=1\textwidth]{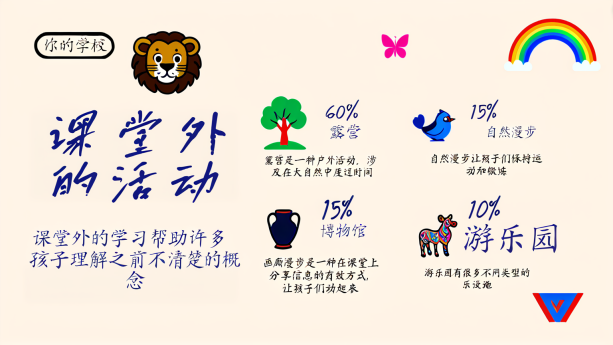}
\vspace{-3mm}
\end{subfigure}
\hfill
\begin{subfigure}[b]{0.32\textwidth}
\centering
\includegraphics[width=1\textwidth]{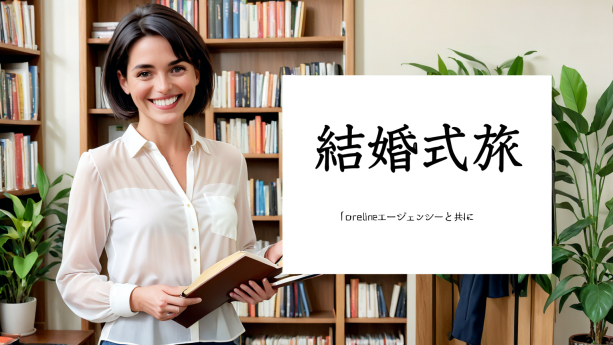}
\vspace{-3mm}
\end{subfigure}
\hfill
\begin{subfigure}[b]{0.32\textwidth}
\centering
\includegraphics[width=1\textwidth]{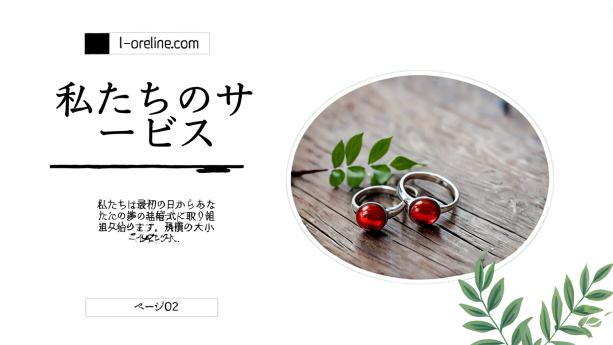}
\vspace{-3mm}
\end{subfigure}
\hfill
\begin{subfigure}[b]{0.32\textwidth}
\centering
\includegraphics[width=1\textwidth]{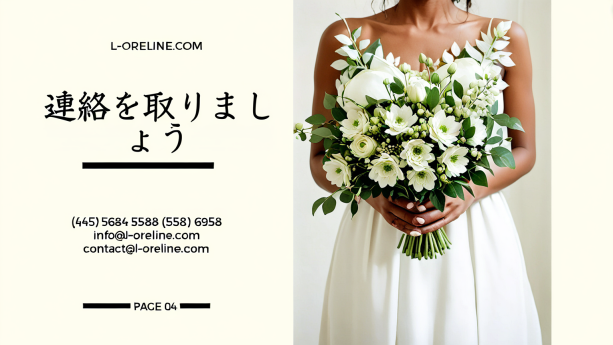}
\vspace{-3mm}
\end{subfigure}
\begin{subfigure}[b]{0.32\textwidth}
\centering
\includegraphics[width=1\textwidth]{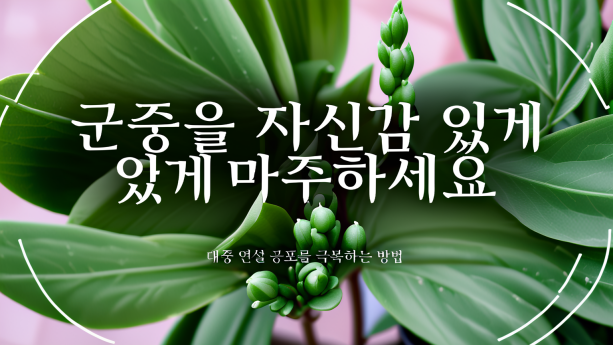}
\vspace{-3mm}
\end{subfigure}
\hfill
\begin{subfigure}[b]{0.32\textwidth}
\centering
\includegraphics[width=1\textwidth]{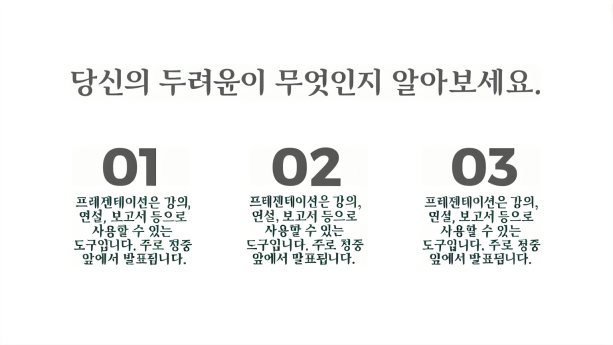}
\vspace{-3mm}
\end{subfigure}
\hfill
\begin{subfigure}[b]{0.32\textwidth}
\centering
\includegraphics[width=1\textwidth]{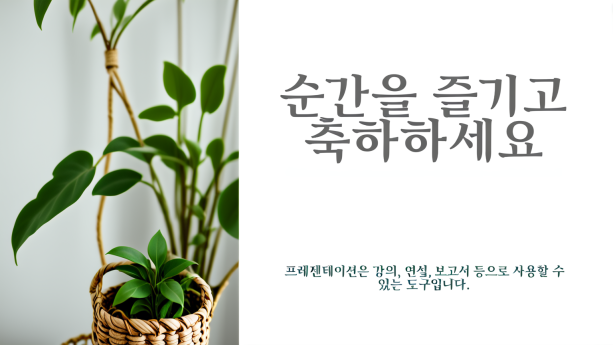}
\vspace{-3mm}
\end{subfigure}
\hfill
\begin{subfigure}[b]{0.32\textwidth}
\centering
\includegraphics[width=1\textwidth]{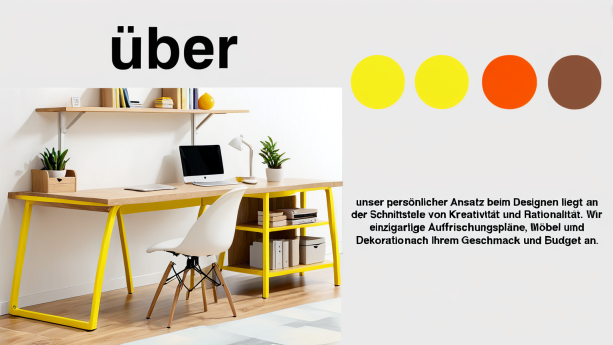}
\vspace{-3mm}
\end{subfigure}
\hfill
\begin{subfigure}[b]{0.32\textwidth}
\centering
\includegraphics[width=1\textwidth]{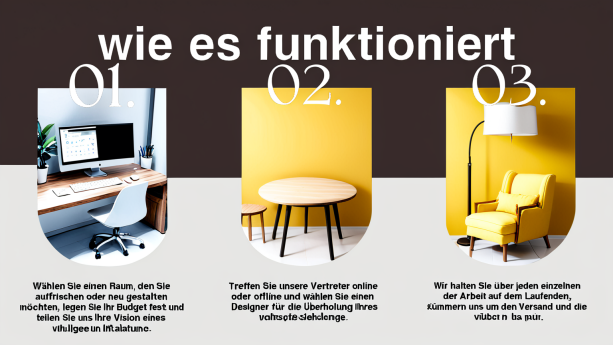}
\vspace{-3mm}
\end{subfigure}
\hfill
\begin{subfigure}[b]{0.32\textwidth}
\centering
\includegraphics[width=1\textwidth]{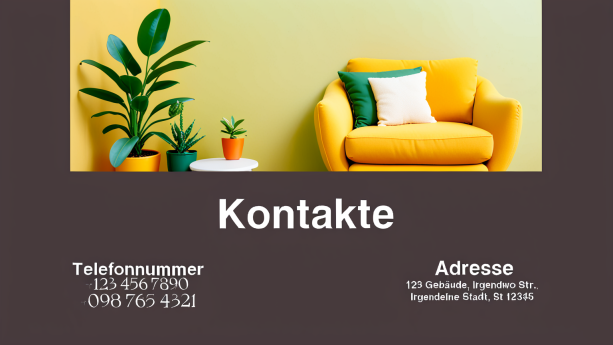}
\vspace{-3mm}
\end{subfigure}
\hfill
\begin{subfigure}[b]{0.32\textwidth}
\centering
\includegraphics[width=1\textwidth]{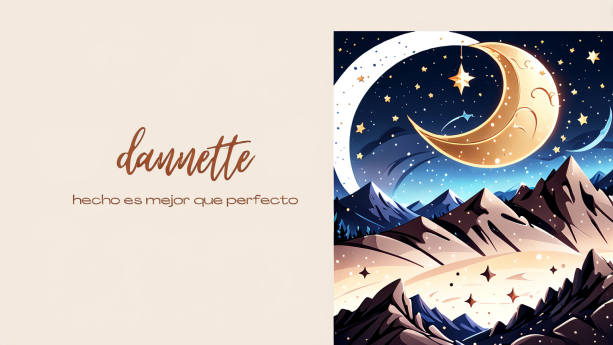}
\vspace{-3mm}
\end{subfigure}
\hfill
\begin{subfigure}[b]{0.32\textwidth}
\centering
\includegraphics[width=1\textwidth]{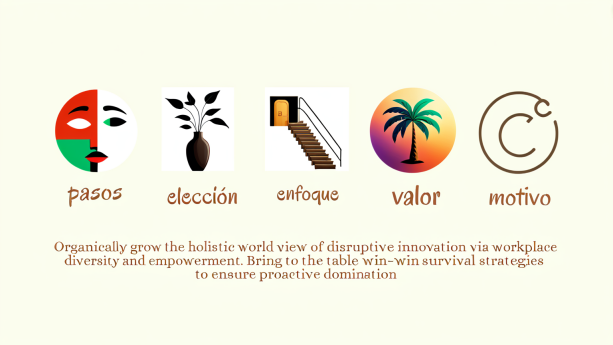}
\vspace{-3mm}
\end{subfigure}
\hfill
\begin{subfigure}[b]{0.32\textwidth}
\centering
\includegraphics[width=1\textwidth]{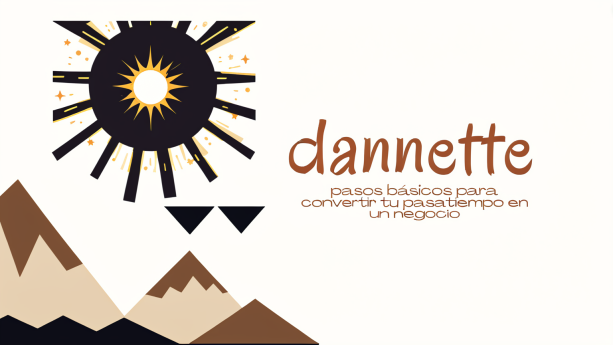}
\vspace{-3mm}
\end{subfigure}
\hfill
\begin{subfigure}[b]{0.32\textwidth}
\centering
\includegraphics[width=1\textwidth]{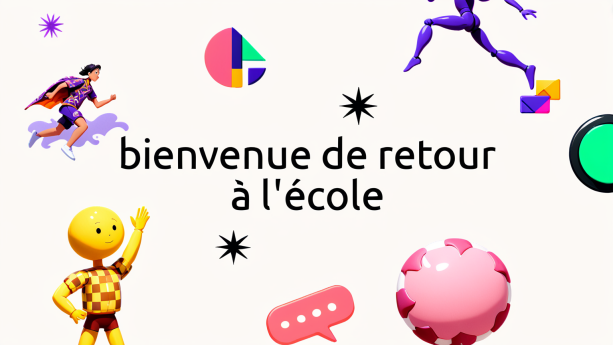}
\vspace{-3mm}
\end{subfigure}
\hfill
\begin{subfigure}[b]{0.32\textwidth}
\centering
\includegraphics[width=1\textwidth]{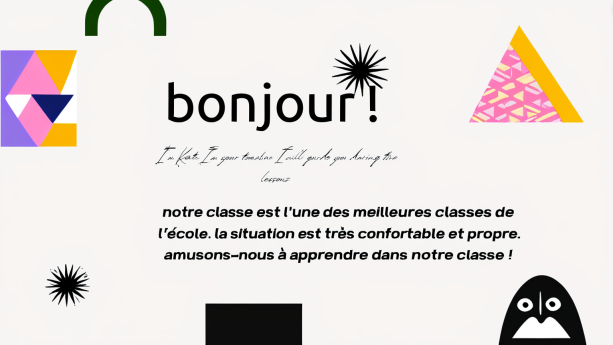}
\vspace{-3mm}
\end{subfigure}
\hfill
\begin{subfigure}[b]{0.32\textwidth}
\centering
\includegraphics[width=1\textwidth]{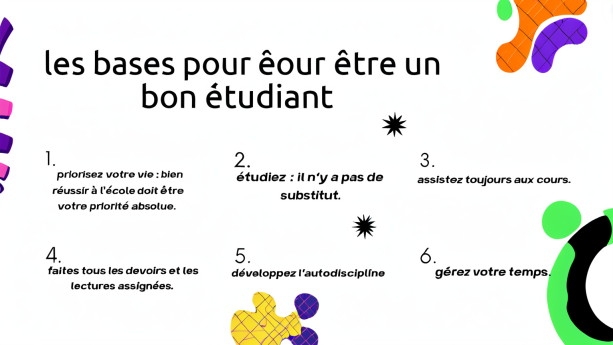}
\vspace{-3mm}
\end{subfigure}
\hfill
\begin{subfigure}[b]{0.32\textwidth}
\centering
\includegraphics[width=1\textwidth]{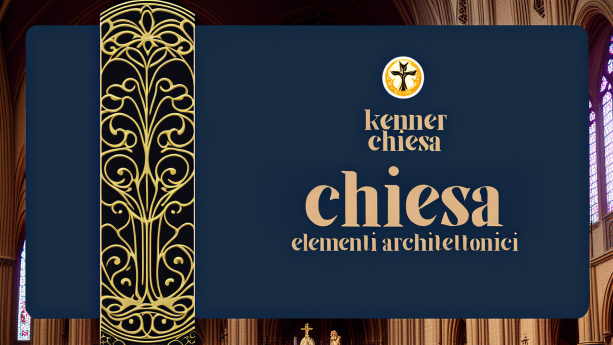}
\vspace{-3mm}
\end{subfigure}
\hfill
\begin{subfigure}[b]{0.32\textwidth}
\centering
\includegraphics[width=1\textwidth]{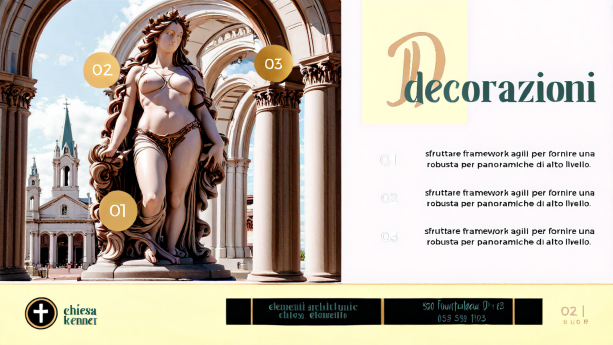}
\vspace{-3mm}
\end{subfigure}
\hfill
\begin{subfigure}[b]{0.32\textwidth}
\centering
\includegraphics[width=1\textwidth]{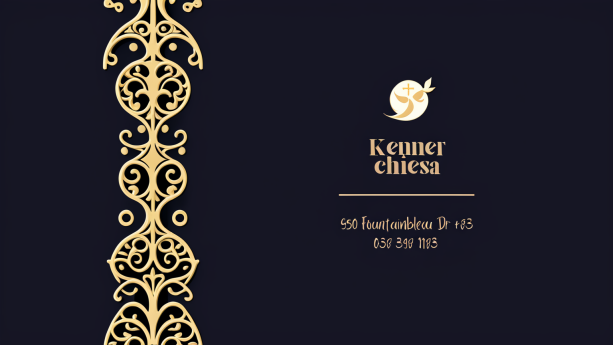}
\vspace{-3mm}
\end{subfigure}
\hfill
\begin{subfigure}[b]{0.32\textwidth}
\centering
\includegraphics[width=1\textwidth]{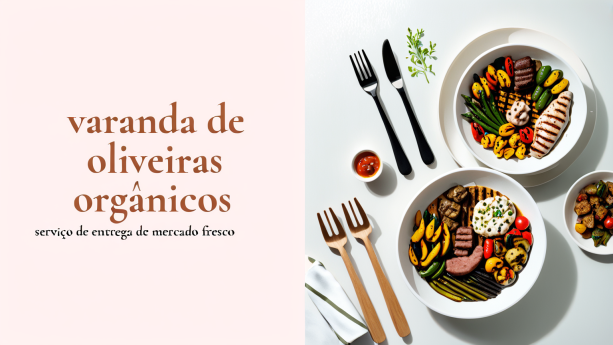}
\vspace{-3mm}
\end{subfigure}
\hfill
\begin{subfigure}[b]{0.32\textwidth}
\centering
\includegraphics[width=1\textwidth]{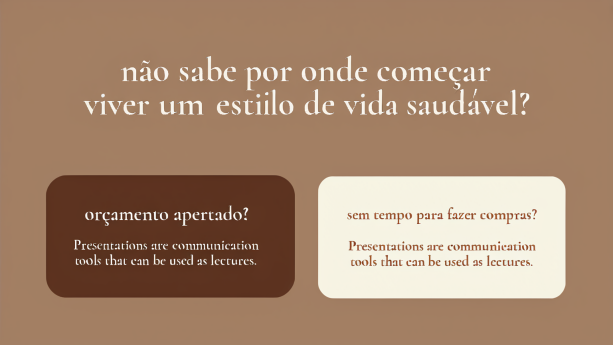}
\vspace{-3mm}
\end{subfigure}
\hfill
\begin{subfigure}[b]{0.32\textwidth}
\centering
\includegraphics[width=1\textwidth]{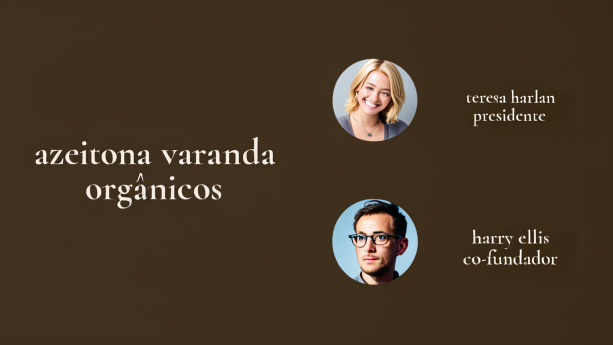}
\vspace{-3mm}
\end{subfigure}
\hfill
\begin{subfigure}[b]{0.32\textwidth}
\centering
\includegraphics[width=1\textwidth]{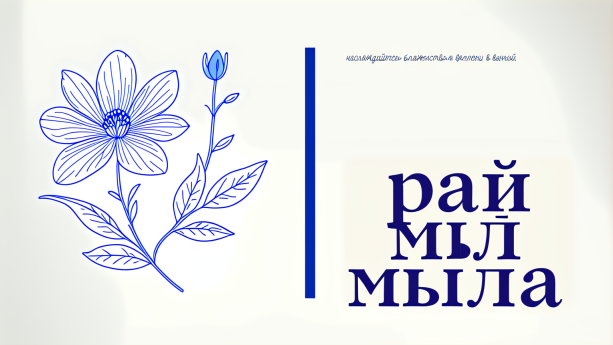}
\vspace{-3mm}
\end{subfigure}
\hfill
\begin{subfigure}[b]{0.32\textwidth}
\centering
\includegraphics[width=1\textwidth]{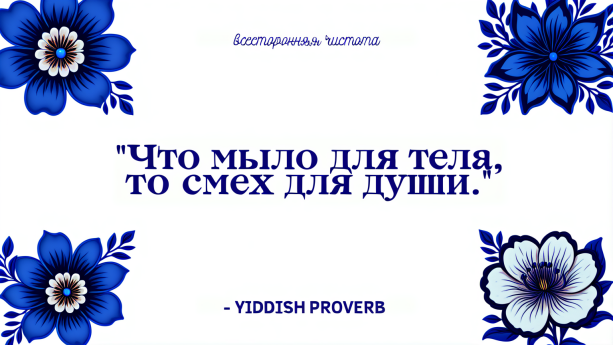}
\vspace{-3mm}
\end{subfigure}
\hfill
\begin{subfigure}[b]{0.32\textwidth}
\centering
\includegraphics[width=1\textwidth]{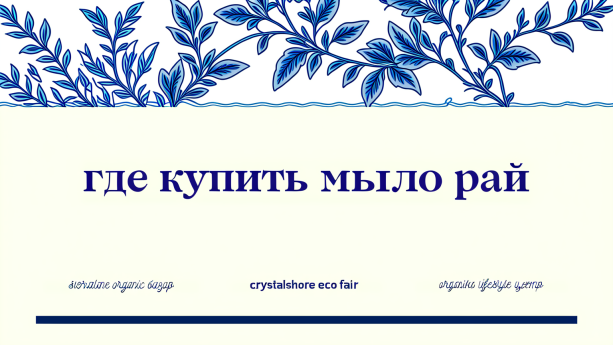}
\vspace{-3mm}
\end{subfigure}
\end{minipage}
\caption{\footnotesize{
\textbf{Qualitative results of multilingual slides generation}.We show the Chinese, Japanese, Korean, German, Spanish, French, Italian, Portuguese, and Russian visual text results in the nine rows subsequently.
}}
\label{fig:multilingual_slide}
\vspace{-5mm}
\end{figure}

\begin{figure}[t]
\centering
\includegraphics[width=1\linewidth]{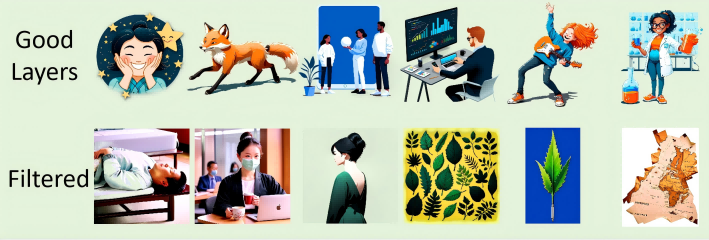}
\caption{\footnotesize{\textbf{Illustrating the transparent layers generated in the data engine}: The first row shows examples of the generated multi-style high-quality transparent layers, while the second row demonstrates the filtered ones.}}
\label{fig:transparent_layer_example}
\end{figure}

\begin{figure}[!t]
\begin{minipage}[!t]{0.95\linewidth}
\begin{subfigure}[b]{0.24\textwidth}
\centering
\includegraphics[width=1\textwidth]{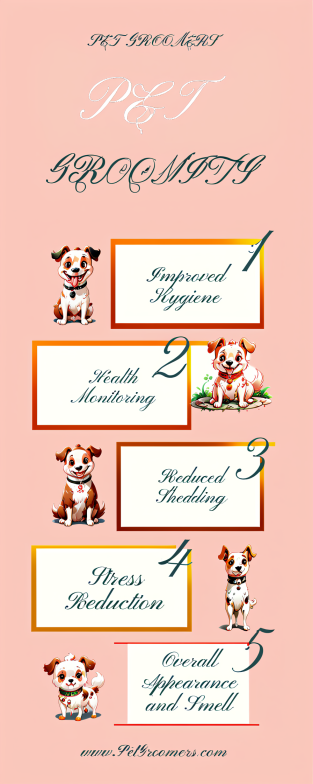}
\vspace{-3mm}
\end{subfigure}
\hfill
\begin{subfigure}[b]{0.24\textwidth}
\centering
\includegraphics[width=1\textwidth]{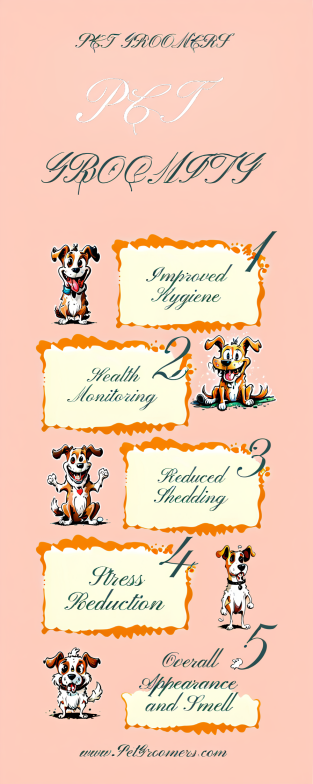}
\vspace{-3mm}
\end{subfigure}
\hfill
\begin{subfigure}[b]{0.24\textwidth}
\centering
\includegraphics[width=1\textwidth]{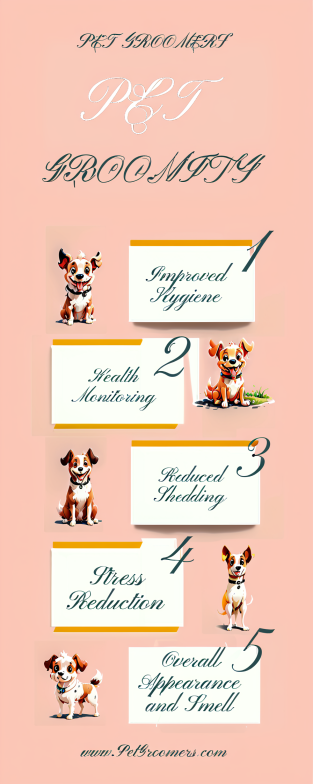}
\vspace{-3mm}
\end{subfigure}
\hfill
\begin{subfigure}[b]{0.24\textwidth}
\centering
\includegraphics[width=1\textwidth]{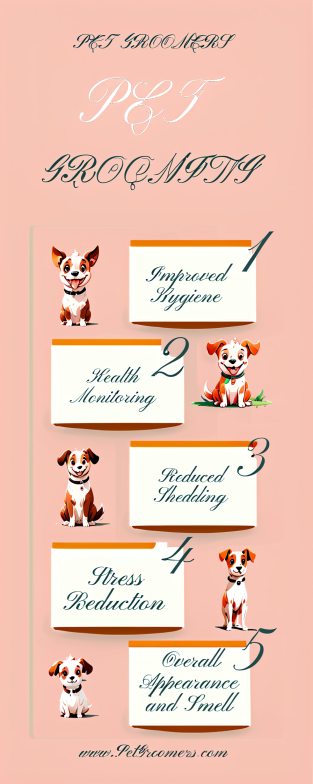}
\vspace{-3mm}
\end{subfigure}
\begin{subfigure}[b]{0.24\textwidth}
\centering
\includegraphics[width=1\textwidth]{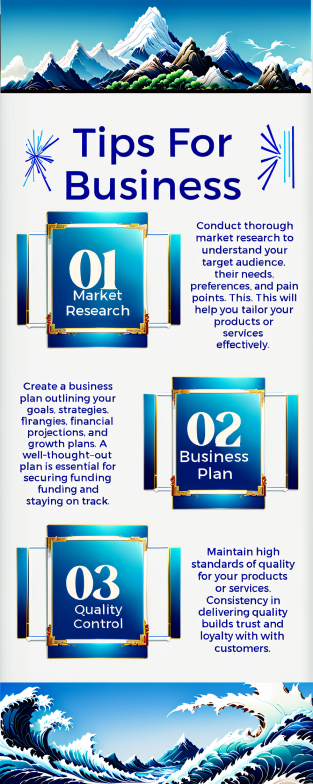}
\vspace{-3mm}
\end{subfigure}
\hfill
\begin{subfigure}[b]{0.24\textwidth}
\centering
\includegraphics[width=1\textwidth]{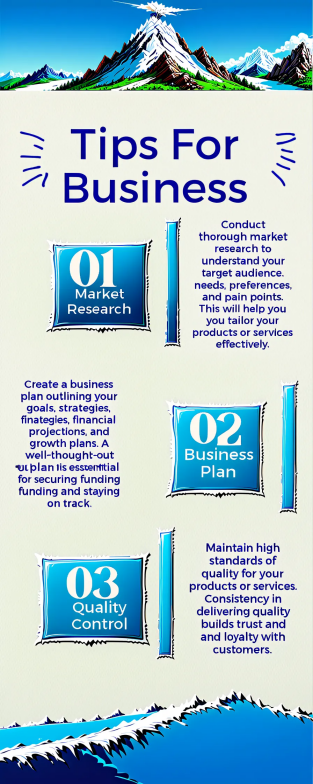}
\vspace{-3mm}
\end{subfigure}
\hfill
\begin{subfigure}[b]{0.24\textwidth}
\centering
\includegraphics[width=1\textwidth]{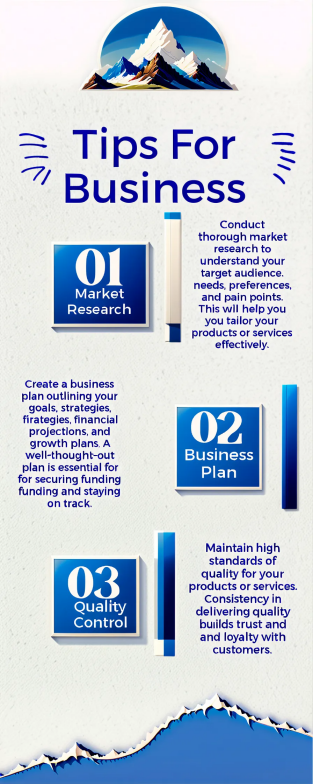}
\vspace{-3mm}
\end{subfigure}
\hfill
\begin{subfigure}[b]{0.24\textwidth}
\centering
\includegraphics[width=1\textwidth]{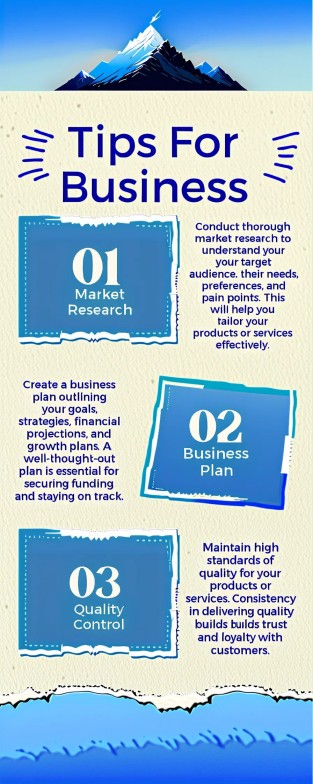}
\vspace{-3mm}
\end{subfigure}
\hfill
\begin{subfigure}[b]{0.24\textwidth}
\centering
\includegraphics[width=1\textwidth]{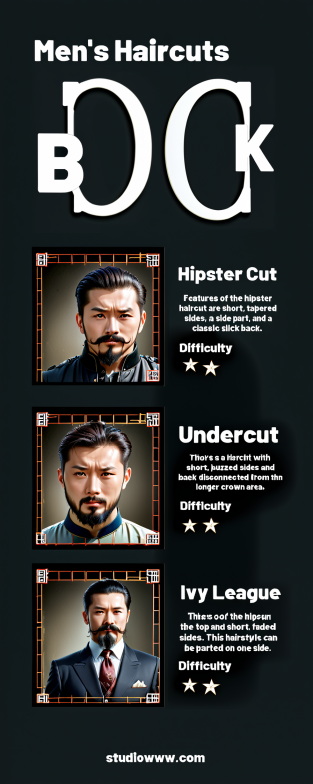}
\vspace{-3mm}
\end{subfigure}
\hfill
\begin{subfigure}[b]{0.24\textwidth}
\centering
\includegraphics[width=1\textwidth]{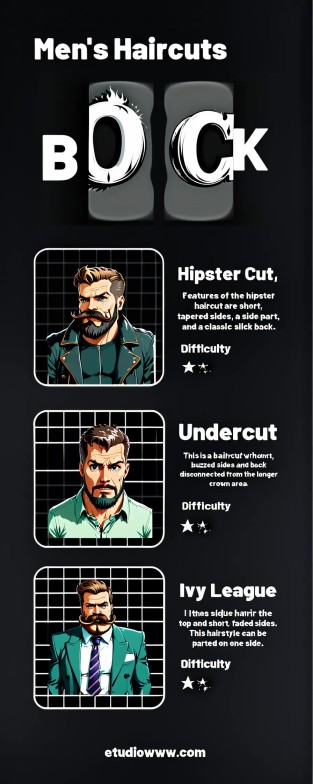}
\vspace{-3mm}
\end{subfigure}
\hfill
\begin{subfigure}[b]{0.24\textwidth}
\centering
\includegraphics[width=1\textwidth]{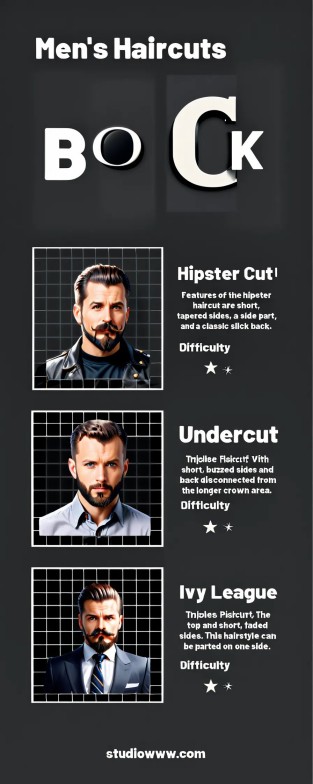}
\vspace{-3mm}
\end{subfigure}
\hfill
\begin{subfigure}[b]{0.24\textwidth}
\centering
\includegraphics[width=1\textwidth]{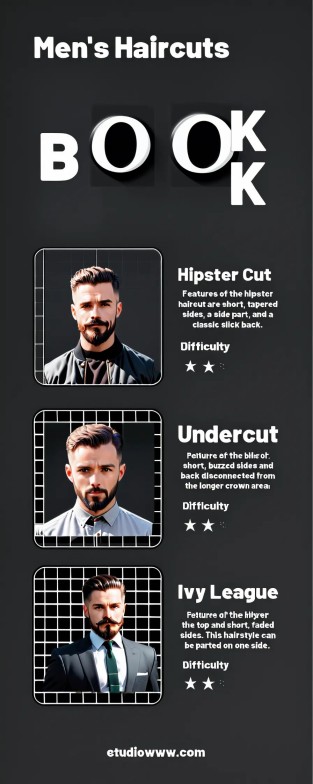}
\vspace{-3mm}
\end{subfigure}
\hfill
\begin{subfigure}[b]{0.24\textwidth}
\centering
\includegraphics[width=1\textwidth]{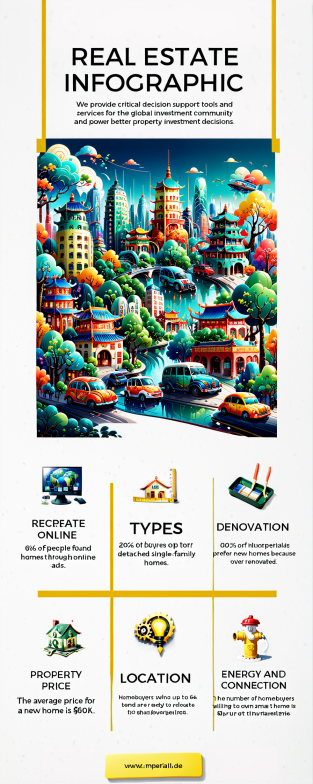}
\vspace{-3mm}
\caption*{\footnotesize{Chinese}}
\end{subfigure}
\hfill
\begin{subfigure}[b]{0.24\textwidth}
\centering
\includegraphics[width=1\textwidth]{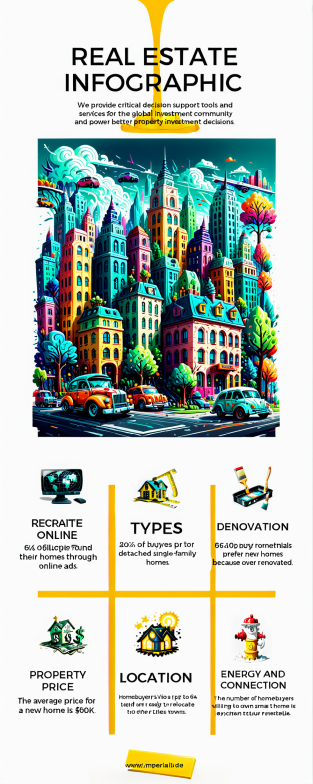}
\vspace{-3mm}
\caption*{\footnotesize{Comic}}
\end{subfigure}
\hfill
\begin{subfigure}[b]{0.24\textwidth}
\centering
\includegraphics[width=1\textwidth]{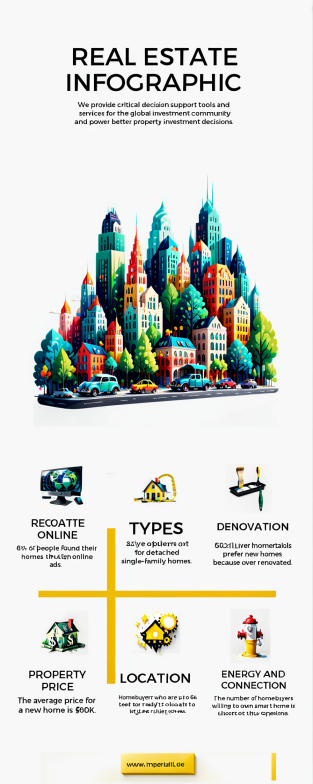}
\vspace{-3mm}
\caption*{\footnotesize{Illustration}}
\end{subfigure}
\hfill
\begin{subfigure}[b]{0.24\textwidth}
\centering
\includegraphics[width=1\textwidth]{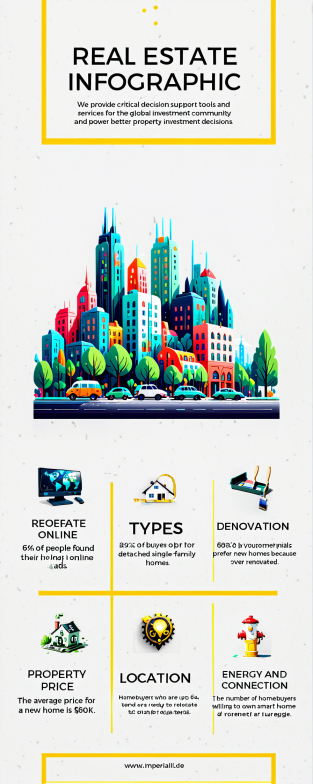}
\vspace{-3mm}
\caption*{\footnotesize{Minimalism}}
\end{subfigure}
\end{minipage}
\caption{\footnotesize{
\textbf{Qualitative results of multi-style infographics generation}. The four columns respectively attribute to four different styles: Chinese, Comic, Illustration and Minimalism.
}}
\label{fig:multi_style}
\vspace{-5mm}
\end{figure}

\begin{figure*}[t]
\centering
\includegraphics[width=0.86\linewidth]{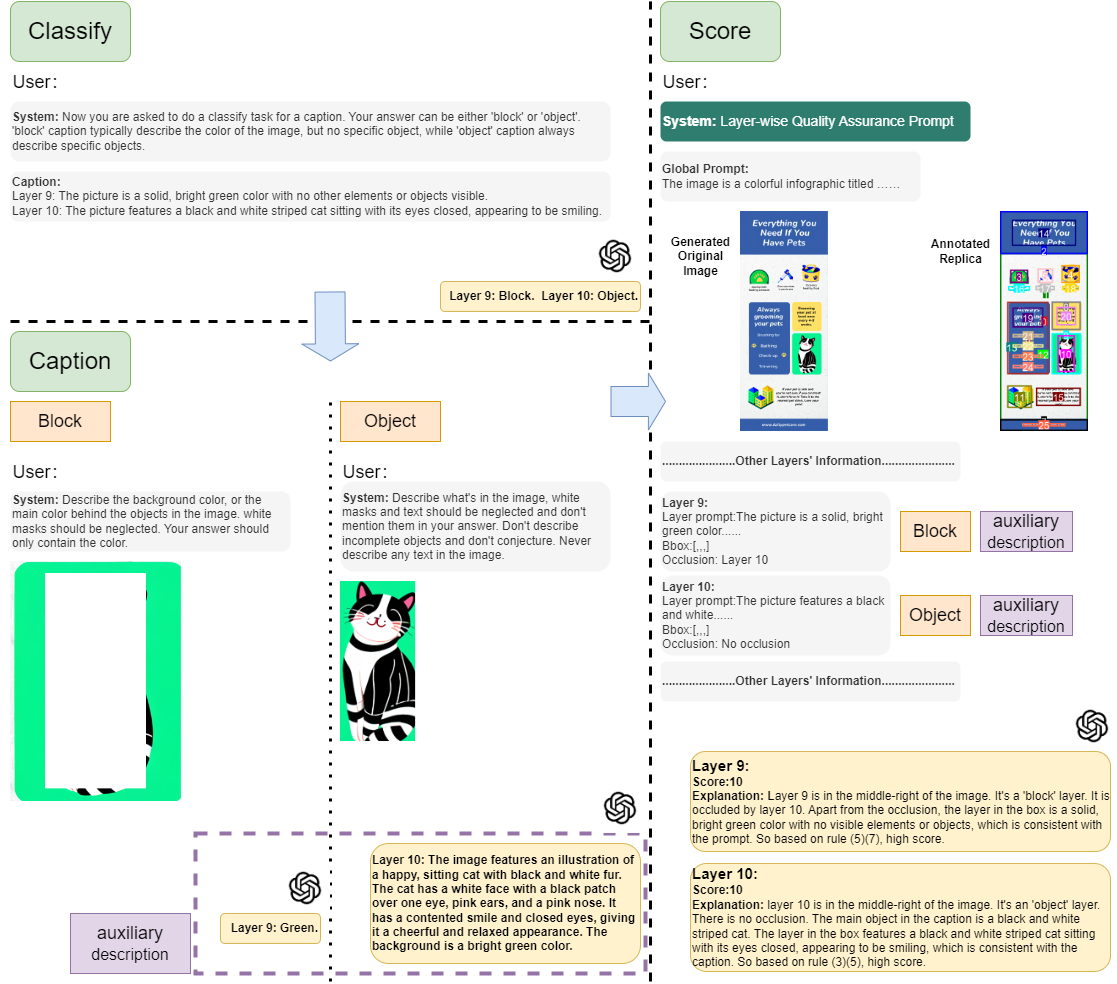}
\caption{\footnotesize{\textbf{Example of LGSR Assessment Pipeline}: We give the examples of two layers go through our LGSR assessment pipeline, which cover the two layer types: ``object'' and ``block''. We 
 demonstrate the different information fed to GPT-4o in every step, and highlight the response by GPT-4o, including its explanation.}}
\label{fig:LGSR_example}
\end{figure*}

In Tables~\ref{tab:ocr_comparison_results} and ~\ref{tab:numerical_quality_comparison_results}, we have already shown that \ourapproach outperforms state-of-the-art methods in almost every metric for slide generation. Figure~\ref{fig:comparison_slide} provides some qualitative comparison results. An interesting observation is that state-of-the-art methods such as FLUX may sometimes generate only a background instead of a reasonable slide page (columns 2, 5, and 6), which is likely related to their training data.

\section{Ablation on Layout Conditional CFG Timestep Range}
\label{sec:Appendix_lcfg_timestep}
\begin{figure*}[t]
\centering
\includegraphics[width=0.86\linewidth]{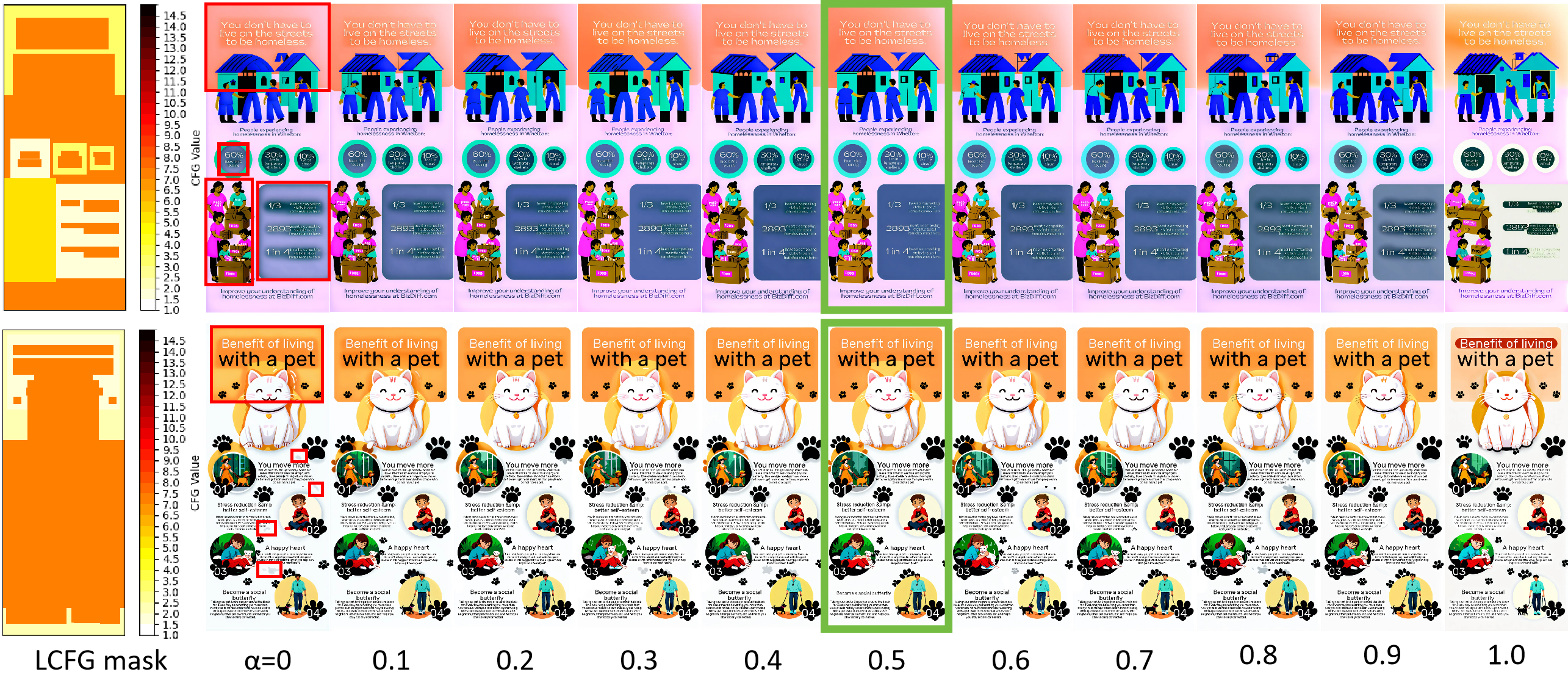}
\caption{\footnotesize{\textbf{Effect of different choices of $\alpha$ for LCFG}: We use red boxes to mark the artifacts in the images generated without layout conditional CFG and use green boxes to highlight the most aesthetically flawless images across all the others generated with different $\alpha$.}}
\label{fig:LCFG_timestep}
\end{figure*}

We demonstrate the effect of choosing different $\alpha$ values for layout conditional CFG in Figure~\ref{fig:LCFG_timestep}. Here, $\alpha$ is the hyperparameter that controls the starting timestep at which layout conditional CFG takes effect in the denoising process, as introduced in Section~\ref{sec:approach_3}. Our goal is to eliminate local artifacts in specific regions without deteriorating others; therefore, we assume that LCFG should be used when $t$ is close to $0$, given that details are settled in the later stages of the diffusion process. Thus, we only alter the starting timestep and apply LCFG toward the end.

From Figure~\ref{fig:LCFG_timestep} we can see that a small $\alpha$ may bring too little changes that are not enough to remove the flaws, while a big $\alpha$ can bring unwanted changes to other regions. Taking the first row as an example, when $\alpha$ is $0.1$ or $0.2$, the shadows and irregular color still exist; when $\alpha$ is $0.8$ or $0.9$, the navy blue block and the light blue rings begin to fade. To strike the balance, we empirically find that $0.5$ is the optimal choice for $\alpha$ in most cases.

\section{Detailed Prompt List}
\label{sec:Appendix_promptlist}
We illustrate the detailed prompts for generated infographics and slides shown in Figure~\ref{fig:teaser}, Figure~\ref{fig:sota} and Figure~\ref{fig:slides} in Table~\ref{tab:prompt_list}. For typesetting convenience, we only list the global prompts and all the text layer prompts.
\onecolumn
\begin{longtable}{l|>{\centering\arraybackslash}m{15cm}} 

\textbf{Image} & \textbf{Prompt} \\ \hline
\endfirsthead

\textbf{Image} & \textbf{Prompt} \\ \hline
\endhead

\multicolumn{2}{r}{\textit{Continued on next page...}} \\\hline
\endfoot
\caption{\footnotesize{{Detailed prompt for generated infographics and slides in Figure 1, Figure 10 and Figure 11.}}} \label{tab:prompt_list} \\
\endlastfoot

\scriptsize{Fig 1, Col1} & 
\scriptsize{Global: The image is a digital graphic with a blue background and a yellow border. At the top, in large white letters, the text reads 'Business Agency.' Below this title, there are three sections, each with a white background and a yellow dot, containing text and icons.The first section is titled 'Services Offered' and includes a brief description: 'Our comprehensive range of services is designed to address your diverse needs and support your business growth.'The second section is titled 'Team Expertise' and highlights: 'bring a diverse range of expertise to the table. We are passionate about collaborating with you to achieve your goals.'The third section is titled 'Successful Projects' and mentions: 'Our commitment to excellence and innovative problem-solving shines through in every endeavor.'At the bottom of the image, there is a stylized illustration depicting a group of people engaged in a meeting or presentation. The individuals are shown with various expressions and postures, suggesting a dynamic and collaborative environment. The central figure is gesturing towards a chart or graph, likely representing data or progress. The overall style of the image is clean, modern, and professional, with a focus on conveying the agency's capabilities and values.
Text: Text "Business" in $<$color-0$>$, $<$en-font-421$>$.  Text "Agency" in $<$color-5$>$, $<$en-font-371$>$.  Text "Services Offered" in $<$color-5$>$, $<$en-font-421$>$.  Text "Our comprehensive range of services is designed to address your diverse needs and support your business growth. " in $<$color-0$>$, $<$en-font-403$>$.  Text "Team Expertise" in $<$color-5$>$, $<$en-font-421$>$.  Text "bring a diverse range of expertise to the table. We are passionate about collaborating with you to achieve your goals." in $<$color-0$>$, $<$en-font-403$>$.  Text "Successful Projects" in $<$color-5$>$, $<$en-font-421$>$.  Text "Our commitment to excellence and innovative problem-solving shines through in every endeavor." in $<$color-0$>$, $<$en-font-403$>$. } \\ 
\hline
\scriptsize{Fig 1, Col2} & 
\scriptsize{Global: The image is an infographic titled 'Benefits of living with Pet'. It features a cartoon illustration of a person holding a cat. The person is smiling and appears to be enjoying the company of the cat. The cat is white with black stripes and is being held gently.Below the illustration, there are four numbered points, each describing a different benefit of living with a pet. The points are as follows:1. 'Increase your social interactions. Having a friend can make it easier to start up a conversation.'2. 'Help with your fitness. Having a pet will, of course, help to increase your energy levels and keep you active.'3. 'Help with loneliness. Having a pet in our side when we’re out and about can make us feel less alone.'4. 'Reduce stress and anxiety. Spending time with a pet can help boost our mental health.'Each point is accompanied by a small illustration of a cat, reinforcing the theme of the infographic. The cats in the illustrations are depicted in various poses and colors, adding visual interest to the text. The overall style of the image is friendly and informative, aimed at highlighting the positive effects of pet ownership on social, physical, emotional, and mental well-being.
Text: Text "Having a friend can make it easier to start up a conversation. " in $<$color-1$>$, $<$en-font-0$>$.  Text "Increase your social interactions" in $<$color-1$>$, $<$en-font-316$>$.  Text "Having a pet will, of course, help to increase your energy levels and keep you active." in $<$color-1$>$, $<$en-font-0$>$.  Text "1" in $<$color-0$>$, $<$en-font-316$>$.  Text "Benefits of living with a        " in $<$color-1$>$, $<$en-font-316$>$.  Text "Pet" in $<$color-1$>$, $<$en-font-316$>$.  Text "2" in $<$color-0$>$, $<$en-font-316$>$.  Text "3" in $<$color-0$>$, $<$en-font-316$>$.  Text "4" in $<$color-0$>$, $<$en-font-316$>$.  Text "Help with your fitness" in $<$color-1$>$, $<$en-font-316$>$.  Text "Having a pet in our side when we’re out and about can make us feel less alone." in $<$color-1$>$, $<$en-font-0$>$.  Text "Help with loneliness" in $<$color-1$>$, $<$en-font-316$>$.  Text "Spending time with a pet can help boost our mental health." in $<$color-1$>$, $<$en-font-0$>$.  Text "Reduce stress and anxiety" in $<$color-1$>$, $<$en-font-316$>$.  } \\ 
\hline
\scriptsize{Fig 1, Col3} & 
\scriptsize{Global: The image is a graphic with a blue background and white text, divided into four sections, each with a different title and accompanying illustration. The title at the top reads 'Tax Planning Strategies' in bold, capital letters.The first section is titled 'Understand Your Tax Bracket' and features an illustration of a target with an arrow pointing at it. The text below the title explains 'By understanding which brackets you fall into, you can make informed decisions about income deferral or acceleration.'The second section is titled 'Maximize Tax-Advantaged Accounts' and shows an illustration of a computer monitor displaying a graph with two lines, one in blue and the other in yellow. The accompanying text explains 'These accounts offer tax deductions, tax-free growth, or tax-free withdrawals for specific purposes.'The third section is titled 'Tax Loss Harvesting' and depicts a hand holding a coin with a dollar sign on it. The text below the title explains 'Offset capital gains by strategically selling investments that have experienced losses.'The fourth and final section is titled 'Plan for Charitable Giving' and features an illustration of two hands shaking, one holding a blue envelope and the other holding a yellow envelope. The text below the title suggests 'Charitable donations can be tax-deductible, so consider making strategic contributions to causes you care about.'Each section is separated by a horizontal line, and the overall style of the image is informational and educational, designed to provide advice on tax planning strategies.
Text: Text "Tax Planning Strategies" in $<$color-0$>$, $<$en-font-78$>$.  Text "By understanding which brackets you fall into, you can make informed decisions about income deferral or acceleration." in $<$color-1$>$, $<$en-font-15$>$.  Text "Understand Your Tax Bracket" in $<$color-0$>$, $<$en-font-15$>$.  Text "These accounts offer tax deductions, tax-free growth, or tax-free withdrawals for specific purposes." in $<$color-1$>$, $<$en-font-15$>$.  Text "Maximize Tax-Advantaged Accounts" in $<$color-0$>$, $<$en-font-15$>$.  Text "Offset capital gains by strategically selling investments that have experienced losses. " in $<$color-1$>$, $<$en-font-15$>$.  Text "Tax Loss Harvesting" in $<$color-0$>$, $<$en-font-15$>$.  Text "Charitable donations can be tax-deductible, so consider making strategic contributions to causes you care about. " in $<$color-1$>$, $<$en-font-15$>$.  Text "Plan for Charitable Giving" in $<$color-0$>$, $<$en-font-15$>$.  } \\
\hline
\scriptsize{Fig 1, Col4} & 
\scriptsize{Global: The image is a graphic with a warm, earthy color palette, primarily in shades of brown and beige. It features a series of circular icons with illustrations of a dog, each accompanied by text that provides tips on how to train a dog. The tips are numbered from 1 to 5 and are presented in a clear, instructional manner.The first tip, '1. One Goal at A Time,' suggests focusing on one behavior at a time during training, explaining 'Potty training, Play dead oe Shake hand whatever it is start out with only one goal in mind.' The second tip, '2. Use A Short Cue Word,' advises 'Use short cue word while training the dog, long sentence only lead to confusion.' The third tip, '3. Use Treats,' emphasizes 'Treats motivate the dog to do the correct action and reward treats for every good behavior.' The fourth tip, '4. Practice, Practice, and Practice,' highlights that 'Repetition is key in perfecting a dog trict keep repeating it until your get hand of it.' The fifth tip, '5. Take it Slow,' suggests 'After practice, give youe dog the much needed rest it deserves, to make him remember the trick.'The text is written in a simple, sans-serif font, and the overall style of the image is clean and modern. The dog in the illustrations is a Corgi, characterized by its short legs and long body, and is depicted in various poses that correspond to the training tips. The background is plain and does not distract from the content. The image includes the URL 'www.yourwebsite.com' at the bottom and is titled 'How to Train Your Dog?'
Text: Text "1. One Goal at A Time" in $<$color-3$>$, $<$en-font-458$>$.  Text "Potty training, Play dead oe Shake hand whatever it is start out with only one goal in mind." in $<$color-3$>$, $<$en-font-0$>$.  Text "2. Use A Short Cue Word" in $<$color-3$>$, $<$en-font-458$>$.  Text "Use short cue word while training the dog, long sentence only lead to confusion." in $<$color-3$>$, $<$en-font-0$>$.  Text "3. Use Treats" in $<$color-3$>$, $<$en-font-458$>$.  Text "Treats motivate the dog to do the correct action and reward treats for every good behavior." in $<$color-3$>$, $<$en-font-0$>$.  Text "4. Practice, Practice, and Practice" in $<$color-3$>$, $<$en-font-458$>$.  Text "Repetition is key in perfecting a dog trict keep repeating it until your get hand of it." in $<$color-3$>$, $<$en-font-0$>$.  Text "5. Take it Slow" in $<$color-3$>$, $<$en-font-458$>$.  Text "After practice, give youe dog the much needed rest it deserves, to make him remember the trick." in $<$color-3$>$, $<$en-font-0$>$.  Text "www.yourwebsite.com" in $<$color-86$>$, $<$en-font-458$>$.  Text "How to Train Your Dog?" in $<$color-86$>$, $<$en-font-458$>$.  } \\
\hline
\scriptsize{Fig 1, Col5} & 
\scriptsize{Global: The image is an infographic titled 'MOVING TIPS' with a light blue background and a white border. It features four illustrated tips for moving, each accompanied by a brief description. The tips are as follows:1. '1.Create a Moving Budget': 'Before moving into a new apartment, create a budget to ensure you have enough money saved for the move. Include deposits, rental fees, moving costs, and any necessary furniture.'2. '2.Ask About Utilities': 'Ask your landlord about the utilities that are included in your rent. Make sure you understand which ones are included and which ones you'll have to pay for yourself.'3. '3.Pack Smart': 'Be sure to pack everything in the right boxes and label them appropriately. You'll thank yourself when you're unpacking your stuff in the new apartment.'4. '4.Make a Plan For today': 'Have a plan before moving day. Make sure you have enough help and the right tools to move all of your stuff safely and efficiently.'In the background, there are two illustrated characters who appear to be in the process of moving. One is carrying a box, while the other is holding a plant. They are standing in front of a moving truck, which is parked next to a stack of boxes.The infographic also includes a website address at the bottom: 'www.reallygoodsite.com'.The overall style of the image is clean and modern, with a clear focus on providing practical advice for those preparing to move.
Text: Text "moving" in $<$color-4$>$, $<$en-font-247$>$.  Text "tips" in $<$color-27$>$, $<$en-font-247$>$.  Text "Before moving into a new apartment, create a budget to ensure you have enough money saved for the move. Include deposits, rental fees, moving costs, and any necessary furniture." in $<$color-1$>$, $<$en-font-326$>$.  Text "1.Create a Moving Budget" in $<$color-4$>$, $<$en-font-247$>$.  Text "Ask your landlord about the utilities that are included in your rent. Make sure you understand which ones are included and which ones you'll have to pay for yourself." in $<$color-1$>$, $<$en-font-326$>$.  Text "2.Ask About Utilities" in $<$color-4$>$, $<$en-font-247$>$.  Text "Be sure to pack everything in the right boxes and label them appropriately. You'll thank yourself when you're unpacking your stuff in the new apartment" in $<$color-1$>$, $<$en-font-326$>$.  Text "3.Pack Smart" in $<$color-4$>$, $<$en-font-247$>$.  Text "Have a plan before moving day. Make sure you have enough help and the right tools to move all of your stuff safely and efficiently." in $<$color-1$>$, $<$en-font-326$>$.  Text "4.Make a Plan For today" in $<$color-4$>$, $<$en-font-247$>$.  Text "www.reallygoodsite.com" in $<$color-4$>$, $<$en-font-247$>$.  } \\
\hline
\scriptsize{Fig 10, Col1} & 
\scriptsize{Global: The image is a digital graphic with a dark brown background. At the top, in large white letters, the text reads 'HOW TO TAME PETS'. Below this title, there are four separate sections, each with a heart icon and a piece of advice on how to tame pets. The first section suggests, 'Approach the animal calmly and slowly, using gentle body language to convey that you are not a threat'. The second section advises, 'Offer the animal treats or food to establish a positive association and encourage them to trust you'. The third section recommends, 'Gradually introduce touch and physical contact, starting with gentle strokes or pats, respecting the animal's boundaries'. The fourth section advises, 'Consistently spend time with the animal, engaging in activities they enjoy, and providing a safe and build a bond of trust'.At the bottom of the image, there is a photograph of a small white dog with a black nose and dark eyes. The dog appears to be a French Bulldog. The dog's expression is joyful, with its mouth open and ears perked up, and it is looking directly at the camera. The photograph is framed by a light beige border with a thin brown line.
 Text: Text "how to tame pets" in $<$color-81$>$, $<$en-font-403$>$.  Text "Offer the animal treats or food to establish a positive association and encourage them to trust you" in $<$color-2$>$, $<$en-font-403$>$.  Text "Approach the animal calmly and slowly, using gentle body language to convey that you are not a threat" in $<$color-81$>$, $<$en-font-403$>$.  Text "Consistently spend time with the animal, engaging in activities they enjoy, and providing a safe and build a bond of trust" in $<$color-81$>$, $<$en-font-403$>$.  Text "Gradually introduce touch and physical contact, starting with gentle strokes or pats, respecting the animal's boundaries" in $<$color-2$>$, $<$en-font-403$>$. } \\
\hline
\scriptsize{Fig 10, Col2} & 
\scriptsize{Global: The image is a graphic representation of a consulting process overview. The title 'CONSULTING PROCESS OVERVIEW' is displayed prominently at the top in bold, black letters within a bordered box. Below the title, there is a subtitle that reads, 'The Consulting Process Overview provides a concise yet comprehensive view of the consulting journey.'The infographic is divided into four distinct stages, each represented by a different section:1. 'Discovery and Research': This stage is depicted with an icon of a computer monitor and light bulbs, suggesting the process of gathering information and analyzing data.2. 'Design and Visualization': This stage is illustrated with a design symbol, indicating the idea generation and conceptualization phase.3. 'Review and Refinement': This stage is represented with a review icon, implying the process of reviewing and refining the design or concept.4. 'Finalization and Delivery': This stage is shown with a delivery icon, representing the finalization of the project and the delivery of the completed work.Each stage is visually distinguished with a different background color and icon, making it easy to follow the sequence of the consulting process. The overall style of the image is clean and modern, with a clear and straightforward layout that is easy to understand.
Text: Text "The Consulting Process Overview provides a concise yet comprehensive view of the consulting journey" in $<$color-2$>$, $<$en-font-216$>$.  Text " Design and Visualization" in $<$color-1$>$, $<$en-font-50$>$.  Text "Review and Refinement" in $<$color-20$>$, $<$en-font-50$>$.  Text "Finalization and Delivery" in $<$color-1$>$, $<$en-font-50$>$.  Text "Consulting Process Overview" in $<$color-1$>$, $<$en-font-216$>$.  Text "Discovery and Research" in $<$color-20$>$, $<$en-font-50$>$.  } \\
\hline
\scriptsize{Fig 10, Col3} & 
\scriptsize{Global: The image is a vibrant infographic that outlines the benefits of pet hotels. It's designed to be visually appealing with a pink background and a playful, cartoon-like style. The infographic is divided into sections, each highlighting a different benefit.At the top, the title 'Benefits of Pet Hotels' is prominently displayed in bold, purple letters. Just below the title, there's a cute illustration of a cat peeking out from a suitcase, symbolizing the pet-friendly nature of the hotels.The first section, titled 'Professional Care:', features an illustration of a cat being groomed by a professional. This suggests that pet hotels offer grooming services.The second section, 'Safety and Security:', shows an illustration of a cat in a safe, secure environment, possibly indicating that pet hotels provide a safe haven for pets.The third section, 'Exercise and Playtime:', depicts a cat playing with a toy, implying that pet hotels offer opportunities for pets to engage in play and exercise.The fourth section, 'Professional Grooming Services:', features another illustration of a cat being groomed, reinforcing the point made in the first section.The final section, 'Peace of Mind for Pet Owners:', shows a cat sitting contentedly next to a bottle of water, which might symbolize the care and attention given to pets at these hotels, providing pet owners with peace of mind.At the bottom of the infographic, there's a website address 'www.Pethotel.com,' suggesting that this is the website for the pet hotels being advertised.Overall, the infographic is a colorful and engaging way to present the advantages of pet hotels to potential customers.
Text: Text "Benefits" in $<$color-83$>$, $<$en-font-179$>$.  Text "of Pet Hotels" in $<$color-83$>$, $<$en-font-179$>$.  Text "www.Pethotel.com" in $<$color-83$>$, $<$en-font-71$>$.  Text "Peace of Mind for Pet Owners: " in $<$color-0$>$, $<$en-font-71$>$.  Text "Safety and Security:" in $<$color-0$>$, $<$en-font-71$>$.  Text "Professional Grooming Services:" in $<$color-0$>$, $<$en-font-71$>$.  Text "Professional Care:" in $<$color-0$>$, $<$en-font-71$>$.  Text "Exercise and Playtime:" in $<$color-0$>$, $<$en-font-71$>$.  } \\
\hline
\scriptsize{Fig 10, Col4} & 
\scriptsize{Global: The image is a vibrant orange poster with a white background. It features two illustrations of a fluffy brown and white dog, which appears to be a corgi, sitting on a yellow background. The dog is depicted in a relaxed pose, looking directly at the viewer.At the top of the poster, there is a bold, white text that reads 'Let's Give Our Pet a New Home!'. Below this, there are two pieces of text in a smaller font. The first one states 'Fluffy buddy still needs new adopters to bring them to become human new homie!' and the second one says 'Around 60\% of volunteers have already adopted fluffy buddies into their new home!'At the bottom of the poster, there is a call to action with a white text that reads 'Are you ready to adopt your new homie or give donations? Contact us on:' followed by 'Call:+123 44 55 66 77', 'Visit:www.yoursite.com', and 'E-mail:youremail@gmail.com'.The overall design of the poster is simple and clear, with a focus on the message of adopting pets and the contact information provided.
Text: Text "Fluffy buddy still needs new adopters to bring them to become human new homie!" in $<$color-14$>$, $<$en-font-0$>$.  Text "Around 60\% of volunteers have already adopted fluffy buddies into their new home!" in $<$color-14$>$, $<$en-font-0$>$.  Text "Are you ready to adopt your new homie or give donations? Contact us on:" in $<$color-0$>$, $<$en-font-0$>$.  Text "Call:+123 44 55 66 77" in $<$color-0$>$, $<$en-font-0$>$.  Text "Visit:www.yoursite.com" in $<$color-0$>$, $<$en-font-0$>$.  Text "E-mail:youremail@gmail.com" in $<$color-0$>$, $<$en-font-0$>$.  Text "Let's Give Our Pet a New Home!" in $<$color-0$>$, $<$en-font-0$>$.  } \\
\hline
\scriptsize{Fig 10, Col5} & 
\scriptsize{Global: The image is a digital graphic with a warm, earthy color scheme, primarily in shades of orange and brown. It features a textured background that resembles a sandy beach or a textured surface. At the top, there is a bold title that reads '4 Ways to Create' and 'FINANCIAL PLAN' in capital letters, with the word 'FINANCIAL PLAN' being the most prominent.Below the title, there is a paragraph of text that explains that 'Business financial health can be determined by implementing a financial plan. Prepare your financial plan by following the steps below.'The graphic is divided into four sections, each representing a step in creating a financial plan. Each section is labeled with a number from 1 to 4 and includes a circular icon with a different color and symbol for each step. The icons are simple and abstract, with no specific objects or characters depicted.The first step, labeled '1', is titled 'Compare Your Goals' and includes an icon with a document and a magnifying glass, suggesting a focus on reviewing or analyzing goals. The description reads: 'Thinking about what your company wants to accomplish with a strategic plan helps. Look at numbers first, then consider what you need to achieve your goals.' The second step, '2', is titled 'Plan Your Finances Goals' and features an icon with a document and a pencil, indicating planning or writing. The description reads: 'Consider what it will cost to achieve your goals. Include various scenarios. Create a range to predict the impact of each optimistic and pessimistic scenario.' The third step, '3', is titled 'Contingency Planning' and shows an icon with gears and a document, which might imply strategizing or organizing. The description reads: 'Examine your cash flow statement and assets and formulate a plan for when the business encounters a crisis or when the cash flow is interrupted.' The fourth step, '4', is titled 'Compare Your Goals' again, with an icon that includes a document and a magnifying glass, similar to the first step. The description reads: 'Consider what it will cost to achieve your goals. Include various scenarios. Create a range to predict the impact of each optimistic and pessimistic scenario.'Each step has a brief description underneath the title, providing a brief explanation of the action to be taken. The text is clear and legible, and the overall layout is clean and organized, designed to be informative and easy to follow.
Text: Text "4 Ways to Create" in $<$color-2$>$, $<$en-font-400$>$.  Text "Financial Plan" in $<$color-1$>$, $<$en-font-485$>$.  Text "Business financial health can be determined byimplementing a financial plan.Prepare your financial plan by following the steps below." in $<$color-2$>$, $<$en-font-342$>$.  Text "Compare Your Goals" in $<$color-2$>$, $<$en-font-342$>$.  Text "1" in $<$color-2$>$, $<$en-font-342$>$.  Text "Thinking about what your company wants to accomplish with a strategic plan helps. Look at numbers first, then consider what you need to achieve your goals." in $<$color-2$>$, $<$en-font-342$>$.  Text "Plan Your Finances Goals" in $<$color-2$>$, $<$en-font-342$>$.  Text "2" in $<$color-2$>$, $<$en-font-342$>$.  Text "Consider what it will cost to achieve your goals. Include various scenarios. Create a range to predict the impact of each optimistic and pessimistic scenario." in $<$color-2$>$, $<$en-font-342$>$.  Text "Contingency Planning" in $<$color-2$>$, $<$en-font-342$>$.  Text "3" in $<$color-2$>$, $<$en-font-342$>$.  Text "Examine your cash flow statement and assets and formulate a plan for when the business encounters a crisis or when the cash flow is interrupted." in $<$color-2$>$, $<$en-font-342$>$.  Text "Compare Your Goals" in $<$color-2$>$, $<$en-font-342$>$.  Text "4" in $<$color-2$>$, $<$en-font-342$>$.  Text "Consider what it will cost to achieve your goals. Include various scenarios. Create a range to predict the impact of each optimistic and pessimistic scenario." in $<$color-2$>$, $<$en-font-342$>$.  } \\
\hline
\scriptsize{Fig 11, Row1, Col1} & 
\scriptsize{Global: The image shows a minimalist concept displayed on a wooden surface. There are three framed pieces of artwork or photographs. The first frame is gold, the second is white with a brown paper insert. To the right of the frames, there is a small potted plant. Below the frames, there are three brown leather-bound books stacked vertically. The overall style of the image is clean and modern, with a focus on simplicity and the use of natural materials. The background is a plain, light-colored wall, which enhances the minimalist aesthetic. There is a circular brown icon on the left side of the image with a white line drawing of a couch and picture frame, accompanied by the text 'Minimalist' and 'CONCEPT'.
Text: Text "Minimalist" in $<$color-0$>$, $<$en-font-57$>$.  Text "Concept" in $<$color-0$>$, $<$en-font-0$>$.  } \\
\hline
\scriptsize{Fig 11, Row1, Col2} & 
\scriptsize{Global: The image shows a modern bedroom interior with a focus on lighting. On the left side of the image, there is a bed with a white bedspread and a beige blanket. The bed is positioned against a white brick wall. To the right of the bed, there is a small wooden side table with a black floor lamp on it. On the right side of the image, there is a text overlay that reads 'Lighting'. Below this title, there is a suggestion for improving lighting in a room: 'use more natural lighting by installing glass windows and also use less lighting, such as study lamps'. The text is set against a brown background. At the bottom of the image, there is a credit line that says 'By Company Name'. The overall style of the image is clean and minimalistic, with a focus on interior design and lighting solutions.
Text: Text "Lighting" in $<$color-0$>$, $<$en-font-57$>$.  Text "use more natural lighting by installing glass windows and also use less lighting, such as study lamps" in $<$color-0$>$, $<$en-font-0$>$.  Text "By Company Name" in $<$color-0$>$, $<$en-font-0$>$.  } \\
\hline
\scriptsize{Fig 11, Row1, Col3} & 
\scriptsize{Global: The image depicts a room with a focus on interior decoration. On the wall, there is a large abstract painting with a mix of colors, predominantly in shades of beige, brown, and white, which gives it a textured appearance. Below the painting, there is a wooden desk with a few items on it, including a small plant in a white pot, a small figure, some paintbrushes in a clear container, and a stack of books. To the right of the desk, there is a small figure of a person, possibly a decorative statue or figurine.On the left side of the image, there is a text box with the word 'Decoration' in a bold, serif font. Below this title, there is a brief description that reads, 'Use abstract painting decorations with minimalist colors and plants and also a small figure.' At the bottom left, there is a text that reads, 'By Company Name'.The overall style of the image is clean and modern, with a focus on simplicity and the use of natural materials. The lighting in the room is soft, creating a warm and inviting atmosphere. The image appears to be a promotional or inspirational graphic, possibly for an interior design company or a lifestyle blog.
Text: Text "Decoration" in $<$color-7$>$, $<$en-font-57$>$.  Text "Use abstract painting decorations with minimalist colors and plants and also a small figure" in $<$color-7$>$, $<$en-font-0$>$.  Text "By Company Name" in $<$color-7$>$, $<$en-font-0$>$.  } \\
\hline
\scriptsize{Fig 11, Row2, Col1} & 
\scriptsize{Global: The image is a digital graphic with a combination of text and a photograph. The text at the top reads 'COMPANY NAME', followed by 'Crafting a Winning BUSINESS STRATEGY' in a large, bold font. The photograph shows a man standing in an office environment. He is wearing glasses, a suit, and a tie. Behind him, there is a large window that lets in natural light, and through the window, you can see a cityscape with buildings and trees. The overall style of the image is professional and corporate, likely intended for a business-related context.
Text: Text "business strategy" in $<$color-98$>$, $<$en-font-250$>$.  Text "Crafting a Winning" in $<$color-98$>$, $<$en-font-250$>$.  Text "company name" in $<$color-98$>$, $<$en-font-37$>$.  } \\
\hline
\scriptsize{Fig 11, Row2, Col2} & 
\scriptsize{Global: The image is a slide from a presentation, specifically an introduction slide. The slide is divided into two main sections. On the left side, there is a photograph of three individuals engaged in a business meeting. They are seated around a table, with papers and a calculator in front of them, suggesting they are discussing financial or strategic matters. The individuals appear to be focused on their work, with one person gesturing towards the papers.The left side of the slide also includes the text 'COMPANY NAME' and 'INTRODUCTION' in a large, bold font, accompanied by a graphic of a yellow sunburst.On the right side of the slide, there is text that serves as an introduction to the topic of the presentation. The text is in a sans-serif font and is written in a clear, readable font size. The text reads:'Greet the audience and introduce the topic of business strategy.Define business strategy as a roadmap to achieve long-term objectives.Explain that the presentation aims to explore the key components of an effective business strategy.'The text is aligned to the left and is set against a light green background, which contrasts with the darker green background of the left side of the slide. The overall layout of the slide is professional and designed to be informative and engaging for the audience.
Text: Text "introduction" in $<$color-98$>$, $<$en-font-250$>$.  Text "company name" in $<$color-98$>$, $<$en-font-37$>$.  Text "Greet the audience and introduce the topic of business strategy." in $<$color-4$>$, $<$en-font-37$>$.  Text "Define business strategy as a roadmap to achieve long-term objectives." in $<$color-4$>$, $<$en-font-37$>$.  Text "Explain that the presentation aims to explore the key components of an effective business strategy." in $<$color-4$>$, $<$en-font-37$>$.  } \\
\hline
\scriptsize{Fig 11, Row2, Col3} & 
\scriptsize{Global: The image is a composite of two separate sections, each with its own content.On the left side, there is a dark green background with light green text. The text at the top reads 'COMPANY NAME'. Below this, in a larger font size, is the phrase 'SECTION BREAK'. Underneath the section break title, there is a smaller text that says 'Explain the importance of allocating resources effectively to execute the strategy.'On the right side of the image, there is a photograph of a man sitting at a desk. He is wearing glasses and a suit, and he appears to be engaged in work, possibly reading or reviewing documents. The desk has a laptop, a cup that might contain a beverage, and some papers or documents. The background of the photograph shows a window with daylight coming through, suggesting an office environment.
Text: Text "section break" in $<$color-98$>$, $<$en-font-250$>$.  Text "company name" in $<$color-98$>$, $<$en-font-37$>$.  Text "Explain the importance of allocating resources effectively to execute the strategy." in $<$color-98$>$, $<$en-font-37$>$.  } \\
\hline
\scriptsize{Fig 11, Row3, Col1} & 
\scriptsize{Global: The image features a vibrant green vineyard with rows of grapevines stretching into the distance. The sun is low on the horizon, casting a warm, golden light over the scene, suggesting either sunrise or sunset. In the foreground, there is a graphic design element that resembles a book cover or a sign. This design includes a dark green background with ornate, swirling patterns in a lighter green. At the center, there is a logo with the text 'BOURGON' in a stylized font, and below it, the words 'Family Vineyard' are written in a larger, bold font. The phrase 'in Italy' is placed at the bottom of the design. At the very bottom, the website 'bourgon.it' is mentioned. The overall style of the image is a blend of a real-life photograph of the vineyard with a graphic overlay, possibly for promotional or informational purposes.
Text: Text "BOURGON" in $<$color-0$>$, $<$en-font-71$>$.  Text "bourgon.it" in $<$color-0$>$, $<$en-font-71$>$.  Text "FamilyVineyard" in $<$color-0$>$, $<$en-font-53$>$.  Text "in Italy" in $<$color-35$>$, $<$en-font-71$>$.  } \\
\hline
\scriptsize{Fig 11, Row3, Col2} & 
\scriptsize{Global: The image is a digital graphic with a split layout, featuring a photograph on the left and text on the right. The photograph depicts a lush green landscape with rows of grapevines. The vines are well-maintained and appear to be in a healthy state, with green leaves and visible grape clusters. In the background, there is a large, historic building with multiple towers and a prominent clock tower, suggesting a castle or a similar architectural structure. The sky is partly cloudy, and the overall atmosphere of the photograph is serene and picturesque.On the left side of the image, there is a dark green section with decorative golden borders. This section includes the text 'BOURGON', 'About us', and '01 / 04'.On the right side of the image, there is text that provides information about the location and climate of the depicted winery. The text is organized into two sections, with the first section titled 'Location' and the second titled 'Climate'. The text in the 'Location' section reads, 'The winery is located in southern Italy, a region that preserved a special rustic charm and unique atmosphere. Wine hills, filled with delicate fragrances, and beautiful scenery over the Mediterranean Sea leave wine lovers with an unforgettable impression.' The text in the 'Climate' section reads, 'The excellent micro-climate of this region and the lime-rich, well-aerated soils provide ideal conditions for producing premium wines. The winemakers produce a wide range of wine grape varieties to everyone’s taste: Pinot Noir, Bonarda, Cabernet Sauvignon, Barbera, Malvasia, Chardonnay, Riesling Italico and Sauvignon Blanc.' The text is in a serif font, which gives it a formal and elegant appearance. The color of the text is a dark shade, contrasting with the lighter background, making it easy to read. The overall style of the image suggests it is likely a promotional or informational graphic for a winery, designed to showcase the beauty of the location and the quality of the wines produced there.
Text: Text "The excellent micro-climate of this region and the lime-rich,well-aerated soils provide ideal conditions for producing premium wines. The winemakers produce a wide range of wine grape varieties to everyone’s taste: Pinot Noir, Bonarda, Cabernet Sauvignon, Barbera, Malvasia, Chardonnay, Riesling Italico and Sauvignon Blanc." in $<$color-2$>$, $<$en-font-71$>$.  Text "The winery is located in southernItaly, a region that preserved aspecial rustic charm and unique atmosphere. Wine hills, filled with delicate fragrances, and beautiful scenery over the Mediterranean Sea leave wine lovers with an unforgettable impression." in $<$color-2$>$, $<$en-font-71$>$.  Text "Climate" in $<$color-2$>$, $<$en-font-53$>$.  Text "BOURGON" in $<$color-0$>$, $<$en-font-71$>$.  Text "Location" in $<$color-2$>$, $<$en-font-53$>$.  Text "About us" in $<$color-0$>$, $<$en-font-53$>$.  Text "01 / 04" in $<$color-35$>$, $<$en-font-53$>$.  } \\
\hline
\scriptsize{Fig 11, Row3, Col3} & 
\scriptsize{Global: The image is a digital graphic with a split layout. On the left side, there is a photograph of a vineyard with green grapevines and clusters of green grapes. The vines are densely packed, and the leaves are lush and green, indicating a healthy plant.On the right side of the image, there is a dark green background with white and gold text. The text is organized into sections with headers such as 'Distance', 'Address', and 'Contact us'. Under the 'Distance' header, there are listed distances to various locations such as 'Airport Mussony 35 km', 'Railway Station Vulcanum 18 km', 'Bus Stop 8 Main Rd. 1 km', and 'Highway 267 Hickory St. 3 km'. The 'Address' section provides a physical address: '8539 West William Lane (30) 2349 4871, (30) 2349 4710 info@bourgon.it'. At the bottom of the right section, the website 'bourgon.it' is listed. The overall style of the image suggests it is a promotional or informational graphic, possibly for a winery or vineyard, given the context of the vineyard photograph. The design is clean and professional, with a clear focus on providing contact and location information. The logo 'BOURGON' is also displayed prominently. The footer shows '04 / 04'.
Text: Text "8539 West William Lane(30) 2349 4871, (30) 2349 4710info@bourgon.it" in $<$color-0$>$, $<$en-font-71$>$.  Text "bourgon.it" in $<$color-0$>$, $<$en-font-71$>$.  Text "AirportRailway Station Bus Stop Highway" in $<$color-0$>$, $<$en-font-71$>$.  Text "Mussony Vulcanum8 Main Rd.267 Hickory St." in $<$color-35$>$, $<$en-font-71$>$.  Text "35 km18 km1 km3 km" in $<$color-0$>$, $<$en-font-71$>$.  Text "Address" in $<$color-35$>$, $<$en-font-53$>$.  Text "BOURGON" in $<$color-0$>$, $<$en-font-71$>$.  Text "Distance" in $<$color-35$>$, $<$en-font-53$>$.  Text "Contact us" in $<$color-0$>$, $<$en-font-53$>$.  Text "04 / 04" in $<$color-35$>$, $<$en-font-53$>$.  } \\
\end{longtable}

\twocolumn
\end{document}